\documentclass{article}

\PassOptionsToPackage{numbers, sort&compress}{natbib}

\usepackage[main, final]{neurips_2026}

\usepackage{graphicx}
\usepackage{amsmath}
\usepackage{amssymb}
\usepackage{amsthm}
\usepackage{bm}
\usepackage{makecell}
\usepackage{multirow}
\usepackage{array}
\usepackage{tabularx}
\usepackage{float}
\usepackage{caption}
\usepackage{subcaption}
\usepackage{wrapfig}
\usepackage[export]{adjustbox}
\usepackage{placeins}
\usepackage{enumitem}
\usepackage{xstring}
\usepackage{xspace}
\usepackage{pifont}
\usepackage[hang,flushmargin]{footmisc}
\usepackage{algorithm}
\usepackage{algorithmicx}
\usepackage{algpseudocode}
\usepackage{listings}
\usepackage{mdframed}
\usepackage{tcolorbox}
\usepackage{pgf}
\usepackage{thmtools}
\usepackage{thm-restate}
\usepackage{lipsum}

\usepackage{kotex}

\usepackage{animate}

\definecolor{tabfirst}{rgb}{1, 0.7, 0.7}
\definecolor{tabsecond}{rgb}{1, 0.85, 0.7}
\definecolor{tabthird}{rgb}{1, 1, 0.7}

\definecolor{GoogleBlue}{RGB}{66, 133, 244}
\definecolor{GoogleRed}{RGB}{234, 67, 53}
\definecolor{GoogleYellow}{RGB}{251, 188, 4}
\definecolor{GoogleGreen}{RGB}{52, 168, 83}

\definecolor{cornellred}{rgb}{0.7, 0.11, 0.11}
\definecolor{cadmiumgreen}{rgb}{0.0, 0.42, 0.24}
\definecolor{aliceblue}{rgb}{0.91, 0.94, 0.97}
\definecolor{darkblue}{rgb}{0.83, 0.89, 0.97}
\definecolor{Red7}{rgb}{0.941, 0.243, 0.243}
\definecolor{Green7}{RGB}{55, 178, 77}
\definecolor{Blue9}{rgb}{0.098, 0.3, 0.9}

\definecolor{codegreen}{rgb}{0, 0.6, 0}
\definecolor{codegray}{rgb}{0.5, 0.5, 0.5}
\definecolor{codepurple}{rgb}{0.58, 0, 0.82}
\definecolor{backcolour}{rgb}{0.95, 0.95, 0.92}

\newcommand{\method}{SierpinskiCam\xspace}

\lstdefinestyle{pytorchstyle}{
    language=Python,
    backgroundcolor=\color{backcolour},
    commentstyle=\color{codegreen}\itshape,
    keywordstyle=\color{codepurple}\bfseries,
    numberstyle=\tiny\color{codegray},
    stringstyle=\color{codepurple},
    basicstyle=\ttfamily\footnotesize,
    breakatwhitespace=false,
    breaklines=true,
    captionpos=b,
    keepspaces=true,
    numbers=left,
    numbersep=5pt,
    showspaces=false,
    showstringspaces=false,
    showtabs=false,
    tabsize=2,
    frame=lines,
    framesep=2mm,
    morekeywords={torch, stack, sum, expand, cat, multinomial, softmax}
}

\usepackage[utf8]{inputenc} %
\usepackage[T1]{fontenc}    %
\usepackage[hidelinks]{hyperref}  
\usepackage{url}            %
\usepackage{booktabs}       %
\usepackage{amsfonts}       %
\usepackage{nicefrac}       %
\usepackage{microtype}      %
\usepackage{xcolor}         %
\usepackage{colortbl}
\usepackage{arydshln}

\title{SierpinskiCam \hspace{-0.6em} \raisebox{-0.4em}{\includegraphics[height=1.3em]{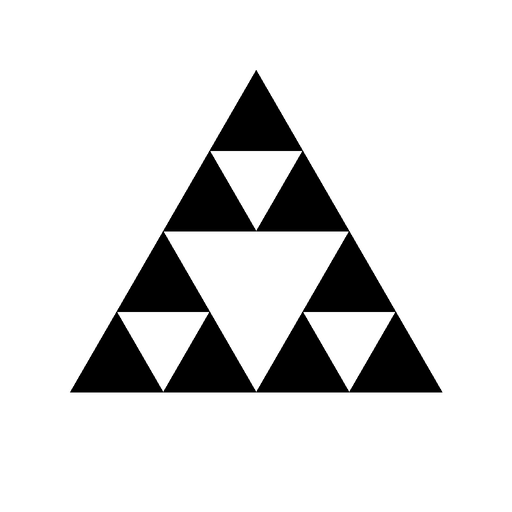}}: Camera-Controlled Video Retaking with Sierpinski Triangle Pattern Cues}

\author{Suttisak Wizadwongsa\textsuperscript{1,2}\thanks{Equal contribution.}, 
Hyelin Nam\textsuperscript{1}\footnotemark[1], 
Supasorn Suwajanakorn\textsuperscript{2}, 
Jeong Joon Park\textsuperscript{1}\thanks{Corresponding author.}
\\
\textsuperscript{1}University of Michigan, Ann Arbor
\qquad
\textsuperscript{2}VISTEC, Thailand
}

\hypersetup{
  colorlinks=true,
  urlcolor=blue,
  citecolor=black,
  linkcolor=black
}

\begin{document}

\maketitle

\vspace{-2em}
\begin{center}
    \includegraphics[width=\textwidth]{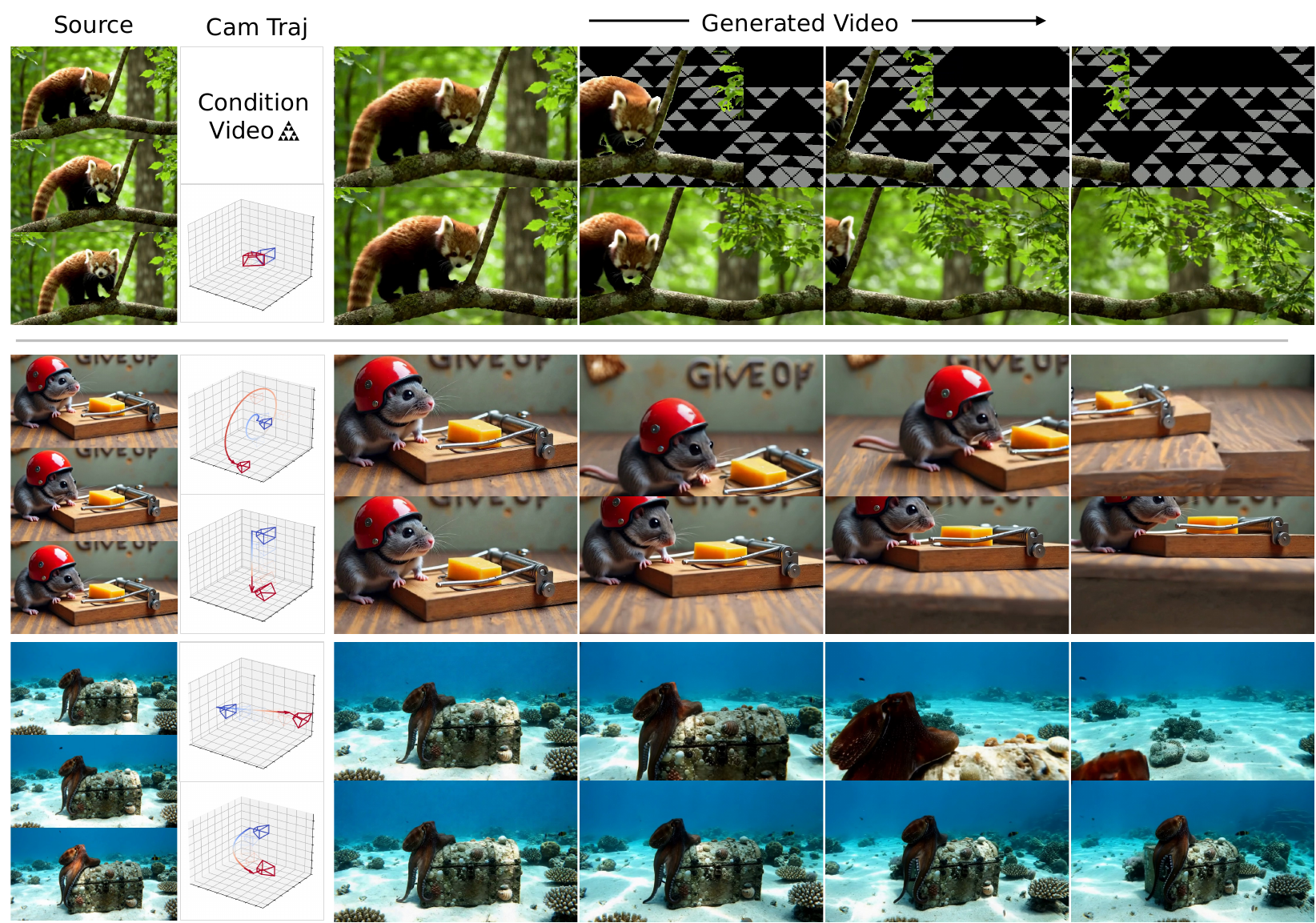}
    \captionof{figure}{
    \textbf{\method for video retake generation.}
    Given a source video and a target camera trajectory (\textcolor{blue}{blue} $\rightarrow$ \textcolor{red}{red}), \method retakes the video under user-defined camera motions.
    Even under large viewpoint changes with sparse source evidence, our Sierpinski textured dome (top \textit{condition video}) and negative rotary position embedding allow faithful following of the target camera trajectory while preserving the original scene content.
    Source videos are generated by Veo3~\citep{DeepMind2025Veo3}. 
    }
    \label{fig:teaser}
\end{center}

\begin{abstract}
Generating novel renderings of a scene along user-defined camera trajectories from a single monocular video, dubbed video retaking, is a compelling but difficult problem in content creation and visual effects.
Existing geometry-guided approaches reconstruct a 4D representation from the source video and render it along the target trajectory to condition video diffusion models. However, this guidance degrades as the target camera departs from the source trajectory, leaving newly revealed regions sparse or entirely missing. We propose \method, which addresses this limitation by augmenting geometry-based guidance with Sierpinski dome texture cues that contain rich trackable features even under large viewpoint changes. 
We further introduce a reference video conditioning mechanism that appends source-video tokens to the target-token sequence and separates the two streams with negative RoPE indices, enabling appearance grounding without architectural modification or per-video adaptation.
Extensive experiments show that \method achieves significant gains in camera controllability, geometric consistency, and video quality across diverse and challenging retaking scenarios. Project page: \href{https://hyelinnam.github.io/SierpinskiCam/}{https://hyelinnam.github.io/SierpinskiCam/}.
\end{abstract}

\section{Introduction}
\label{sec:intro}

Video retaking aims to re-render an existing video under a new camera trajectory while preserving the scene identity, object appearance, and temporal dynamics of the original capture, enabling new creative possibilities in filmmaking, virtual production, and user-generated content. The main challenge is to follow the target camera motion without losing source fidelity, even when the target view reveals regions that were not visible in the original video.

Recent advances in video diffusion models (VDMs)~\cite{kong2024hunyuanvideo, yang2024cogvideox, wan2025wan} provide a strong generative prior for realistic appearance, temporal coherence, and plausible completion of unseen regions.
Adapting these models for video retaking, however, requires addressing two conditioning questions: (i) how to represent and inject the target camera trajectory for precise viewpoint control, and (ii) how to inject the source video for faithful preservation of the original content.

For target camera control, existing methods fall into implicit and explicit formulations. Implicit methods~\cite{van2024generative, bai2025recammaster, park2025redirector} encode the target camera trajectory as pose representations, such as 6-DoF parameters, Plücker rays, or learned embeddings, and inject them into the generative backbone. The model must then infer from data how to disentangle camera motion from scene dynamics and how to interpret the trajectory scale relative to each source video, leaving camera control ambiguous.

Explicit methods~\cite{xiao2024trajectory, chen2025reconstruct, jeong2025reangle, yu2025trajectorycrafter, zhang2025recapture, park2025zero4d} reduce this ambiguity by grounding camera control in geometric transformations of the source video. They estimate depth, point tracks, or point clouds from the source video, warp it under the target camera trajectory, and pass the result to a VDM to refine and synthesize missing regions. This provides an interpretable signal that directly specifies where observed content should appear in the target view. However, geometry derived from a monocular source video provides only partial coverage: when the target camera reveals regions outside the original observation, the warped proxy becomes sparse, with few valid scene points and large empty regions devoid of meaningful motion cues. Such uninformative guidance increases the risk that the generative model fails to follow the target trajectory or hallucinates implausible content.

In addition to target-camera guidance, video retaking must condition on the source video to preserve content appearance and dynamics. This is non-trivial because the source video is not spatially aligned with the target view, making direct feature reuse unreliable. Existing methods address this through channel-wise inputs~\cite{van2024generative}, dedicated reference-attention modules~\cite{yu2025trajectorycrafter}, or per-video fine-tuning~\cite{chen2025reconstruct, zhang2025recapture}. However, these designs require architectural modifications or video-specific adaptation that complicate deployment and limit generalization across different backbones, resolutions, and sequence lengths.

To address these limitations, we introduce SierpinskiCam, which tackles both conditioning questions with a Sierpinski-patterned texture dome and position-disentangled source injection.
For target camera control, we complement geometry-based proxy rendering with a textured dome that fills empty target-view regions with a Sierpinski fractal pattern, a self-similar texture whose multi-scale edges and corners remain trackable across camera distances, turning otherwise uninformative regions into coherent camera-motion cues.
For source preservation, we append source-video tokens to the target-token sequence without architectural modification or per-video adaptation, and assign them negative spatial RoPE indices (NegRoPE) to prevent the model from spuriously attending to source and target tokens that share the same spatial position indices rather than semantically related content.
In summary, our contributions are: (i) a Sierpinski texture dome that provides dense, scale-robust camera-motion cues in newly revealed regions, (ii) NegRoPE, a position-disentangled source injection mechanism that requires no architectural modification or per-video adaptation, and (iii) comprehensive experiments validating our design choices and demonstrating state-of-the-art performance in camera controllability, geometric consistency, and video quality across diverse and challenging retaking scenarios (\autoref{fig:teaser}, ~\autoref{fig:qualitative} and Tab.~\ref{tab:quantitative_comparison}), with 15\% higher user preference scores (\autoref{tab:user_study}).

\section{Related work}
\label{sc:related_work}
\subsection{Camera Control for Video Retaking}
Building on recent advances in video diffusion models (VDMs)~\cite{kong2024hunyuanvideo, yang2024cogvideox, wan2025wan}, recent studies have explored controlling video generation through camera pose conditioning.
We organize relevant works into two categories: implicit methods, in which the model must infer geometry on its own, and explicit methods, which rely on external 3D information as constraints during synthesis.

\paragraph{Implicit Methods.}
Implicit methods~\cite{van2024generative, bai2025recammaster, park2025redirector} inject both sources of control directly into video generative models by conditioning on camera extrinsic parameters and input latents, thereby learning the retaking process through dataset supervision.
Recent approaches following this paradigm construct large-scale synthetic triplet datasets of input videos, target camera trajectories, and corresponding retaken videos using rendering pipelines, and train large diffusion models on these datasets.
However, because camera poses can vary substantially in distribution and scale across datasets, the model must implicitly infer how the target trajectory relates to the source video from supervision alone. 
As a result, their performance remains strongly tied to the coverage of synthetic training data, often leading to limited generalization capability.
ReDirector~\cite{park2025redirector} mitigates this issue by additionally providing the source camera trajectory and explicitly modeling the relative pose difference through positional encoding; however, it still struggles to fully disentangle camera motion from object motion within the video, reflecting an inherent limitation of implicit methods.

\paragraph{Explicit Methods.}
On the other hand, explicit approaches~\cite{xiao2024trajectory, chen2025reconstruct, jeong2025reangle, yu2025trajectorycrafter, park2025zero4d} approximate the underlying scene geometry to reuse the original video content.
They typically estimate depth for each frame, lift the sequence into a 3D proxy, and warp the frames along a new camera trajectory.
Several methods combine this procedure with fine-tuning of generative video models to inpaint or refine regions that become invalid after warping.
While these geometric cues are effective for reusing observed content, they become less informative when the target camera moves far from the original viewpoint or departs substantially from the observed scene.
In such cases, unlike implicit methods that can continue to receive camera pose conditioning, explicit warping provides little additional signal about the intended camera motion.
As a result, the generative model may hallucinate content that is inconsistent with the original scene context.
We target this failure mode by replacing the commonly used black background with texture patterns that allow the model to continue inferring camera motion even in regions where warped content is unavailable.
Recent work~\cite{cao2026freeorbit4d, lin2026vista4d} addresses the same broad limitation from a different angle by improving the 4D proxy itself.
This direction is complementary to ours, as our method provides additional texture-based camera-motion cues that can be applied on top of explicit retaking pipelines.

\subsection{Source Preservation in Video Retaking}
Preserving the source video is a separate challenge from specifying the target camera motion.
Since the source and target views are generally not spatially aligned, prior work differs in how the source video is exposed to the generator.
Generative Camera Dolly~\cite{van2024generative} uses channel-wise latent conditioning, while ReCamMaster~\cite{bai2025recammaster} concatenates source and target clips along the temporal axis so that 3D self-attention can learn source-target relationships. TrajectoryCrafter~\cite{yu2025trajectorycrafter} instead introduces a dedicated Ref-DiT cross-attention module for reference-video conditioning.
Other approaches instead rely on per-video adaptation of the generative prior to preserve source-specific appearance and motion~\cite{chen2025reconstruct, jeong2025reangle, zhang2025recapture}.
These strategies provide effective source preservation, but they either introduce additional conditioning modules, rely on per-video adaptation, or tie source conditioning to a fixed interface.
Our method also uses token concatenation, but makes it architecture-preserving and position-aware by assigning negative RoPE indices to appended source tokens.

\section{Background: Rotary Position Embedding (RoPE)}
\label{sec:prelim}

With the shift from convolutional architectures to transformer-based diffusion models, positional encoding becomes a key component for modeling  relationships among latent tokens.
RoPE~\citep{su2024roformer} is commonly used as a positional encoding in diffusion transformer (DiT) models, providing a simple and general mechanism for preserving relative positional structure among latent tokens.
In DiT-based video diffusion models such as Wan~\cite{wan2025wan}, RoPE is applied to patchified latent tokens before computing attention.
Specifically, after VAE encoding and patchification, each attention head applies linear projections to obtain queries and keys
$q, k \in \mathbb{C}^{N \times (d_{\mathrm{head}}/2)}$, where real and imaginary components represent adjacent channels and $d_{\mathrm{head}}$ is even.
For a token at position $n$ and channel $c$, the rotation is applied via
\begin{equation}
    \bar{q}_{n,c} = q_{n,c} \exp(i \omega_c n), \qquad
    \bar{k}_{n,c} = k_{n,c} \exp(i \omega_c n),
\end{equation}
where the angular frequencies follow an exponential schedule
$\omega_c = 10000^{-2(c-1)/d_{\mathrm{head}}}$ for
$c = 1, \ldots, d_{\mathrm{head}}/2$.
This yields attention scores
\begin{equation}
A'(n,m)
= \operatorname{Re}\!\left[
  \sum_{c=1}^{d_{\mathrm{head}}/2}
  q_{n,c}\,k_{m,c}^{*}\,\exp\!\bigl(i\omega_c(n-m)\bigr)
\right].
\end{equation}
RoPE makes the attention scores depend on token positions only through
the relative offset $(n-m)$, enabling the model to capture spatial
relationships independently of absolute token positions.
In video diffusion models, this formulation extends to latent tokens by assigning RoPE components along the frame, height, and width axes.
In this work, we manipulate only the height and width components, leaving temporal positional encoding unchanged; The details are presented in \autoref{method:subsec:negrope}.

\begin{figure}[t]
    \centering
    \includegraphics[width=\textwidth]{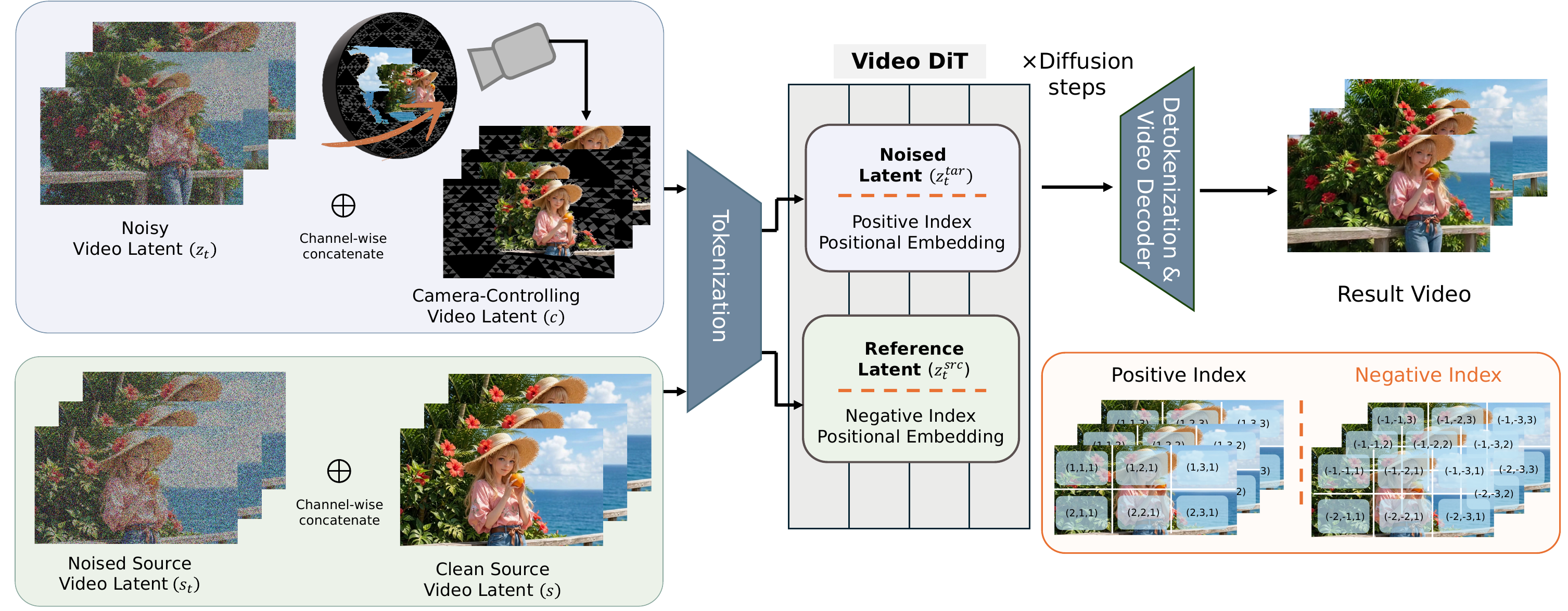}
    \caption{\textbf{Overview of \method}. Our model processes two parallel streams: (1) a target stream (top) where noisy target latents $z_t$ are concatenated with the Sierpinski-dome camera-controlling video $c$, using positive RoPE indices for spatial alignment; and (2) a source stream (bottom) where noised and clean source latents are concatenated using negative RoPE indices. This negative indexing isolates the source content spatially while enabling semantic attention. Both streams are tokenized and processed jointly by the Video DiT to denoise and generate the final result.}
    \label{fig:pipe_fig}
    \vspace{-0.5em}
\end{figure}

\section{Methods}
\label{sec:methods}
In this section, we present \method, a camera-controlled video retaking method that effectively handles target views even when visual guidance from the source video becomes sparse.
In \autoref{method:subsec:sierpinski}, we describe how Sierpinski triangle patterns are rendered under the target trajectory to produce dense, persistent camera-motion cues. 
Then, in \autoref{method:subsec:negrope}, we introduce negative rotary position embedding (NegRoPE) for position-disentangled source injection, which appends source-video tokens to the target-token sequence while assigning them negative spatial RoPE indices to avoid direct positional collisions with target tokens.

\subsection{Generating Camera-Motion Cues Using Sierpinski Triangle Fractal Patterns}
\label{method:subsec:sierpinski}

\paragraph{Geometry-based camera-motion proxy.}
We first construct geometry-based proxies by lifting the source video into 3D and rendering it under the target camera trajectory. 
Following explicit camera-control pipelines such as TrajectoryCrafter~\cite{yu2025trajectorycrafter}, we use dense 3D point clouds reconstructed from monocular depth and sparse 3D point tracks.
For the dense proxy, we use DepthAnything-V3~\cite{lin2025depth} to estimate per-frame depth, extrinsics, and intrinsics, and temporally smooth the estimated source cameras to reduce frame-wise reconstruction jitter.
The source pixels are then forward-splatted from the estimated source cameras to the target camera trajectory, yielding a target-view RGB proxy and a validity mask.
In parallel, we obtain sparse 3D point tracks using SpaTracker-V2~\cite{xiao2025spatialtrackerv2}; projected tracks are rendered as fixed-radius colored points, with colors sampled from their corresponding first-frame pixels.
Although these proxies provide direct geometric guidance where valid source points are available, they often struggle to disentangle global perspective changes from local subject movements, especially when the motion signals are sparse or absent.

\begin{wrapfigure}[11]{r}{0.4\linewidth}
  \centering
  \vspace{-1.2em}
  \includegraphics[width=\linewidth]{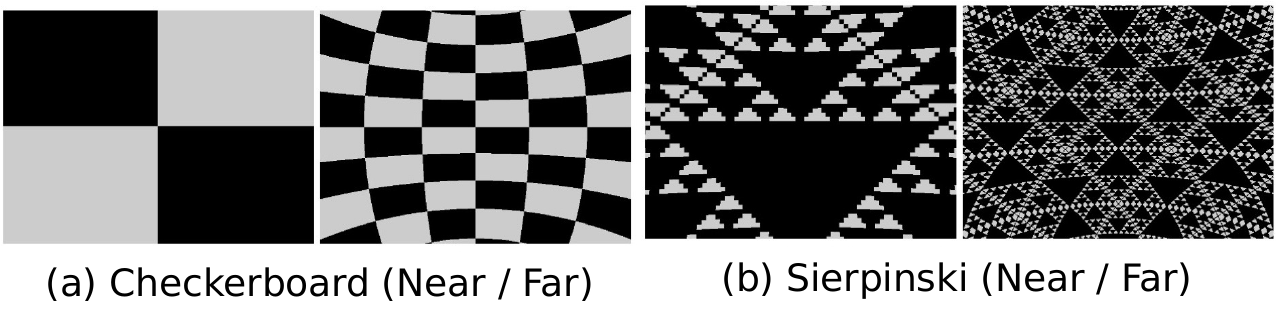}
  \caption{\textbf{Multi-scale robustness.} Unlike the Checkerboard pattern (a), the Sierpinski fractal pattern (b) provides structural details in both near and far views, ensuring camera pose control across scales.}
  \label{fig:pattern_comparison}
  \vspace{-1em}
\end{wrapfigure}

\paragraph{Sierpinski textured dome.}
This motivates an additional camera-motion cue that remains informative beyond the valid coverage of the source-derived geometry.
We texture the dome with a Sierpinski fractal triangle pattern, a structured high-contrast pattern that supplies repeated edges and corners across the image.
This choice follows the common use of Sierpinski patterns in motion tracking, where reliable features from multi-scale fractal pattern help recover geometric transformations across camera distances~\cite{wooley2017scale,zhang2000flexible} (Fig.~\ref{fig:pattern_comparison}).
We analyze this choice in \autoref{exp:subsec:tex_analysis}.

We implement this auxiliary cue by rendering a textured spherical dome around the estimated proxy scene geometry.
For each target camera, we cast a ray from each pixel using the target intrinsics and intersect it with the dome in the proxy 3D coordinate frame.
The dome radius is set to enclose the reliable proxy geometry, with a fixed upper bound to avoid placing the texture excessively far from the camera.
Each intersection point is converted to spherical coordinates and used to sample a 2D texture defined on the dome.
The final dense conditioning image $I_{\mathrm{cond}}$ is obtained by compositing the source-derived warp $I_{\mathrm{warp}}$ with the rendered dome image $I_{\mathrm{dome}}$:
\begin{equation}
I_{\mathrm{cond}} = M \odot I_{\mathrm{warp}} + (1-M) \odot I_{\mathrm{dome}},
\label{eq:dense_conditioning}
\end{equation}
where $M \in \{0,1\}^{H \times W}$ is the forward-warp validity mask; $M=1$ selects source-derived warped pixels, while $M=0$ selects the rendered dome.

We divide the 2D texture into a $16 \times 16$ grid and place a depth-3 recursive Sierpinski motif in each cell, alternating the triangle orientation across rows and columns.
Rendered from the target camera pose, this spherical texture moves coherently with the camera, replacing the commonly used black background with a motion-dependent structured cue wherever the source-derived warp is invalid.
The resulting conditioning video preserves source-derived content where available while supplying a geometrically consistent motion signal in newly revealed regions.
\autoref{fig:ex_pattern} and \autoref{fig:added_traj} in appendix show examples of the resulting textured dome, and further implementation details are provided in \autoref{app:subsec:dome_impl}.

\subsection{NegRoPE for Position-Disentangled Source Injection}
\label{method:subsec:negrope}
The camera-controlling videos are spatially aligned with the generated output and can therefore be treated as aligned conditional signals.
In contrast, the source video, which provides visual characteristics of the scene, is generally captured from a different viewpoint and lacks spatial correspondence with the output video, thus requiring a fundamentally different injection strategy. 
Rather than introducing a dedicated reference-attention module or relying on per-video adaptation, we use an architecture-preserving token-wise injection mechanism inspired by Flux Kontext~\cite{labs2025flux}.
We convert the source and target components into separate token sequences and concatenate them into a shared transformer sequence.

Let $s_t$ denote the noised source video latent and $s$ its clean counterpart.
Similarly, let $z_t$ denote the noised target video latent, and let $c$ denote the clean camera-controlling video latent, which is spatially aligned with the target trajectory.
We construct the source and target token sequences, $x^{\text{src}}_t$ and
$x^{\text{tar}}_t$, respectively, by patchifying the channel-wise concatenation of the corresponding components:
\begin{align}
    x^{\text{src}}_t &= \operatorname{Patchify}([s_t, s]_{\text{channel}}), \\
    x^{\text{tar}}_t &= \operatorname{Patchify}([z_t, c]_{\text{channel}}).
\end{align}
We then concatenate them along the token dimension as $x_t = [x^{\text{src}}_t, x^{\text{tar}}_t]_{\text{token}}$.
This allows them to jointly participate in self-attention and propagate reference information without requiring spatial alignment or fixed frame counts.
This strategy works directly with the base model without any architectural modification.

However, if we assign positional embedding indices to the tokens in $x^{\text{src}}_t$ and $x^{\text{tar}}_t$ in the same way, performance drops significantly.
This is because tokens with the same index are more likely to yield high dot product values, causing the model to incorrectly associate source and target tokens based on positional similarity rather than semantic content. 
To distinguish token types, we employ a simple yet effective trick: target tokens use positive spatial indices $n > 0$, while source tokens use negative spatial indices $-n$, with temporal ordering preserved within each sequence.
A key property of RoPE makes this particularly elegant: the rotary embedding of a negative index $-n$ is equal to the complex conjugate of the rotary embedding of index $n$. Specifically, for queries and keys:
\begin{equation}
\bar{q}_{-n,c} = {q}_{-n,c} \, \exp(-i \omega_c n) = {q}_{-n,c} \, \exp(i \omega_c n)^*,
\end{equation}
\begin{equation}
\bar{k}_{-n,c} = {k}_{-n,c} \, \exp(-i \omega_c n) = {k}_{-n,c} \, \exp(i \omega_c n)^*.
\end{equation}
This encourages attention scores between target and reference tokens to incorporate relative distance while maintaining distinct positional signatures, reducing spurious positional correlations.

\section{Experiments}
\label{sec:exp}

\subsection{Experimental Setup}

\paragraph{Implementation details.}

We adopt Wan2.1 Fun-Control 14B~\cite{wan2025wan} as our base model, using the 1.3B version for ablation study. %
Originally conditioned on text, first-frame images, and depth/human pose controls, the models are fine-tuned via LoRA (rank 64, lr $5 \times 10^{-5}$, 100 epochs) to adapt to our new control signals. To improve robustness, we randomly zero-out the first-frame signal with 10\% probability.
The models were trained and evaluated using NVIDIA L40S GPUs.

We train primarily on the MultiCamVideo~\cite{bai2025recammaster} dataset, a synthetic collection rendered in Unreal Engine~\cite{unrealengine} where each scene features a 3D character captured by ten synchronized cameras.
For each training sample, we randomly select two views: one acts as the source to construct the camera-controlling video, and the other serves as the ground-truth target.
We generate two types of control signals: (1) sparse 3D point tracks via SpaTrackerV2~\cite{xiao2025spatialtrackerv2} and (2) dense point clouds derived from DepthAnything~v3~\cite{lin2025depth}. To mitigate synthetic bias and preserve realistic appearance, we augment the training with real-world static scenes from RealEstate10K \cite{zhou2018stereo}. 
Control videos for this subset are derived from a single video using a first-frame point cloud projected onto subsequent frames.
Crucially, this dataset is used without a reference source video, which enables the model to function effectively in scenarios where no reference is available.
All inputs are resized and center-cropped to $512 \times 320$ pixels with 49 frames.

\paragraph{Evaluation protocol.}
We evaluate our method on two complementary benchmarks.
First, following prior video-retaking works~\citep{bai2025recammaster}, we evaluate on 90 videos from DAVIS~\citep{pont20172017}, using the 10 standard target camera trajectories from ReCamMaster and 4 additional challenging trajectories, yielding 1,260 video--trajectory test cases in total.
DAVIS provides diverse in-the-wild dynamic scenes and serves as our primary benchmark for evaluating retaking quality under realistic object and scene motion.
We additionally evaluate on 56 clips from the Charge dataset~\citep{nazarczuk2026charge}, which provides synchronized target-view videos and ground-truth camera trajectories.
This setting provides an independent benchmark for evaluating
camera controllability and geometric consistency with known target cameras.

Across both benchmarks, we report VBench~\citep{huang2024vbench} metrics for visual quality, Dyn-MEt3R~\citep{park2025steerx} for the geometric consistency of generated retakes, and per-frame MEt3R~\citep{asim2025met3r} for consistency with the source video.
On Charge, where ground-truth target videos are available, we additionally report PSNR and LPIPS. We treat these metrics as complementary, since newly synthesized regions may plausibly differ from the ground truth.
For camera controllability, we report Rotation Error (RotErr), Translation Error (TransErr), and Absolute Trajectory Error (ATE).
Generated-video camera trajectories are estimated using ViPE~\citep{huang2025vipe}, selected through validating three pose estimators on Charge (Appendix~\ref{app:cam_pose}).

\begin{figure}[t]
    \centering
    \includegraphics[width=\textwidth]{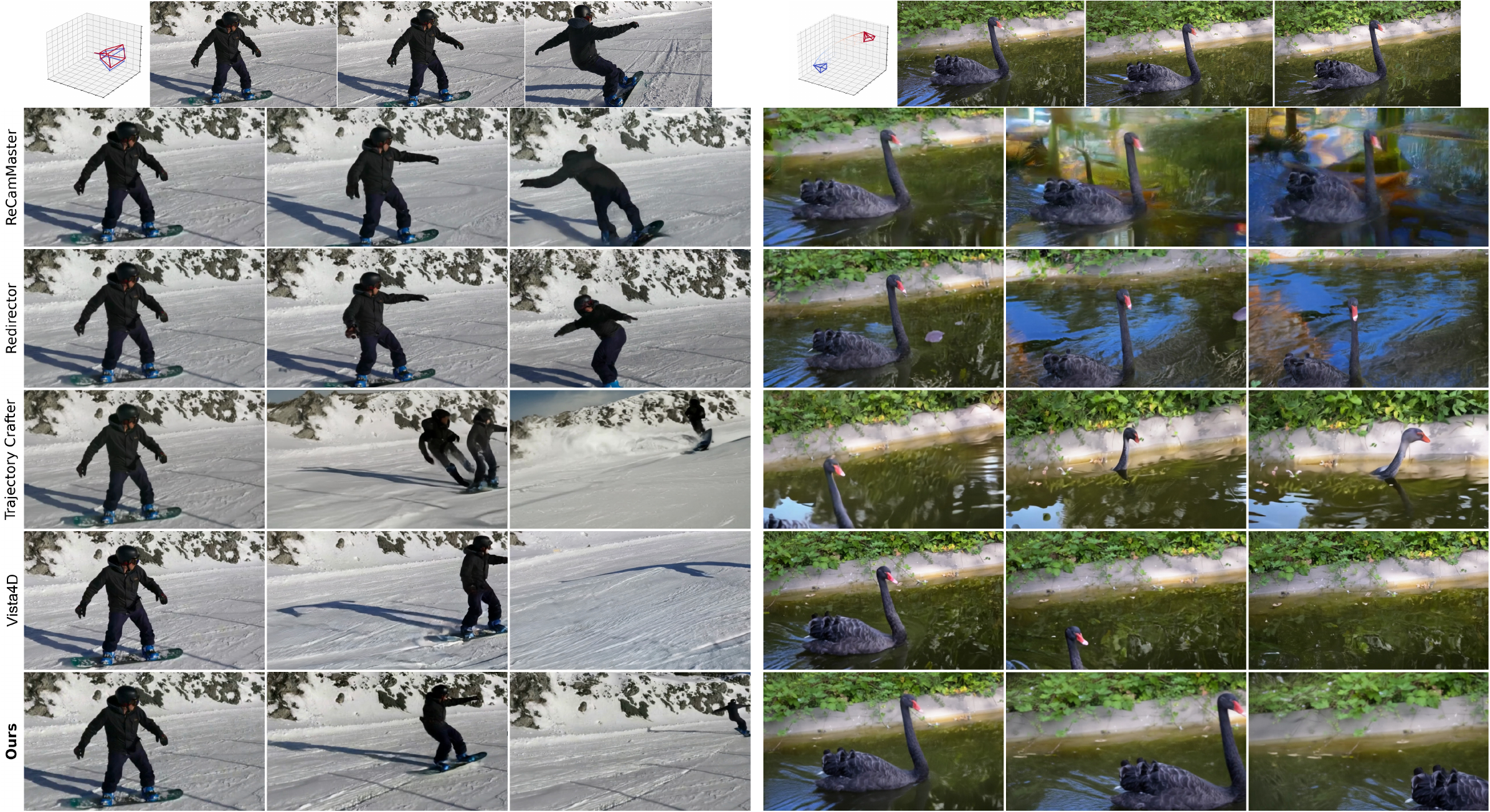}
    \caption{\textbf{Qualitative comparison on the DAVIS dataset.}
    For each scene, top row shows the source video and target trajectory, where the camera moves from \textcolor{blue}{blue} $\rightarrow$ \textcolor{red}{red}.
    Note how prior methods incorrectly put main characters (human/swan) at the center of the frames or create spurious artifacts, while \method accurately follows the user-defined camera paths. Left: the person should go down since the camera is fixed. Right: the swan should leave the frame as the camera moves along the target trajectory.
    }
    \label{fig:qualitative}
\end{figure}

\subsection{Comparison with State-of-the-Art Methods}
We compare our method with recent camera-controlled video-to-video approaches: the implicit methods ReCamMaster~\cite{bai2025recammaster} and ReDirector~\cite{park2025redirector}, and the explicit method TrajectoryCrafter~\cite{yu2025trajectorycrafter} and Vista4D~\cite{lin2026vista4d}.

\vspace{-0.2em}
\paragraph{Qualitative results.}
A known limitation of implicit methods is that they often conflate object and camera motion, causing dynamic objects to appear static or anchored during retaking.
As shown across all examples in \autoref{fig:qualitative}, ReCamMaster and ReDirector tend to keep objects visible even when they should leave the camera frustum; for example, the subject moves forward in the source video but remains nearly stationary in their retaken results.
Although TrajectoryCrafter alleviates this issue with explicit guidance, it becomes unreliable when the guidance becomes sparse; for example, it hallucinates new objects after the original subject leaves the view or fails to sufficiently realize the specified camera trajectory.
Vista4D likewise employs explicit 4D point-cloud guidance, but still exhibits a subject-following bias in challenging cases.
In the snowboard example, its synthesized camera partially follows the moving subject.
This behavior is consistent with a conflict between its explicit point-cloud guidance and implicit source-video prior, which favors subject preservation over strict adherence to the specified camera trajectory.
In contrast, \method preserves a clear separation between camera motion and object dynamics and produces context-consistent videos even under sparse geometric guidance.

\begin{table}[t]
\centering
\vspace{-1.7em}
\caption{
\textbf{Quantitative comparison on DAVIS and Charge.}
We evaluate recent camera-controlled video retaking methods across
visual quality, geometric consistency, and camera accuracy.
DAVIS provides an in-the-wild dynamic-video evaluation, while Charge
provides synchronized target views and ground-truth camera trajectories.
\textbf{Bold}: best; \underline{underline}: second best.
}
\vspace{0.5em}
\label{tab:quantitative_comparison}
\small

\resizebox{\linewidth}{!}{%
\setlength{\tabcolsep}{1pt}

\begin{tabular}{
c
@{\hspace{6pt}}
l
cccccccccc
}
\toprule

&
\multirow{2}{*}{Method}
& \multicolumn{5}{c}{Visual Quality$\uparrow$}
& \multicolumn{2}{c}{Geometric Consistency}
& \multicolumn{3}{c}{Camera Accuracy}
\\

\cmidrule(lr){3-7}
\cmidrule(lr){8-9}
\cmidrule(lr){10-12}

&
& Subject
& Background
& Aesthetic
& Imaging
& Temporal
& \multirow{2}{*}{Dyn-MEt3R$\uparrow$}
& \multirow{2}{*}{MEt3R$\downarrow$}
& \multirow{2}{*}{RotErr$\downarrow$}
& \multirow{2}{*}{TransErr$\downarrow$}
& \multirow{2}{*}{ATE$\downarrow$}
\\

&
& Consistency
& Consistency
& Quality
& Quality
& Flickering
&
&
&
&
\\

\midrule

\cellcolor{blue!12}
& ReCamMaster
& 0.8983
& \textbf{0.9170}
& 0.4875
& 0.4849
& \underline{0.9636}
& 0.7003
& 0.4562
& \underline{0.2106}
& \textbf{1.2312}
& 0.7221
\\

\cellcolor{blue!12}
& ReDirector
& \textbf{0.9033}
& 0.9115
& \underline{0.4943}
& \underline{0.5186}
& 0.9614
& \underline{0.7459}
& \underline{0.3937}
& 0.2262
& \underline{1.2537}
& \underline{0.7087}
\\

\cellcolor{blue!12}
& TrajectoryCrafter
& 0.8625
& 0.9069
& 0.4883
& 0.4883
& 0.9439
& 0.7020
& 0.4421
& 0.2514
& 1.3091
& 0.7798
\\

\cellcolor{blue!12}
& Vista4D
& 0.8819
& 0.9153
& 0.4726
& \textbf{0.5227}
& 0.9463
& 0.7323
& \textbf{0.3113}
& 0.3249
& 1.3443
& 0.7253
\\

\cellcolor{blue!12}
\multirow{-5}{*}{\rotatebox[origin=c]{90}{\textsc{DAVIS}}}
& \textbf{Ours}
& \underline{0.8986}
& \underline{0.9157}
& \textbf{0.4980}
& 0.5061
& \textbf{0.9699}
& \textbf{0.7477}
& 0.4097
& \textbf{0.2058}
& 1.3087
& \textbf{0.6921}
\\

\midrule

\cellcolor{lime!12}
& ReCamMaster
& 0.9134
& 0.9338
& 0.4950
& 0.4397
& \textbf{0.9904}
& 0.7978
& 0.4380
& 11.3046
& 0.4391
& 0.3580
\\

\cellcolor{lime!12}
& ReDirector
& \underline{0.9210}
& \textbf{0.9514}
& 0.4874
& \underline{0.5358}
& 0.9803
& 0.7996
& \textbf{0.2665}
& 12.4306
& \textbf{0.3144}
& 0.2991
\\

\cellcolor{lime!12}
& TrajectoryCrafter
& 0.9204
& \underline{0.9512}
& 0.5197
& 0.4413
& \underline{0.9815}
& \underline{0.8291}
& 0.3699
& 13.6855
& 0.3893
& 0.2985
\\

\cellcolor{lime!12}
& Vista4D
& 0.9168
& 0.9490
& 0.5243
& \underline{0.5430}
& 0.9767
& 0.8231
& 0.3056
& 13.8536
& 0.3873
& \underline{0.2968}
\\

\cellcolor{lime!12}
\multirow{-5}{*}{\rotatebox[origin=c]{90}{\textsc{Charge}}}
& \textbf{Ours}
& \textbf{0.9226}
& 0.9462
& \textbf{0.5466}
& \textbf{0.5568}
& 0.9812
& \textbf{0.8445}
& \underline{0.2957}
& \textbf{11.2907}
& \underline{0.3165}
& \textbf{0.2889}
\\

\bottomrule
\end{tabular}%
}

\vspace{-0.5em}
\end{table}

\vspace{-0.8em}
\paragraph{Quantitative results \& User study.}

\begin{wraptable}{r}{0.44\columnwidth}
    \centering
    \footnotesize
    \caption{
        \textbf{User study on DAVIS.}
        Higher scores indicate stronger preference.
    }
    \label{tab:user_study}
    \vspace{-0.5em}
    \resizebox{0.85\linewidth}{!}{%
        \setlength{\tabcolsep}{1pt}
\begin{tabular}{lccc}
\toprule
\multirow{2}{*}{Method}
& \multicolumn{3}{c}{User Study} \\
\cmidrule(lr){2-4}
& Overall$\uparrow$ & Motion$\uparrow$ & Stab.$\uparrow$ \\
\midrule
ReCamMaster~\cite{bai2025recammaster}
& 2.70 & 3.17 & 2.72 \\
TrajectoryCrafter~\cite{yu2025trajectorycrafter}
& 2.56 & 3.12 & 2.44 \\
ReDirector~\cite{park2025redirector}
& \underline{2.80} & \underline{3.22} & \underline{2.88} \\
\midrule
\textbf{Ours} (SierpinskiCam)
& \textbf{3.23} & \textbf{3.33} & \textbf{3.32} \\
\bottomrule
\end{tabular}

    }
    \vspace{-1.0em}
\end{wraptable}

As reported in \autoref{tab:quantitative_comparison}, \method performs favorably across both DAVIS and Charge, with particularly consistent performance in geometric consistency and camera controllability.
The results on Charge, which provides ground-truth camera trajectories, further support the effectiveness of our structured camera cues for following the intended target motion under a controlled evaluation setting.
The user study further supports these findings from a perceptual perspective.
As shown in \autoref{tab:user_study} our method achieves the highest mean score across all criteria among the evaluated methods.
The study compares \method with ReCamMaster, TrajectoryCrafter, and ReDirector on 10 diverse DAVIS videos with varying camera motions.
A total of 41 participants rated each method on overall preference, camera motion accuracy, and source consistency using a 1--5 Likert scale.

\vspace{-0.5em}
\subsection{Texture Design for Trackable Dome Conditioning}
\label{exp:subsec:tex_analysis}

The dome texture is intended to provide auxiliary trackable structure for the target camera transformation, complementing real image cues that are preserved wherever forward warping yields valid RGB evidence.
We therefore analyze which texture design best supports reliable feature tracking for dome conditioning.
We compare six patterns and a uniform baseline. The textured patterns are the proposed Sierpinski triangle, a recursive circle fractal, random dots, a triangle grid, Perlin noise, and a checkerboard.
To isolate the effect of the pattern itself, we render each texture directly on the virtual dome without scene content or foreground compositing, using 10 canonical camera trajectories from ReCamMaster.
We extract SIFT features~\citep{lowe04sift}, match descriptors using Lowe's ratio test, and perform RANSAC-based geometric verification~\citep{fischler1981ransac}.
For each trajectory, we average the number of verified inliers over adjacent-frame pairs and report the mean and standard deviation across trajectories.

\begin{figure}[t]
    \centering
    \includegraphics[width=\linewidth]{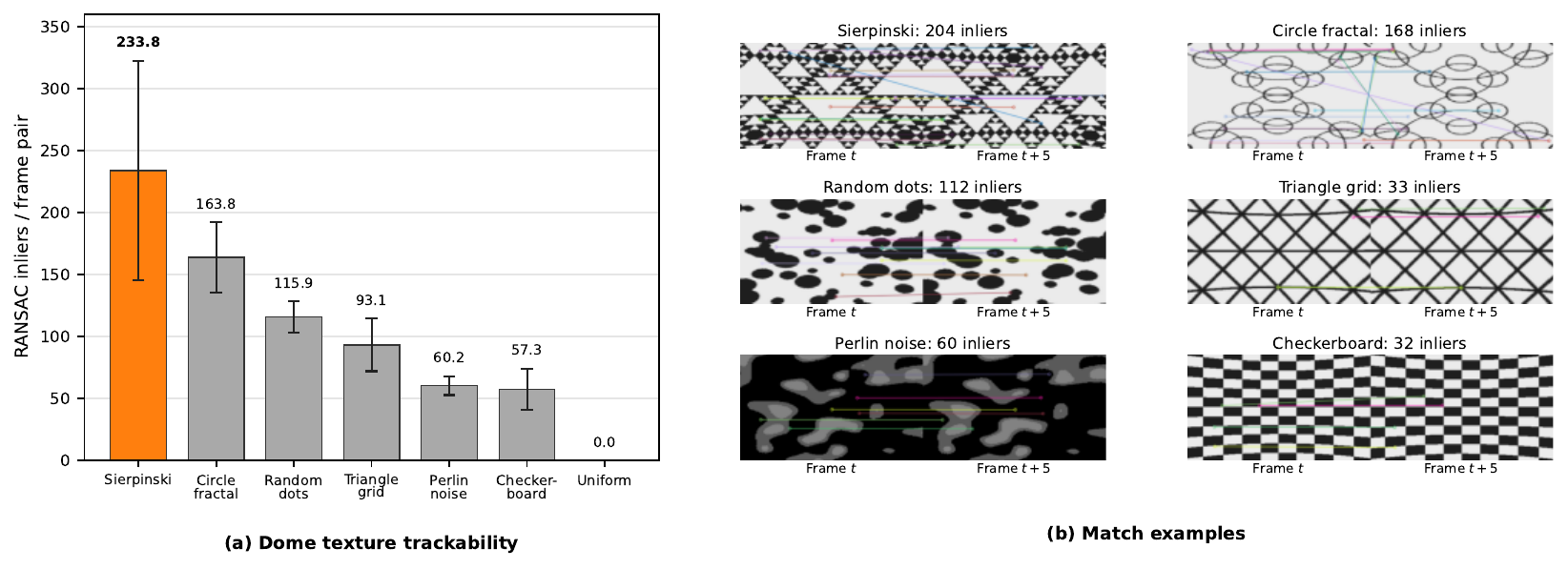}
    \caption{
     \textbf{(a)} Pattern-only trackability over 10 camera trajectories. Ours (Sierpinski) yields the most RANSAC-verified SIFT inliers per frame pair.
    \textbf{(b)} Representative frame-$t$ to frame-$t{+}5$ matches; for clarity, only the top 10\% inliers by Lowe ratio are drawn, while titles report total
      inliers.
    }
    \label{fig:dome_texture_trackability}
    \vspace{-1em}
\end{figure}

\autoref{fig:dome_texture_trackability} summarizes the resulting pattern trackability and shows representative matches under a larger temporal gap.
The Sierpinski pattern achieves the highest mean number of verified correspondences, with 233.8 inliers per frame pair, followed by the circle-fractal pattern with 163.8.
Random dots, the triangle grid, Perlin noise, and the checkerboard yield 115.9, 93.1, 60.2, and 57.3 inliers, respectively.
Thus, the two recursive-multi-scale patterns provided more verified correspondences on average than the regular-grid and stochastic controls.
In particular, the Sierpinski pattern produces a larger pool of geometrically consistent correspondences than the smoother circle-fractal baseline, consistent with its dense multi-scale edges and corners.
Based on this evidence, we adopt the Sierpinski triangle as the default dome texture.
It supplies dense, geometrically verifiable cues in dome-conditioned regions, providing additional camera-motion evidence beyond the warped RGB signal.

\begin{table*}[t]
\centering
\caption{
    \textbf{Ablations on Charge.}
    \textbf{(a)} The Sierpinski dome provides the strongest geometric
    consistency and target-view fidelity among the tested textures,
    whereas Random Dots yields the lowest ATE.
    \textbf{(b)} NegRoPE best preserves content and 3D consistency;
    removing source spatial RoPE improves pose accuracy at a substantial
    cost to content consistency and fidelity.
    All variants are independently trained under matched settings.
    \textbf{Bold}: best; \underline{underline}: second best
    within each ablation.
}
\label{tab:component_ablation}
\small
\resizebox{\linewidth}{!}{%
\setlength{\tabcolsep}{2.5pt}
\begin{tabular}{
    c
    @{\hspace{5pt}}
    l
    cccccccc
}
\toprule
&
\multirow{3}{*}{Variant}
& \multicolumn{2}{c}{Visual Quality}
& \multicolumn{2}{c}{Geometric Consistency}
& \multicolumn{2}{c}{Camera Accuracy}
& \multicolumn{2}{c}{Fidelity}
\\
\cmidrule(lr){3-4}
\cmidrule(lr){5-6}
\cmidrule(lr){7-8}
\cmidrule(lr){9-10}

&
& Subject
& Background
& \multirow{2}{*}{Dyn-MEt3R$\uparrow$}
& \multirow{2}{*}{MEt3R$\downarrow$}
& \multirow{2}{*}{RotErr$\downarrow$}
& \multirow{2}{*}{ATE$\downarrow$}
& \multirow{2}{*}{PSNR$\uparrow$}
& \multirow{2}{*}{LPIPS$\downarrow$}
\\

&
& Cons.$\uparrow$
& Cons.$\uparrow$
&
&
&
&
&
\\
\midrule

\cellcolor{gray!12}
& Solid Black (No Texture)
& \textbf{0.9251}
& \underline{0.9504}
& 0.8218
& \underline{0.5622}
& 12.2411
& 0.2982
& 17.2483
& \underline{0.4455}
\\

\cellcolor{gray!12}
& Checkerboard
& 0.9195
& 0.9462
& \underline{0.8273}
& 0.5773
& 11.7870
& \underline{0.2821}
& 17.1045
& 0.4505
\\

\cellcolor{gray!12}
& Perlin Noise
& 0.9188
& 0.9471
& 0.8112
& 0.5741
& 12.0491
& 0.3000
& 17.0997
& 0.4514
\\

\cellcolor{gray!12}
& Random Dots
& \underline{0.9215}
& 0.9484
& 0.8201
& 0.5745
& \underline{11.5109}
& \textbf{0.2769}
& \underline{17.3298}
& 0.4492
\\

\cellcolor{gray!12}
\multirow{-5}{*}{
    \rotatebox[origin=c]{90}{\textbf{(a) Dome}}
}
& \textbf{Sierpinski (Ours)}
& 0.9208
& \textbf{0.9566}
& \textbf{0.8310}
& \textbf{0.3492}
& \textbf{11.2265}
& 0.2951
& \textbf{17.9980}
& \textbf{0.4433}
\\

\midrule

\cellcolor{gray!12}
& Frame Concat
& \underline{0.8923}
& \underline{0.9471}
& 0.7590
& 0.3916
& 12.7208
& 0.2991
& 17.1581
& \underline{0.5305}
\\

\cellcolor{gray!12}
& No RoPE Mod.
& 0.8168
& 0.9268
& 0.6726
& 0.6683
& 11.3965
& \underline{0.2918}
& 10.7398
& 0.7064
\\

\cellcolor{gray!12}
& Offset RoPE (+4096)
& 0.8118
& 0.9206
& 0.7278
& 0.5666
& 13.1224
& 0.3455
& 16.6975
& 0.5903
\\

\cellcolor{gray!12}
& No Spatial RoPE
& 0.8880
& 0.9322
& 0.7660
& 0.7548
& \textbf{11.1133}
& \textbf{0.2665}
& 13.6331
& 0.6782
\\

\cellcolor{gray!12}
& Temporal Offset (+0.1)
& 0.8921
& 0.9466
& \underline{0.7783}
& \underline{0.3883}
& \underline{11.2191}
& 0.3062
& \underline{17.2652}
& 0.5377
\\

\cellcolor{gray!12}
\multirow{-5}{*}{
    \rotatebox[origin=c]{90}{\textbf{(b) RoPE}}
}
& \textbf{NegRoPE (Ours)}
& \textbf{0.9208}
& \textbf{0.9566}
& \textbf{0.8310}
& \textbf{0.3492}
& 11.2265
& 0.2951 
& \textbf{17.9980}
& \textbf{0.4433}
\\

\bottomrule
\end{tabular}%
}
\vspace{-0.5em}
\end{table*}

\subsection{Ablation Study}
\label{exp:subsec:ablation}

\paragraph{Dome texture.}
We first compare five dome textures through controlled end-to-end
ablations on Charge: solid-black filling, checkerboard, Perlin noise,
random dots, and Sierpinski.
Whereas \autoref{exp:subsec:tex_analysis} analyzes pattern trackability in isolation, this end-to-end ablation directly measures how the choice of dome texture affects the trained video-generation model.
As shown in \autoref{tab:component_ablation}-(a), Sierpinski achieves
the strongest geometric consistency and target-view fidelity among the tested textures, while also providing the best background consistency and rotational accuracy.
Random Dots yields a lower ATE, and solid-black filling slightly better preserves subject consistency, but neither provides the same overall performance across geometry, camera control, and fidelity.
These results show that the benefit does not arise merely from filling disoccluded regions with an arbitrary dense texture.

\paragraph{RoPE indexing strategies.}
We next examine how the RoPE assignment of source-video tokens affects
generation under the same Charge protocol.
We compare NegRoPE with Frame Concat and four alternative indexing
schemes: shared source--target RoPE indices, a large positive spatial
offset, removal of source spatial RoPE, and a small temporal offset.
Except for Frame Concat, these variants use the same source-token
concatenation and differ only in their RoPE index assignment, directly isolating the effect of NegRoPE.
As shown in \autoref{tab:component_ablation}(b), NegRoPE achieves the strongest content consistency, geometric consistency, and target-view fidelity among the evaluated strategies.
Removing source spatial RoPE yields lower camera-pose errors, but
substantially degrades content preservation, geometric consistency,
and fidelity.
We observe cases in which this variant generates subjects with
altered identities or backgrounds inconsistent with the source video,
suggesting that it uses source-video information less reliably.
NegRoPE avoids this trade-off by retaining spatial cues within the
source stream while assigning source and target tokens to disjoint
spatial RoPE domains.

\begin{wraptable}{r}{0.44\columnwidth}
    \vspace{-1em}
    \centering
    \footnotesize
    \caption{
        \textbf{ReAngle-A-Video extension.}
        The Sierpinski cue improves translation accuracy and ATE.
    }
    \label{tab:reangle_a_video}
    \resizebox{0.85\linewidth}{!}{%
        \centering
\small
\resizebox{0.9\linewidth}{!}{%
\setlength{\tabcolsep}{3pt}
\begin{tabular}{lccc}
\toprule
Method & RotErr$\downarrow$ & TransErr$\downarrow$ & ATE$\downarrow$ \\
\midrule
ReAngle-A-Video
& \textbf{0.1591} & 1.3968 & 2.1409 \\
w/ Sierpinski
& 0.1613 & \textbf{1.3087} & \textbf{1.6015} \\
\bottomrule
\end{tabular}%
}

\vspace{0.3em}
\label{tab:camera_accuracy_comparison}
    }
    \vspace{-1.0em}
\end{wraptable}

\subsection{Extension to an explicit camera-control method}
To examine whether our structured camera cue can extend beyond our own pipeline, we instantiate it in an existing explicit method.
Specifically, we use ReAngle-A-Video~\citep{jeong2025reangle}, which provides training code. 
We follow the original codebase and modify only the construction of the conditioning video used for optimization.
We evaluate this extension on 7 DAVIS scenes with 5 target camera trajectories.
The results in \autoref{tab:reangle_a_video} suggest that the structured cue can improve trajectory adherence, particularly in translation and ATE.
This experiment shows the plug-in applicability of the proposed structured camera cue to an explicit camera-control pipeline beyond our own implementation.

\section{Conclusion and Limitations}
\label{sec:conclusion} 

In this work, we presented \method, a dynamic video retaking framework that combines reconstructed 4D geometry with Sierpinski-patterned dome conditioning.
The Sierpinski dome provides trackable multi-scale structure that complements warped RGB evidence and helps guide the target camera motion. 
Our analyses validate this design and confirm its benefit for video model conditioning.
We further introduce NegRoPE to disentangle source and target positions, allowing unaligned source videos to guide generation without spurious positional matching.
Results show that \method maintains strong camera control, geometric consistency, and visual fidelity across diverse scenarios, including large camera deviations where little original scene evidence remains available.
As an inherent consequence of relying on video generation and 4D estimation models, \method also inherits their limitations.
For example, complex motion in original video or unreliable 4D reconstruction can lead to degraded results.
Future advances in both model families may help mitigate these limitations.

\section{Acknowledgement}
\label{sec:ack} 

This work was supported in part by the National Science Foundation under CAREER Award No.~2544116. We also acknowledge scholarship support from SCB Public Company Limited and PTT Public Company Limited (Thailand).

{\small
\bibliographystyle{plainnat}
\bibliography{main}
}

\newpage
\appendix
\clearpage

\section{Motivation}
\begin{figure}[h]
    \centering
    \includegraphics[width=0.8\textwidth]{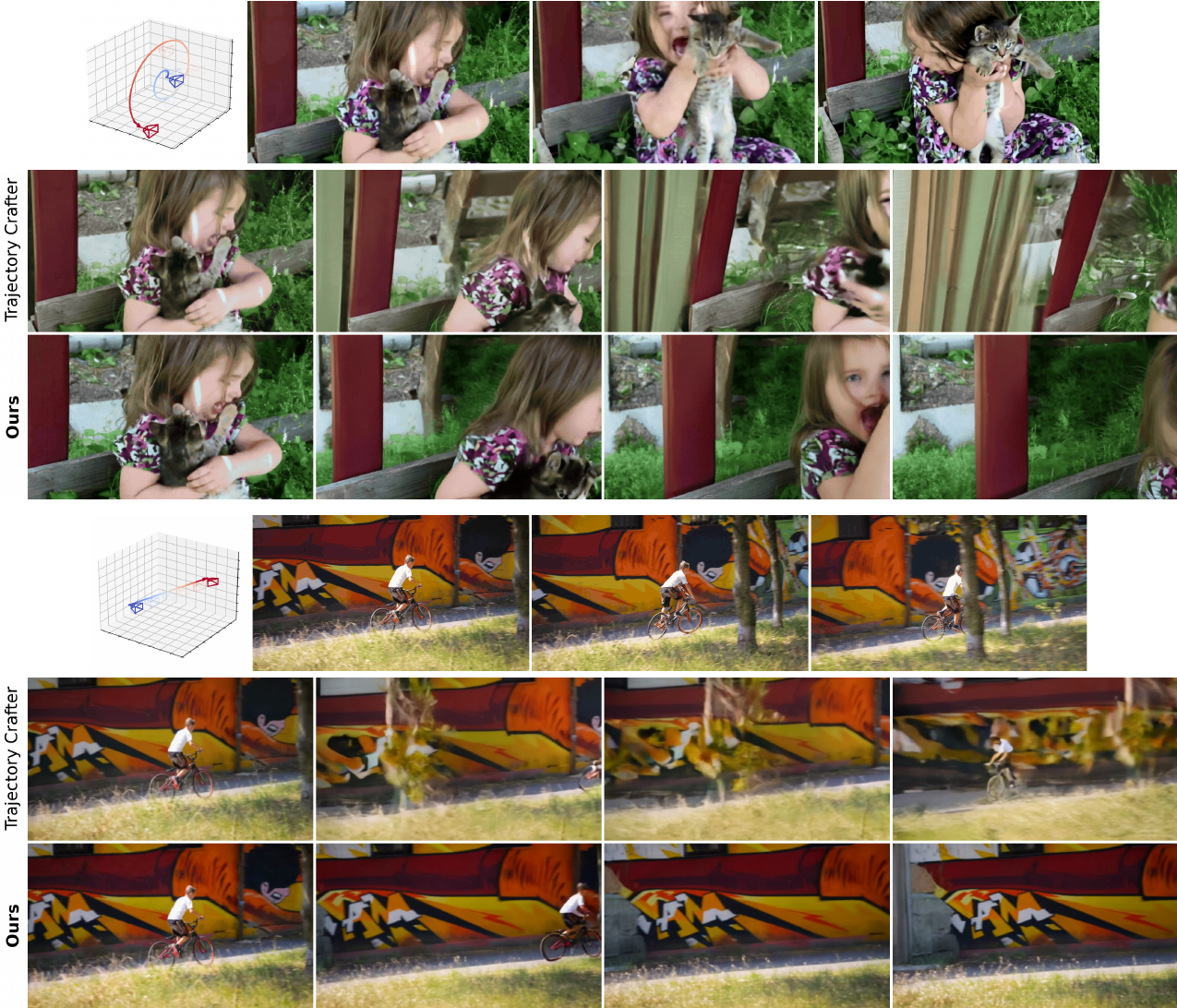}
    \caption{\textbf{Failure cases of TrajectoryCrafter.} When the target camera moves beyond the original scene coverage, it may hallucinate content or fail to follow the camera pose, especially as the main object becomes small or leaves the target-view frustum.}
    \label{fig:motivation}
    \vspace{-0.5em}
\end{figure}

We observed that TrajectoryCrafter~\cite{yu2025trajectorycrafter} often fails when the target camera trajectory extends far beyond the original scene coverage. As shown in \autoref{fig:motivation}, when the camera moves away from the main object or the object leaves the target-view frustum, the model hallucinates content or fails to follow the given camera pose.
We attribute this partly to the common practice of rendering the proxy over a black background: when the warped RGB signal covers only a small part of the target view, the remaining regions provide little cue about depth or camera motion.
We therefore add a Sierpinski-textured dome to the proxy rendering, providing trackable multi-scale structure in empty regions and making the target camera motion more explicit.

\section{Implementation Details}
\label{app:impl_detail}
\subsection{Textured Dome Generation}
\label{app:subsec:dome_impl}
We provide additional details on the construction of the textured dome used for the dense conditioning signal. The dome is implemented as a sphere in the proxy 3D coordinate frame.
For each target camera, we cast one ray per output pixel
using the target intrinsics and intersect the ray with the sphere. The intersection point is converted to spherical coordinates, and its latitude and longitude are used to sample a 2D texture image. In our implementation, the sphere radius is set as
\begin{equation}
R = \min(R_{\max}, d_{\max}),
\end{equation}
where $d_{\max}$ is the maximum distance of confident points in the estimated proxy geometry and $R_{\max}=30$. This clipping prevents the dome from being placed excessively far from the camera while still enclosing the reliable proxy
geometry. In fixed-radius analyses, we directly set $R=30$.

The final dense conditioning frame is obtained by compositing the forward-warped source proxy with the rendered dome:
\begin{equation}
I_{\mathrm{cond}} = M \odot I_{\mathrm{warp}} + (1-M) \odot I_{\mathrm{dome}},
\end{equation}
where $I_{\mathrm{warp}}$ is the source-derived target-view proxy, $I_{\mathrm{dome}}$ is the rendered textured dome, and
$M \in \{0,1\}^{H \times W}$ is the forward-warp validity mask.
Pixels with $M=1$ use the warped source content, while pixels with $M=0$ are filled by the dome.

The default dome texture is a tiled Sierpinski-triangle pattern. We generate a $2048 \times 2048$ RGB texture and divide it into a $16 \times 16$ grid.
Each grid cell contains a recursive Sierpinski triangle with recursion depth 3.
The triangle orientation alternates across rows and columns, and additional half-cell-shifted triangles are inserted according to the row/column parity to avoid large texture regions with a single dominant orientation. 
This produces a repeated multi-scale pattern of corners and edges across the entire sphere.

\begin{figure}[h]
    \centering
    \includegraphics[width=\textwidth]{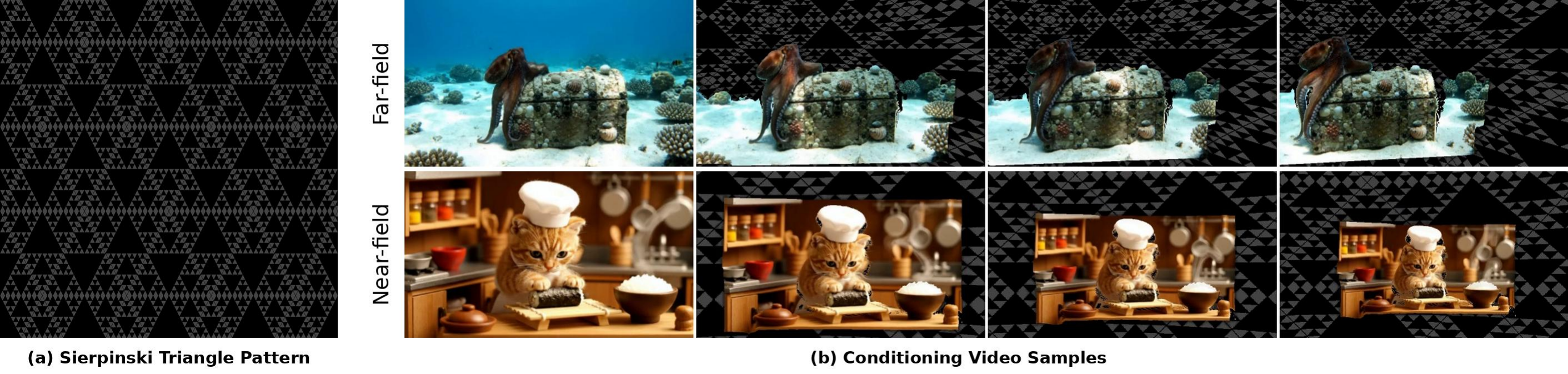}
    \caption{\textbf{Geometry proxy design and conditioning video examples.} \textbf{(a)} The Sierpinski triangle pattern used as a geometry proxy for camera motion conditioning.
    Its self-similar structure provides rich spatial cues without scene-specific texture, enabling generalizable control.
  \textbf{(b)} Example conditioning videos under two representative depth regimes: far-field (distant background) and near-field (close background), illustrating the range of depth variation handled by our framework.}
    \label{fig:ex_pattern}
    \vspace{-0.5em}
\end{figure}

Concretely, let $s$ denote the grid-cell size. For each cell, we recursively subdivide a triangle of size $s$ into three child triangles of size $s/2$ until depth 3 is reached, at which point the leaf triangles are rasterized into the texture.
The orientation of the root triangle is alternated by row index, and half-cell-shifted triangles with the opposite orientation are added on alternating columns.
This staggered tiling reduces large empty bands and keeps local high-contrast structures visible under different target camera views.
\autoref{fig:ex_pattern} shows the Sierpinski triangle pattern used as our
geometry proxy alongside conditioning video examples across far-field and near-field scenes.
Triangle size varies with the scene depth (smaller for far-field and larger for near-field), providing the model with spatial depth cues.

\section{Experimental Details}
\label{app:exp_detail}

\subsection{Caption Preparation}
\label{app:subsec:caption}
The videos used in our experiments are automatically captioned using CogFlorence\footnote{\url{https://huggingface.co/thwri/CogFlorence-2.2-Large}}, a fine-tuned variant of Microsoft’s Florence-2 model, and the resulting captions are used for downstream evaluation.

\subsection{Additional Evaluation Trajectories}
\label{app:additional-trajectories}

\begin{figure}[t]
    \centering
    \includegraphics[width=\textwidth]{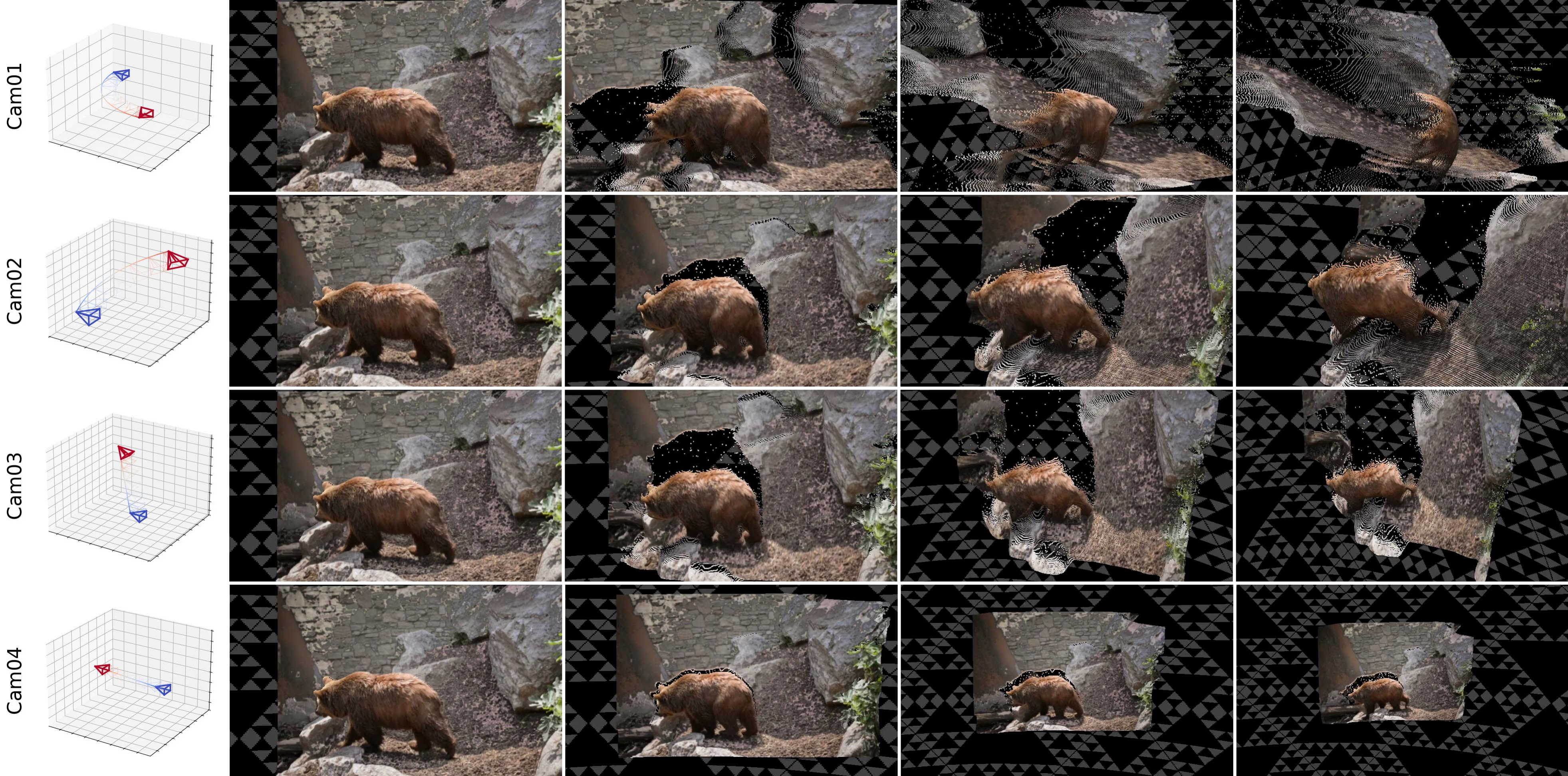}
    \caption{\textbf{Additional camera trajectories used for evaluation.} Each row shows the geometry proxy frames used as video conditioning input for four additional camera trajectories.
    The sampled frames illustrate the spatial extent and viewpoint variation induced by each trajectory.}
    \label{fig:added_traj}
    \vspace{-0.5em}
\end{figure}

To evaluate robustness under more challenging camera motions, we augment the 10 standard ReCamMaster trajectories with 4 additional target trajectories.
These trajectories are designed to stress large viewpoint changes and out-of-frame extrapolation, including large panning, zoom-out, and whirl motions.
Figure~\ref{fig:added_traj} visualizes the added trajectories used in our DAVIS evaluation.

\subsection{Camera Pose Evaluation}
\label{app:cam_pose}

We evaluate camera controllability by estimating the camera trajectory of each generated video and comparing it with the reference trajectory. 
To select the pose estimator, we benchmark DA3~\citep{lin2025depth}, ViPE~\citep{huang2025vipe}, and MegaSaM~\citep{li2025megasam} on Charge, which provides ground-truth camera trajectories.

\begin{wraptable}{r}{0.46\textwidth}
\centering
\caption{Pose estimator validation on Charge. \textbf{Bold}: best; \underline{underline}: second best.}
\label{tab:pose_estimator}
\small
\resizebox{\linewidth}{!}{%
\begin{tabular}{lccc}
\toprule
Estimator
& RotErr$\downarrow$ Mean (Median)
& TransErr$\downarrow$ Mean (Median)
& ATE$\downarrow$ Mean (RMSE) \\
\midrule
DA3
& 0.6442 (0.4051)
& \underline{0.1961} (0.1368)
& 0.1751 (0.2025) \\
ViPE
& \textbf{0.6151} (0.4184)
& \textbf{0.1919} (0.1292)
& \textbf{0.1691} (0.1966) \\
MegaSAM
& \underline{0.6283} (0.4043)
& \textbf{0.1919} (0.1521)
& \underline{0.1717} (0.1937) \\
\bottomrule
\end{tabular}%
}
\vspace{-0.5\baselineskip}
\end{wraptable}

As shown in Table~\ref{tab:pose_estimator}, ViPE achieves the lowest mean RotErr and ATE and ties for the lowest mean TransErr. We therefore use ViPE throughout our evaluation. This choice also avoids coupling the evaluation to DA3, which is used to construct our conditioning videos.

\subsection{User Study}
\label{app:user_study}

For user study, participants were recruited through Prolific\footnote{
https://www.prolific.com/}
and were paid approximately \$4 each.
They were shown the source video, an explanation of the target camera trajectory, and four retaking results.
The four candidate videos were randomly ordered and labeled as A, B, C, and D for each question.
Participants were then asked to rate each candidate independently using the following instructions.

\begin{table*}[t]
\centering
\begin{minipage}[t]{0.48\textwidth}
    \centering
    \begin{tcolorbox}
        \raggedright
        \textcolor{Blue9}{\textbf{User study instructions:}\\}
        Please rate each result from 1 to 5 (1 = Very poor, 5 = Excellent) based on three criteria: \\

        \vspace{1mm}
        \textbf{1. Overall preference} \\
        Please rate each result based on your overall preference, considering visual quality, realism, temporal coherence, and similarity to the source video. \\

        \vspace{1mm}
        \textbf{2. Camera motion accuracy} \\
        Please rate each result based on how well its camera motion follows the target trajectory described in the question. \\

        \vspace{1mm}
        \textbf{3. Stability \& source consistency} \\
        Please rate each result based on temporal stability and consistency with the source video. A higher score means less flickering, fewer unexpected changes in the subject/background, and better preservation of the source identity and geometry.
    \end{tcolorbox}
    \caption{Instructions used for the user study.}
    \label{tab:user_study_instructions}
\end{minipage}
\hfill
\begin{minipage}[t]{0.45\textwidth}
    \centering
    \includegraphics[width=\linewidth]{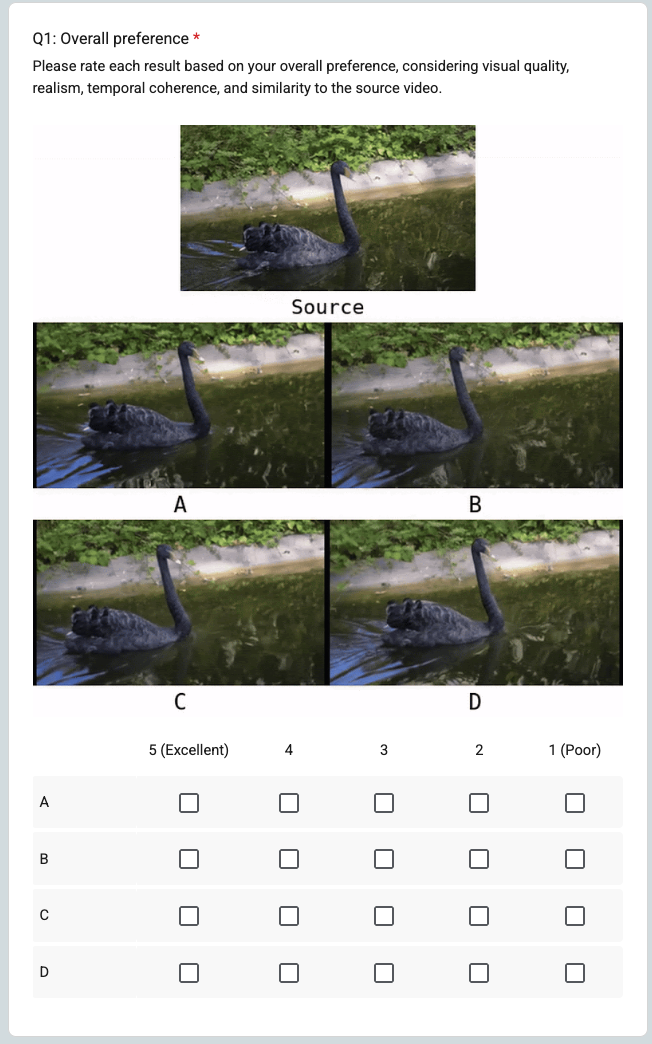}
    \captionof{figure}{Screenshot of the user study interface.}
    \label{fig:user_study_screenshot}
\end{minipage}
\end{table*}

\subsection{ReAngle-A-Video Extension Details.}
We extend ReAngle-A-Video to the DAVIS 480p setting and evaluate whether the generated videos preserve the intended camera motion. For each DAVIS sequence, we use the first 49 frames and resize/crop them to the ReAngle training resolution of $720 \times 480$. We evaluate five camera trajectories: left, right, up, down, and zoom-out. 
For the vanilla baseline, we use the default ReAngle-A-Video conditioning pipeline. For our variant, we replace the black disocclusion fill with the proposed Sierpinski-pattern background while keeping the same target camera trajectory and Depth Anything 3 (DA3) -estimated scene geometry. Both methods use DA3 depth and intrinsics for constructing the conditioning videos. We use 150 optimization steps.
To quantify camera-motion adherence, we estimate camera poses from the generated videos using DA3.
We then compare the estimated trajectory against the target camera trajectory. Rotation error is reported in radians and is computed relative to the first frame to reduce sensitivity to fixed camera-coordinate convention differences.

\section{Additional Qualitative Results}
\label{app:additional_qual_results}

We provide additional qualitative results and comparisons with selected baselines beyond the main evaluation dataset.
These results further illustrate the generalization of \method across different data sources and camera-motion scenarios.
We also include supplementary videos to better visualize camera-control behavior and temporal dynamics.

\begin{figure}[t]
    \centering
    \includegraphics[width=0.8\linewidth]{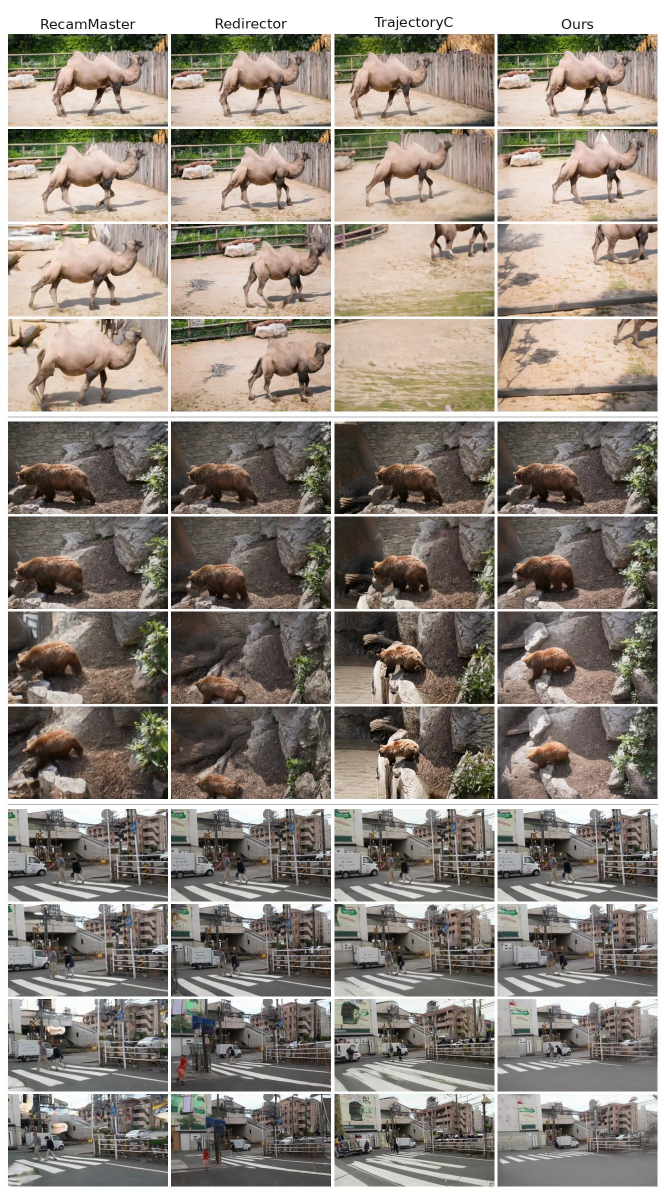}
    \caption{\textbf{Additional comparison on newly added challenging camera trajectories.}}
    \label{fig:added_traj_more}
\end{figure}

\begin{figure}[htbp]
    \centering
    \begin{tabular}{@{}c@{\hspace{2pt}}c@{\hspace{2pt}}c@{\hspace{2pt}}}
        \raisebox{0.cm}{\rotatebox{90}{%
            \sffamily\tiny\parbox{1.1cm}{\centering ReCamMaster}%
        }} &
        \includegraphics[width=0.17\textwidth]{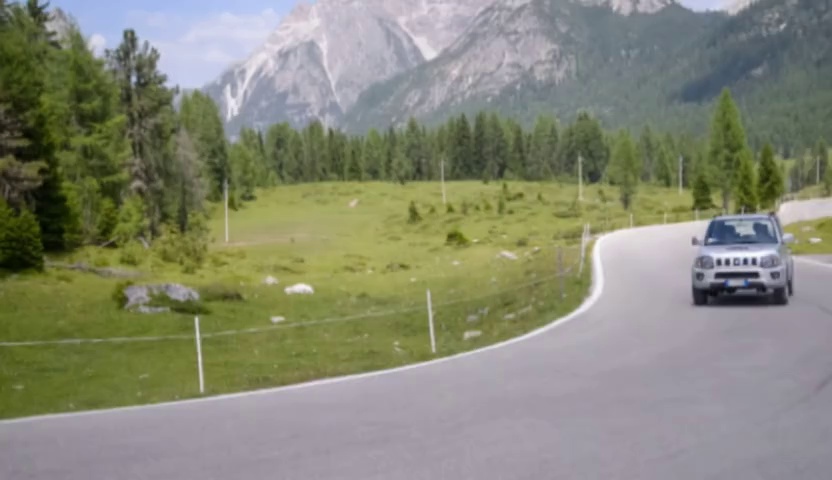}\hspace{0.5pt}  
        \includegraphics[width=0.17\textwidth]{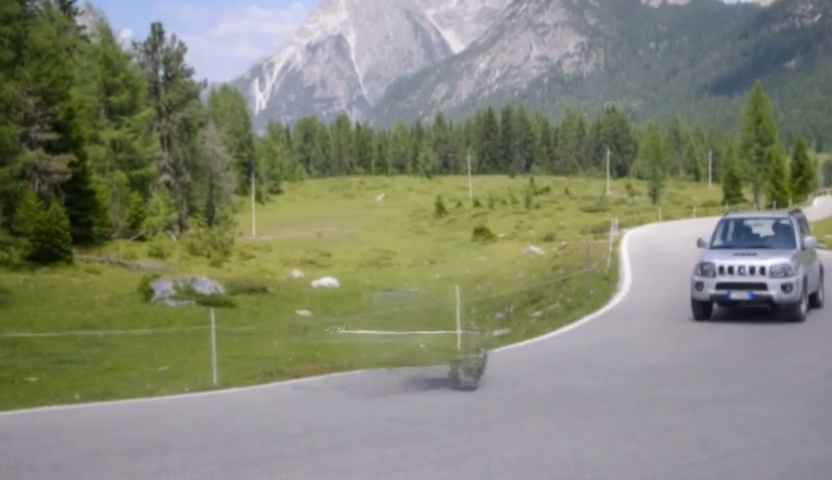}\hspace{0.5pt}  
        \includegraphics[width=0.17\textwidth]{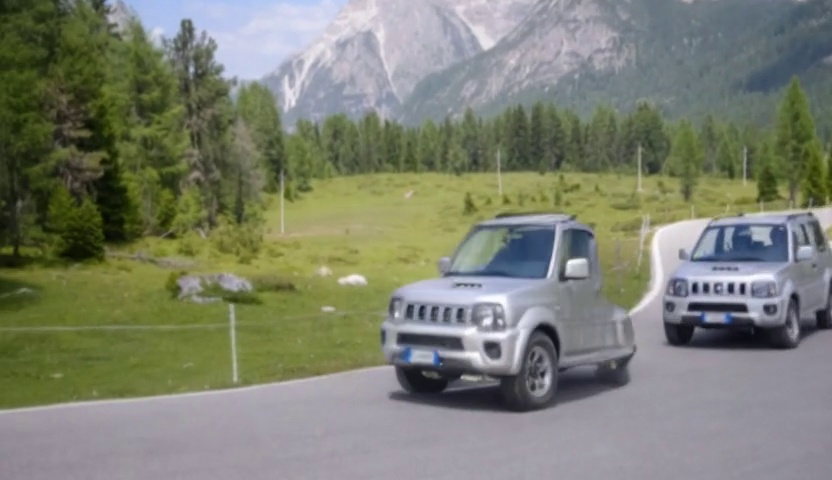}\hspace{0.5pt}  
        \includegraphics[width=0.17\textwidth]{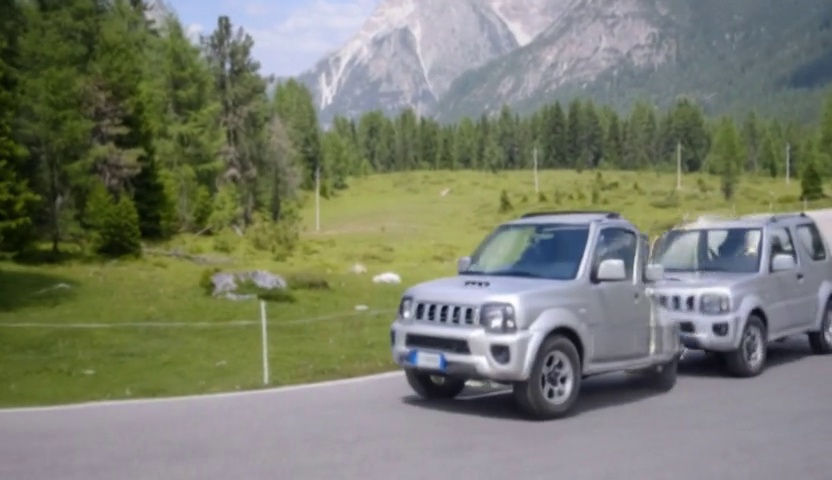}\hspace{0.5pt}  
        \includegraphics[width=0.17\textwidth]{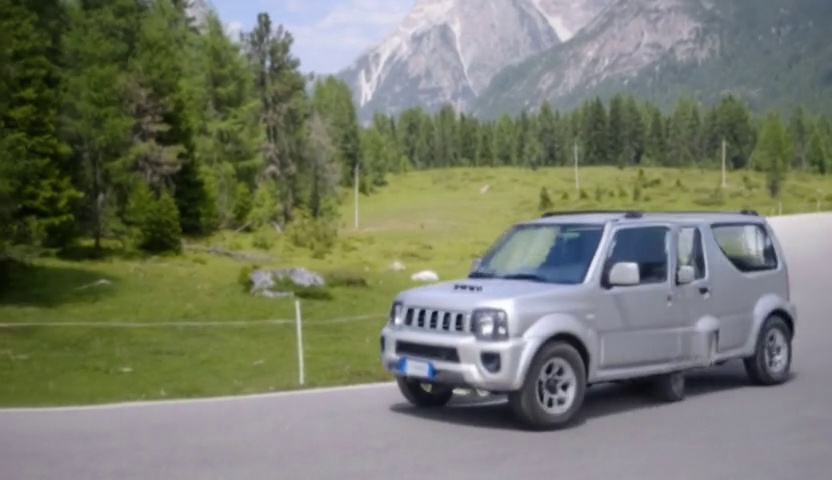} 
        \\
        \raisebox{0.cm}{\rotatebox{90}{%
            \sffamily\tiny\parbox{1.1cm}{\centering ReDirector}%
        }} &
        \includegraphics[width=0.17\textwidth]{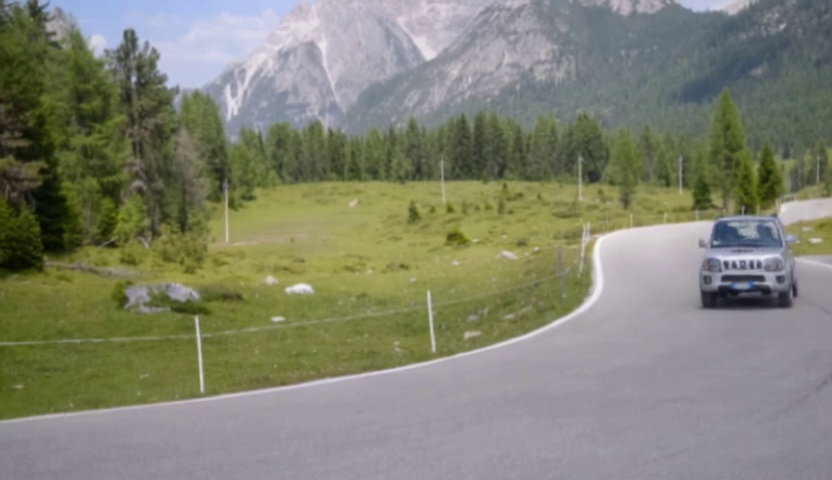}\hspace{0.5pt}  
        \includegraphics[width=0.17\textwidth]{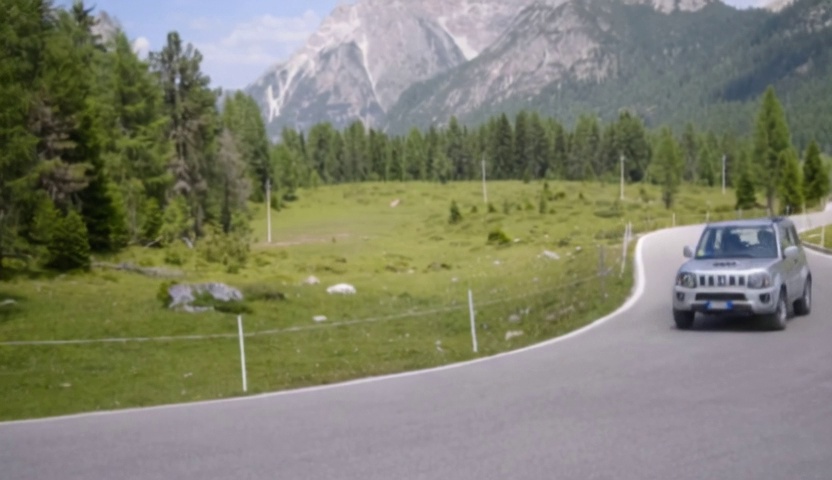}\hspace{0.5pt}  
        \includegraphics[width=0.17\textwidth]{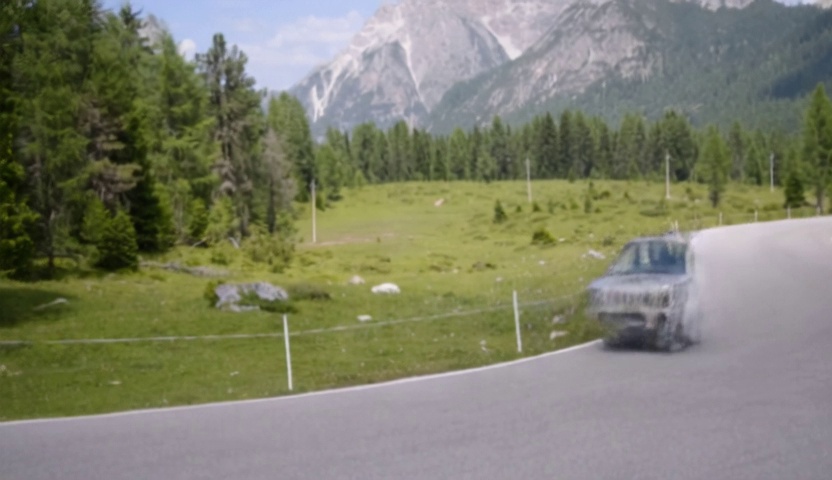}\hspace{0.5pt}  
        \includegraphics[width=0.17\textwidth]{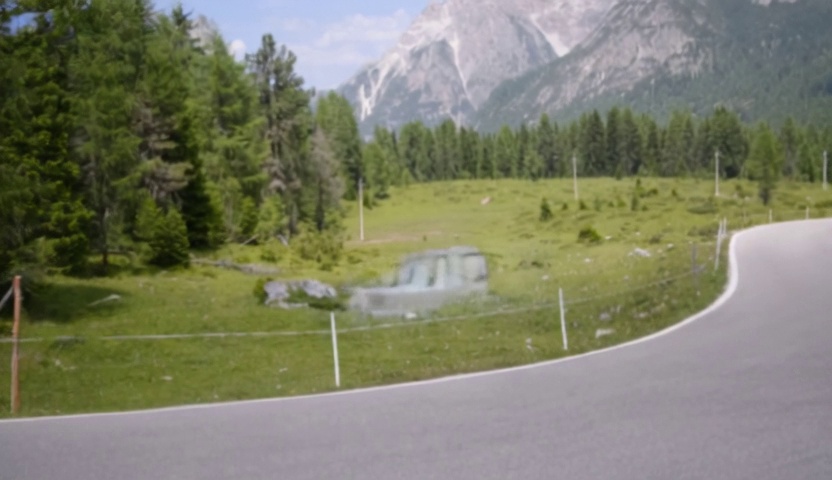}\hspace{0.5pt} 
        \includegraphics[width=0.17\textwidth]{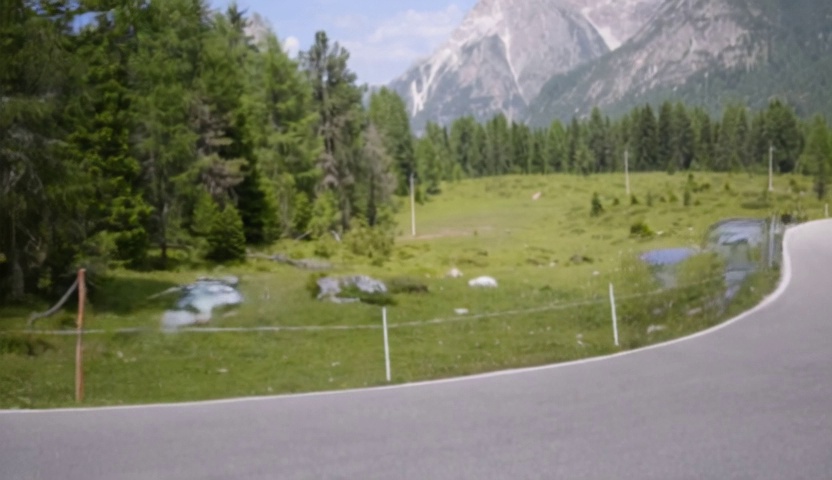}\hspace{0.5pt}  
        \\
        \raisebox{0.cm}{\rotatebox{90}{%
            \sffamily\tiny\parbox{1.1cm}{\centering TrajectorC}%
        }} &
        \includegraphics[width=0.17\textwidth]{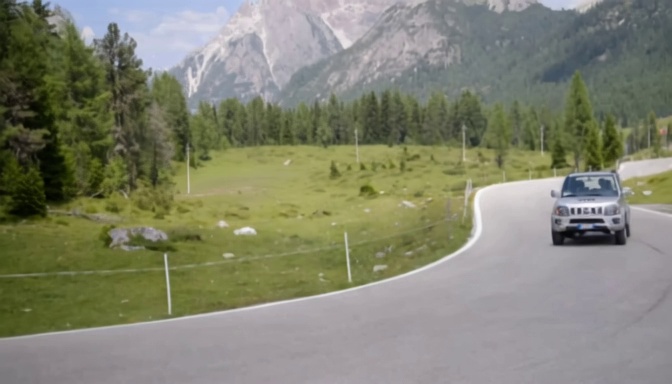}\hspace{0.5pt}  
        \includegraphics[width=0.17\textwidth]{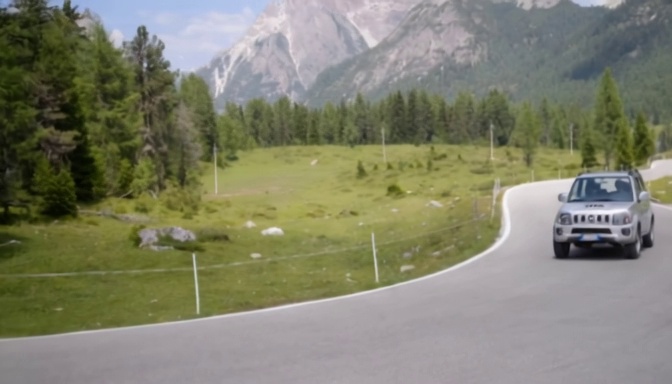}\hspace{0.5pt}  
        \includegraphics[width=0.17\textwidth]{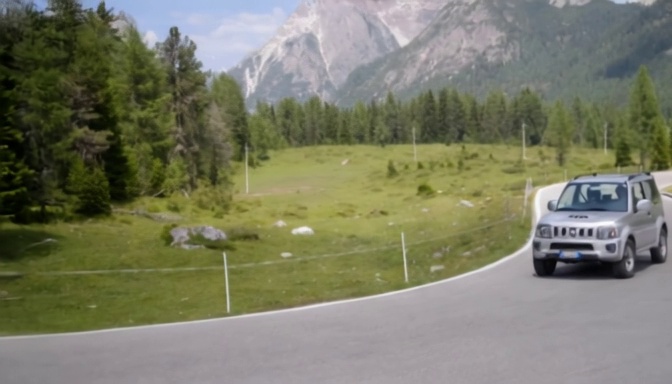}\hspace{0.5pt}  
        \includegraphics[width=0.17\textwidth]{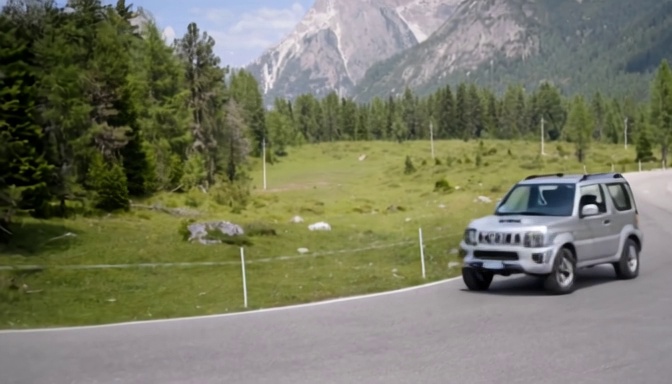}\hspace{0.5pt}    
        \includegraphics[width=0.17\textwidth]{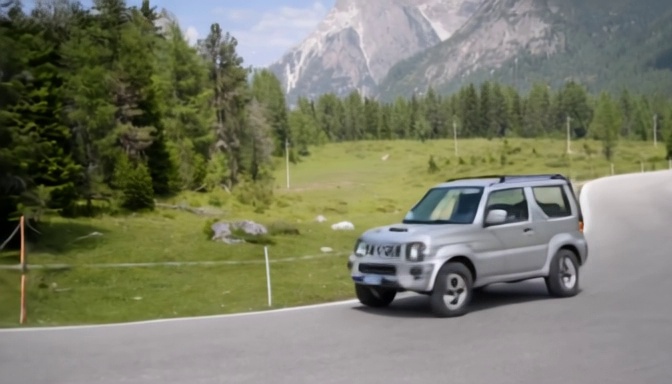}  
        \\
        \raisebox{0.cm}{\rotatebox{90}{%
            \sffamily\tiny\parbox{1.1cm}{\centering Ours}%
        }}  &
        \includegraphics[width=0.17\textwidth]{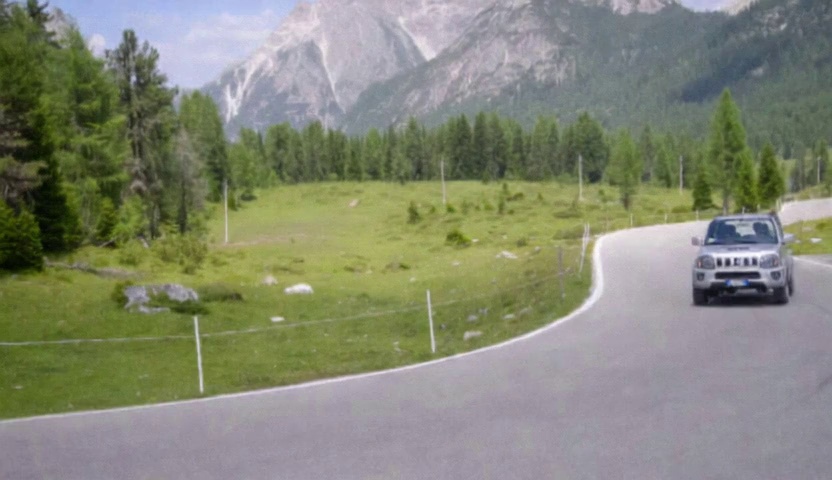}\hspace{0.5pt}  
        \includegraphics[width=0.17\textwidth]{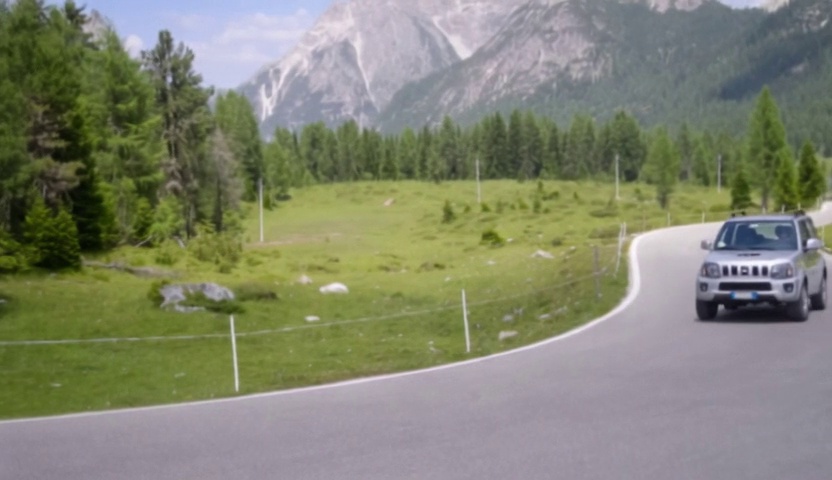}\hspace{0.5pt}  
        \includegraphics[width=0.17\textwidth]{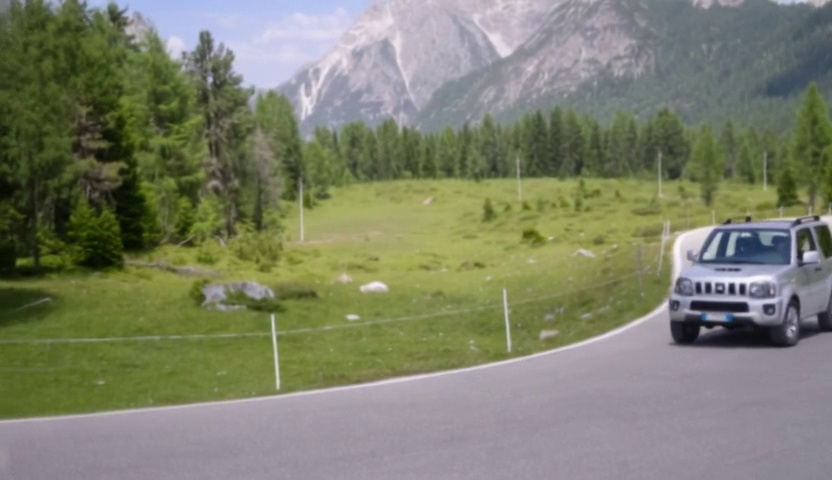}\hspace{0.5pt}  
        \includegraphics[width=0.17\textwidth]{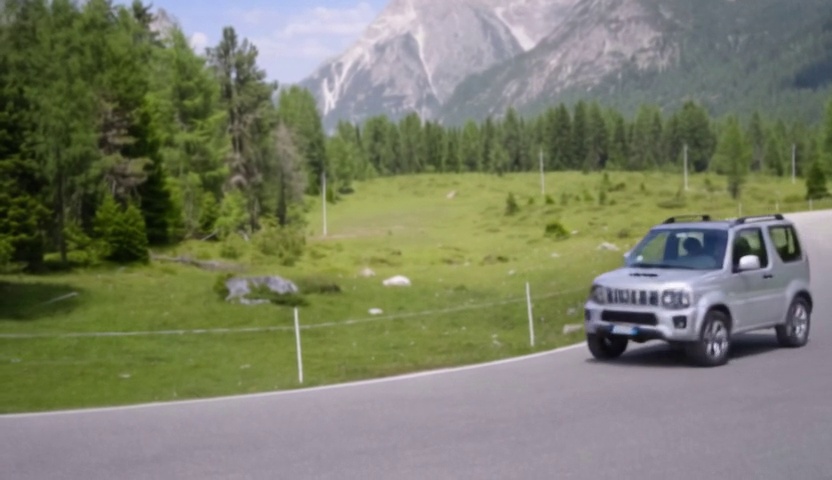}\hspace{0.5pt} 
        \includegraphics[width=0.17\textwidth]{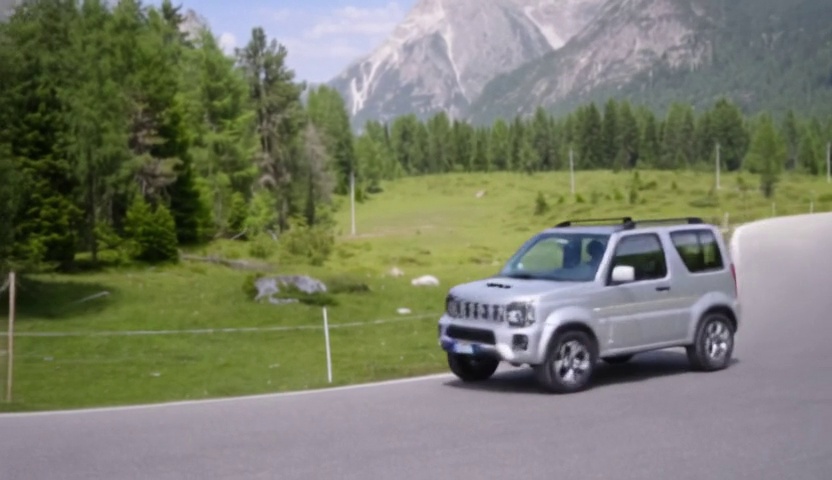}\hspace{0.5pt}  
        \\
        \\
        \raisebox{0.cm}{\rotatebox{90}{%
            \sffamily\tiny\parbox{1.1cm}{\centering ReCamMaster}%
        }}  &
        \includegraphics[width=0.17\textwidth]{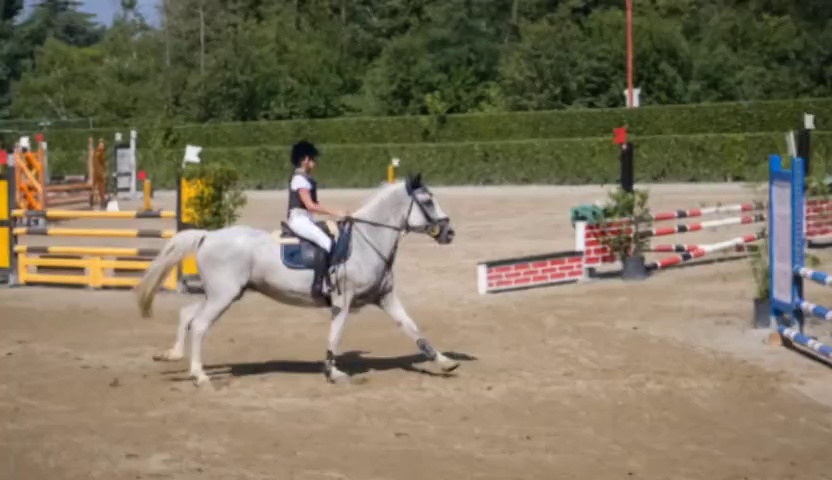}\hspace{0.5pt} 
        \includegraphics[width=0.17\textwidth]{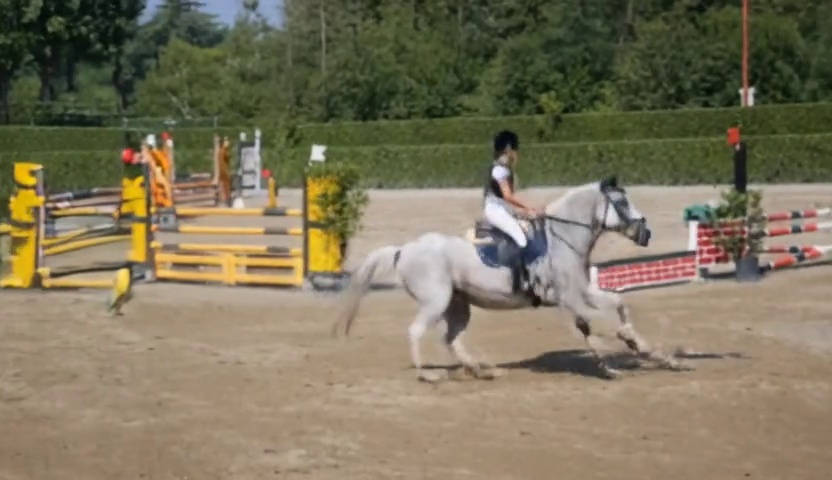}\hspace{0.5pt}
        \includegraphics[width=0.17\textwidth]{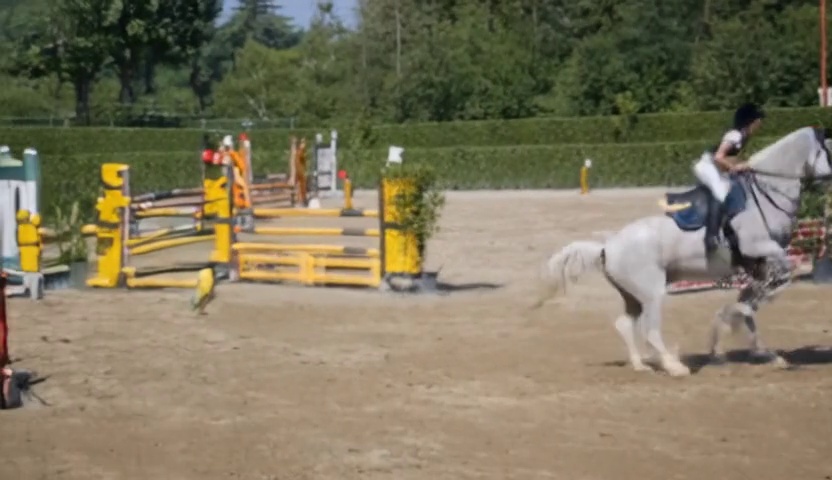}\hspace{0.5pt} 
        \includegraphics[width=0.17\textwidth]{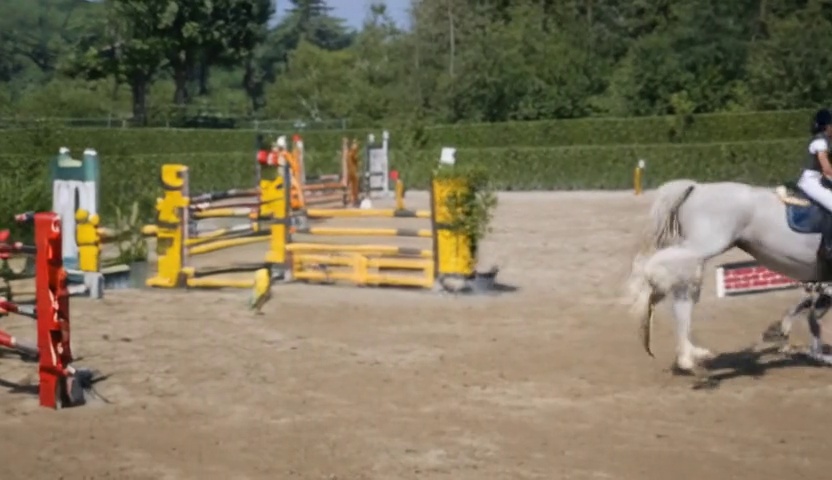}\hspace{0.5pt}   
        \includegraphics[width=0.17\textwidth]{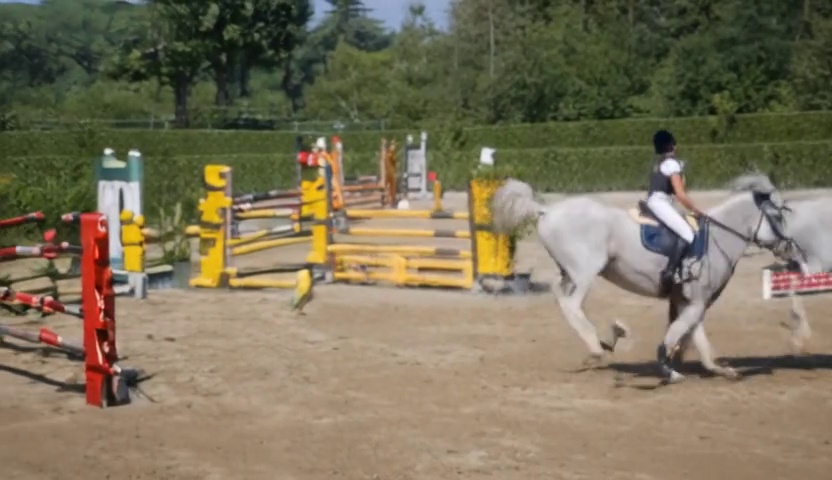}\hspace{0.5pt}   
        \\
        \raisebox{0.cm}{\rotatebox{90}{%
            \sffamily\tiny\parbox{1.1cm}{\centering ReDirector}%
        }} &
        \includegraphics[width=0.17\textwidth]{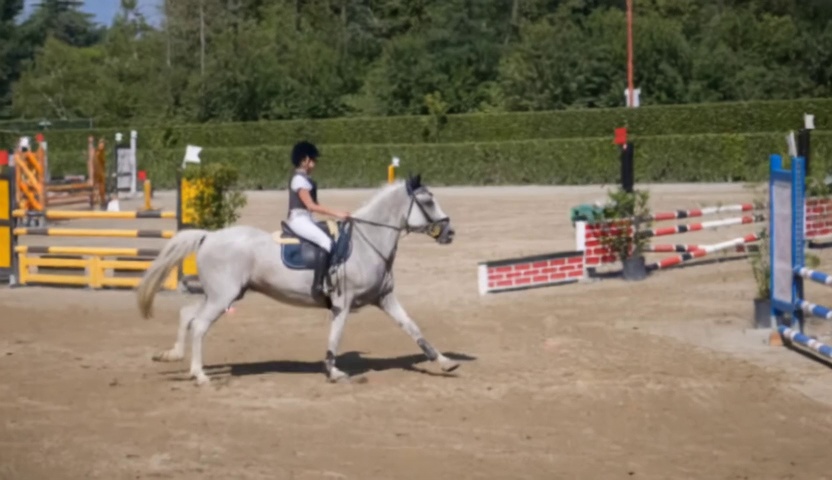}\hspace{0.5pt}  
        \includegraphics[width=0.17\textwidth]{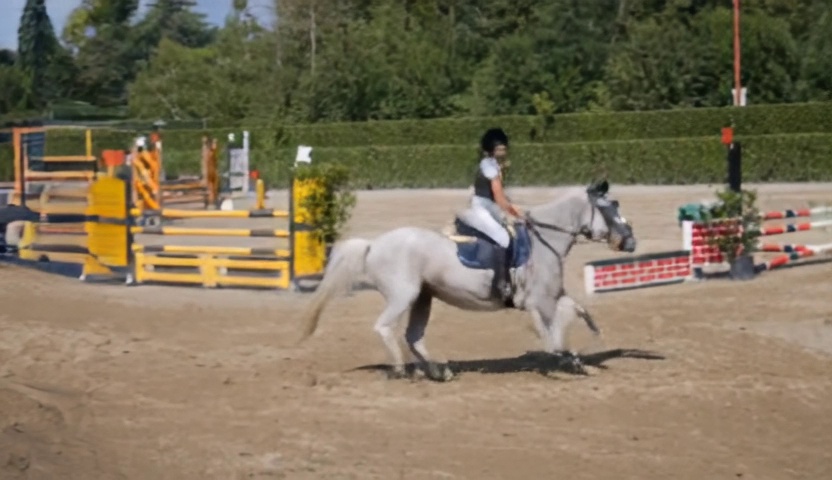}\hspace{0.5pt}  
        \includegraphics[width=0.17\textwidth]{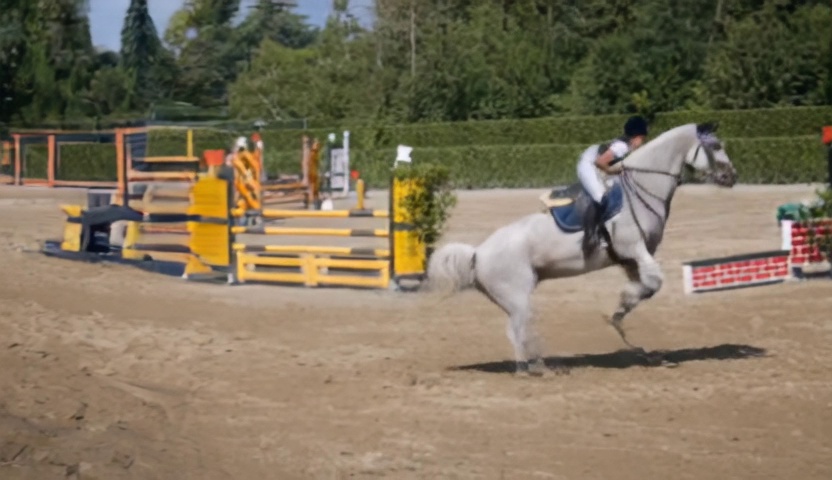}\hspace{0.5pt}  
        \includegraphics[width=0.17\textwidth]{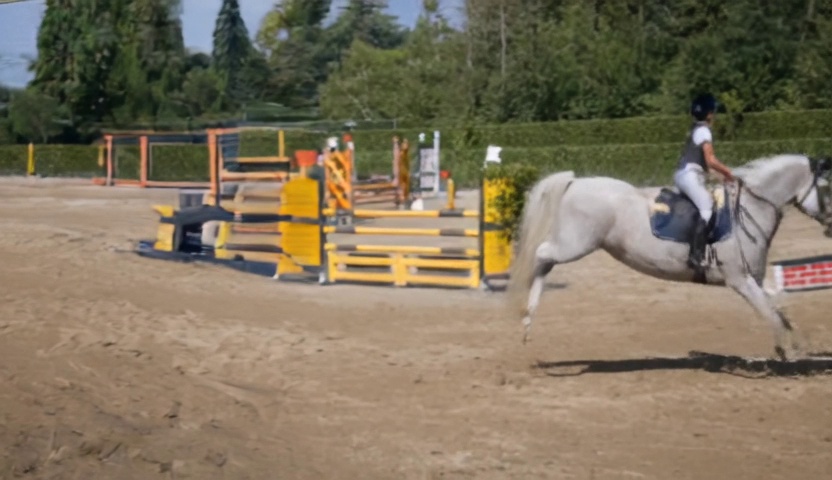}\hspace{0.5pt} 
        \includegraphics[width=0.17\textwidth]{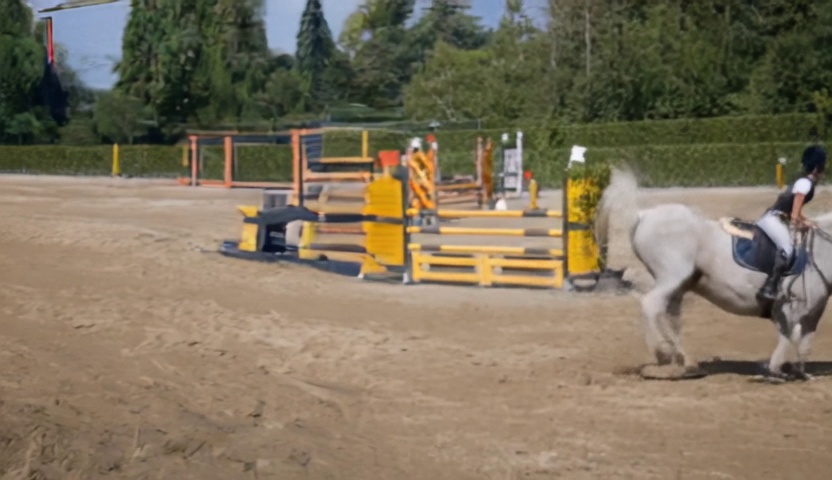}\hspace{0.5pt} 
        \\
        \raisebox{0.cm}{\rotatebox{90}{%
            \sffamily\tiny\parbox{1.1cm}{\centering TrajectorC}%
        }}  &
        \includegraphics[width=0.17\textwidth]{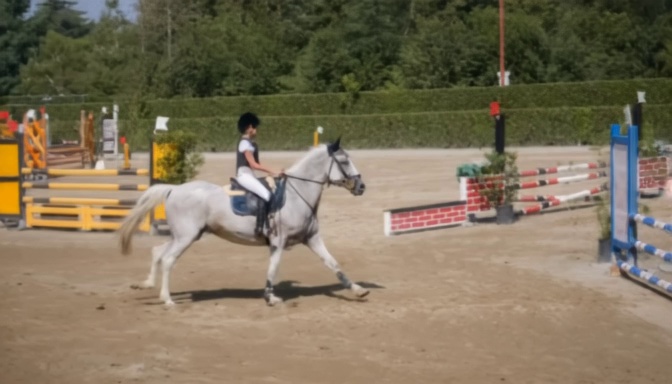}\hspace{0.5pt}  
        \includegraphics[width=0.17\textwidth]{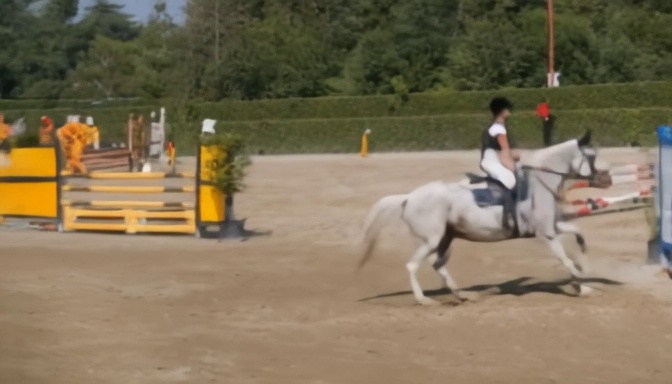}\hspace{0.5pt}  
        \includegraphics[width=0.17\textwidth]{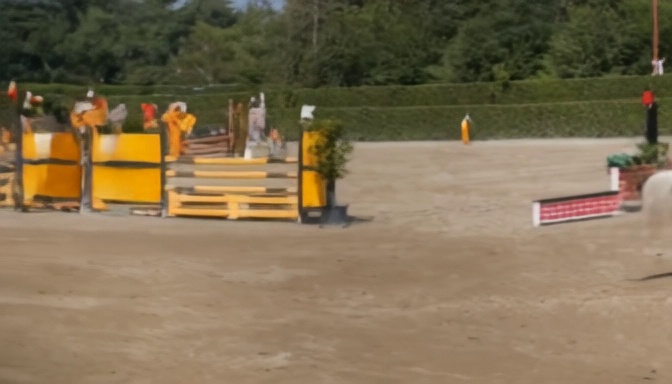}\hspace{0.5pt}  
        \includegraphics[width=0.17\textwidth]{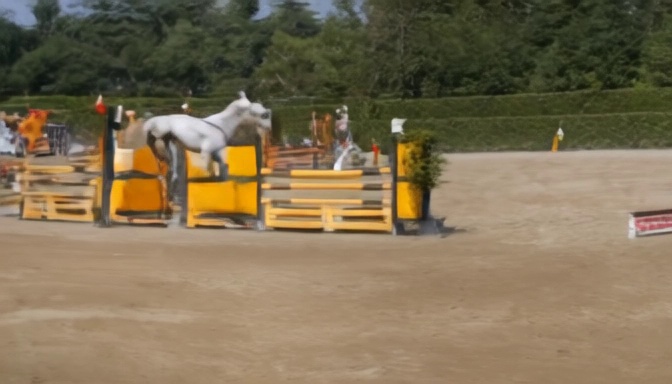}\hspace{0.5pt}     
        \includegraphics[width=0.17\textwidth]{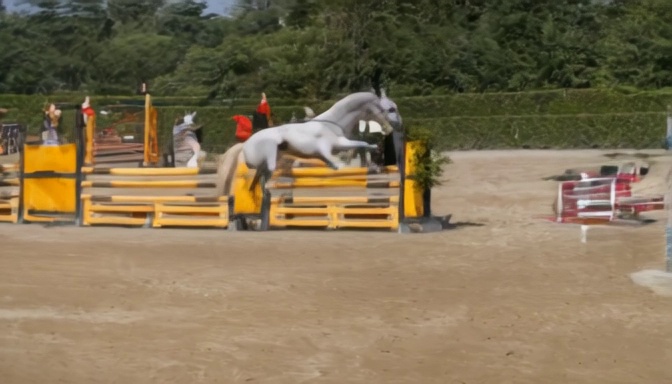}\hspace{0.5pt}   
        \\
        \raisebox{0.cm}{\rotatebox{90}{%
            \sffamily\tiny\parbox{1.1cm}{\centering Ours}%
        }}   &
        \includegraphics[width=0.17\textwidth]{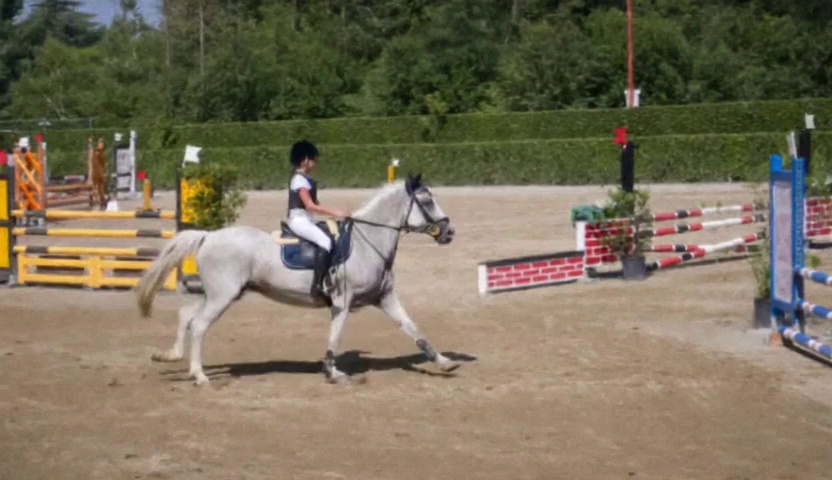}\hspace{0.5pt}  
        \includegraphics[width=0.17\textwidth]{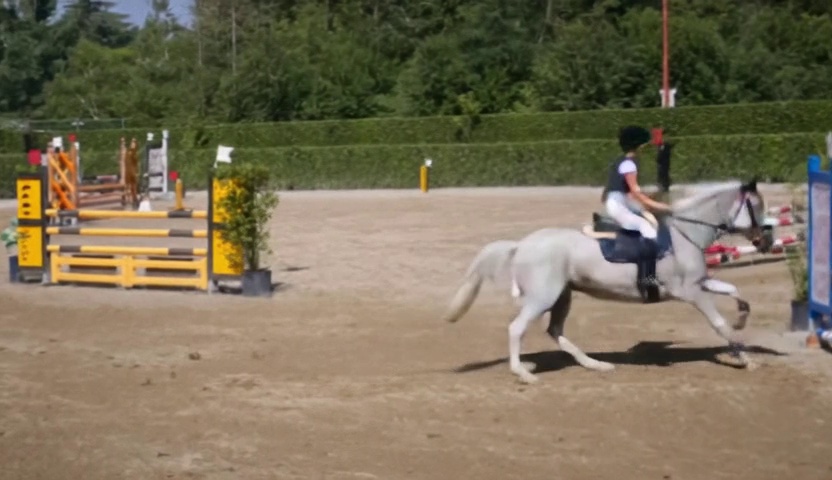}\hspace{0.5pt}  
        \includegraphics[width=0.17\textwidth]{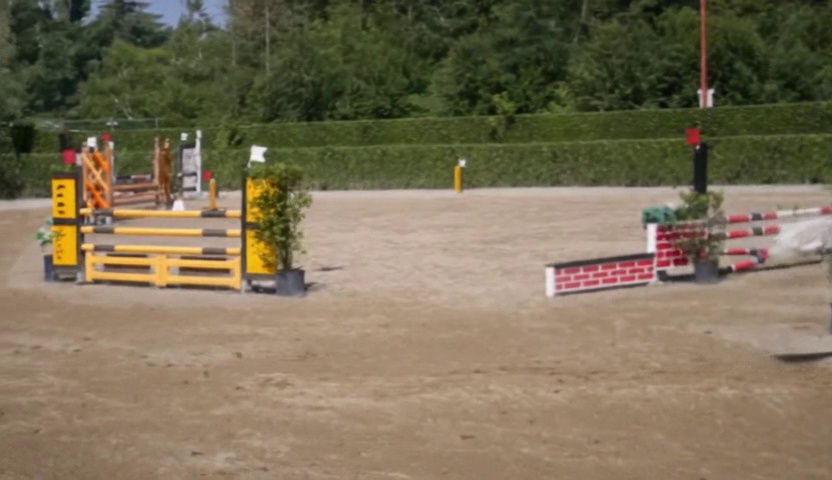}\hspace{0.5pt}  
        \includegraphics[width=0.17\textwidth]{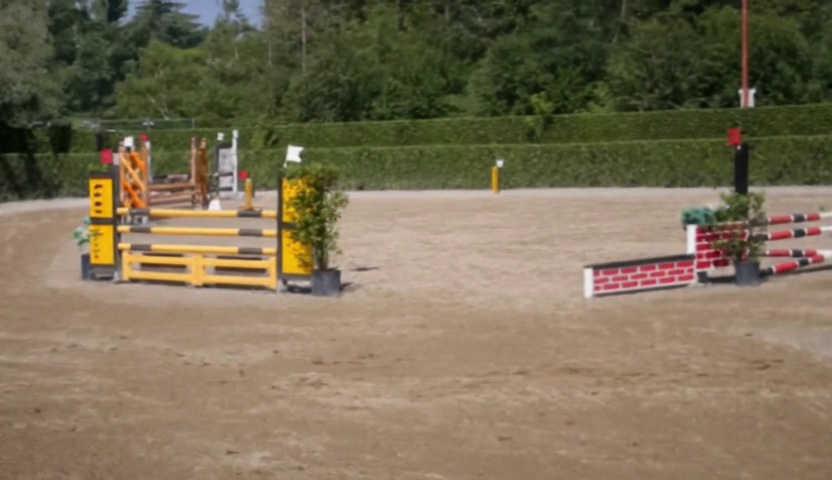}\hspace{0.5pt}   
        \includegraphics[width=0.17\textwidth]{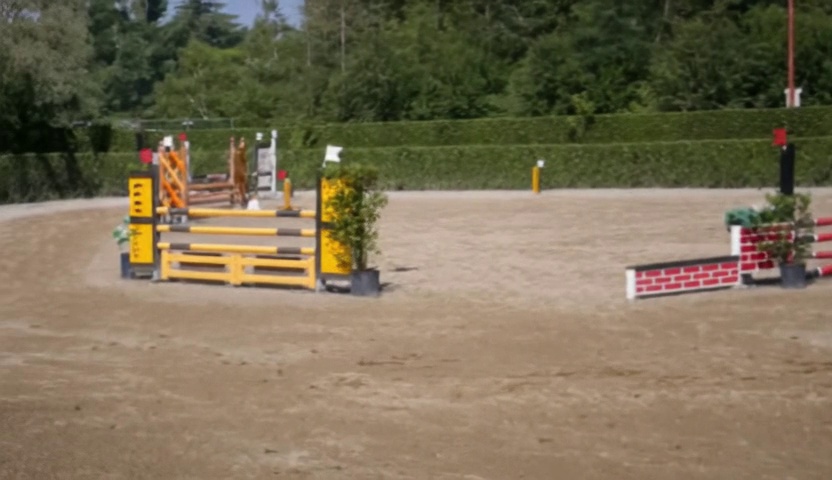}\hspace{0.5pt}  
        \\
        \\
        \raisebox{0.cm}{\rotatebox{90}{%
            \sffamily\tiny\parbox{1.1cm}{\centering ReCamMaster}%
        }}  &
        \includegraphics[width=0.17\textwidth]{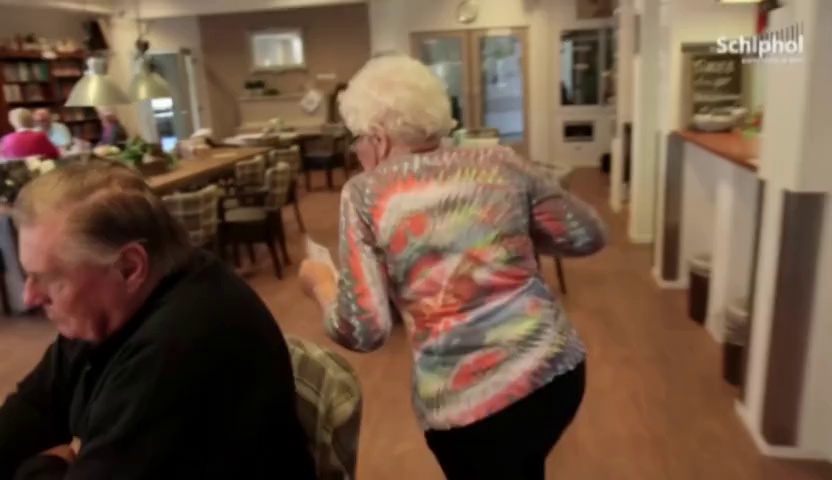}\hspace{0.5pt}  
        \includegraphics[width=0.17\textwidth]{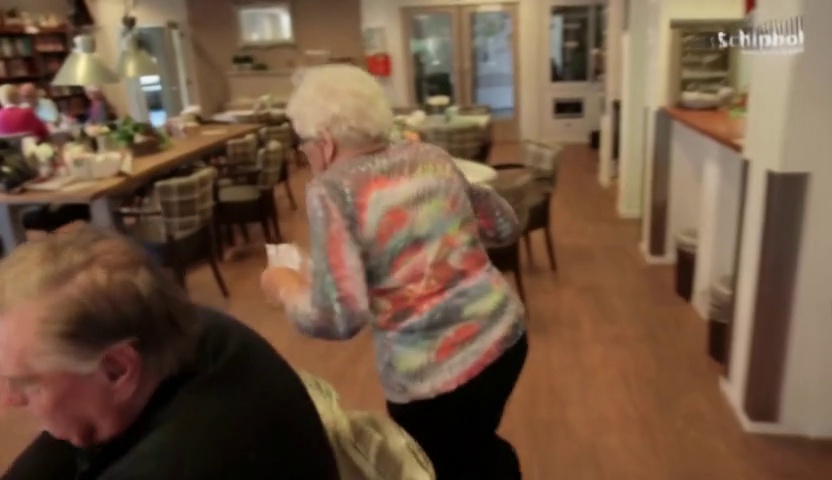}\hspace{0.5pt} 
        \includegraphics[width=0.17\textwidth]{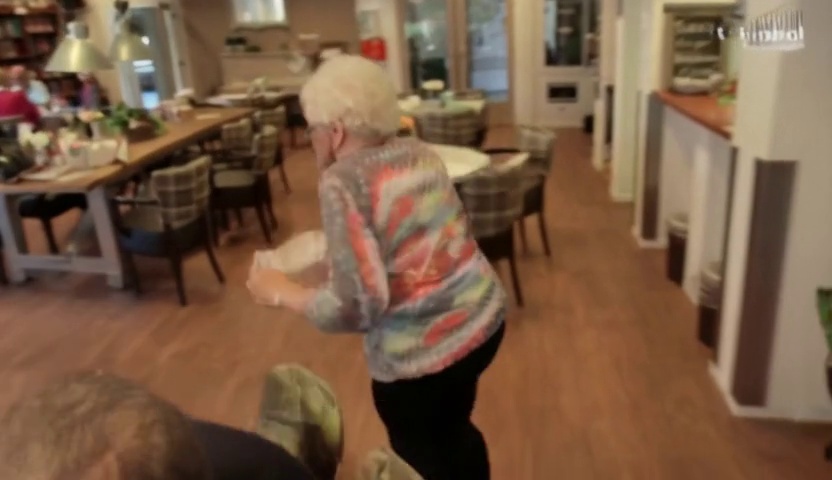}\hspace{0.5pt}   
        \includegraphics[width=0.17\textwidth]{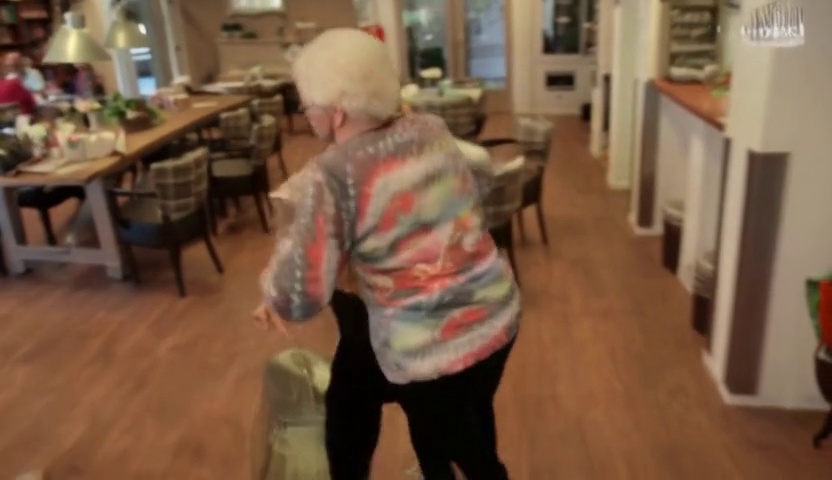}\hspace{0.5pt}  
        \includegraphics[width=0.17\textwidth]{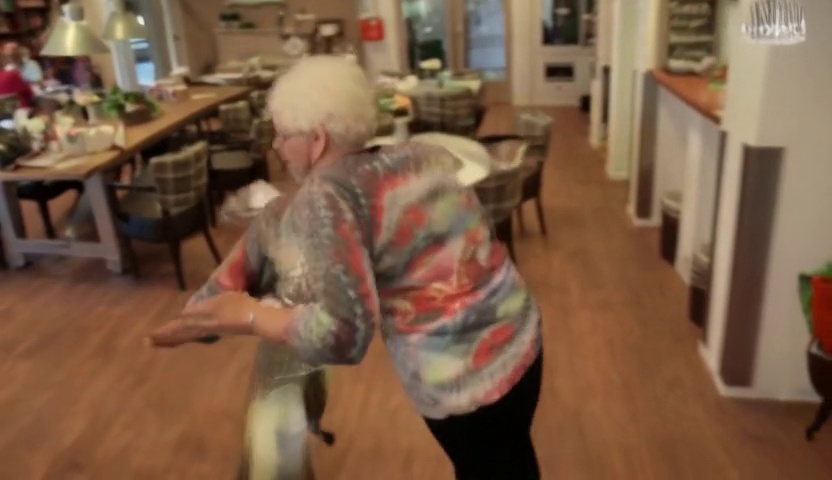}\hspace{0.5pt}  
        \\
        \raisebox{0.cm}{\rotatebox{90}{%
            \sffamily\tiny\parbox{1.1cm}{\centering ReDirector}%
        }}  &
        \includegraphics[width=0.17\textwidth]{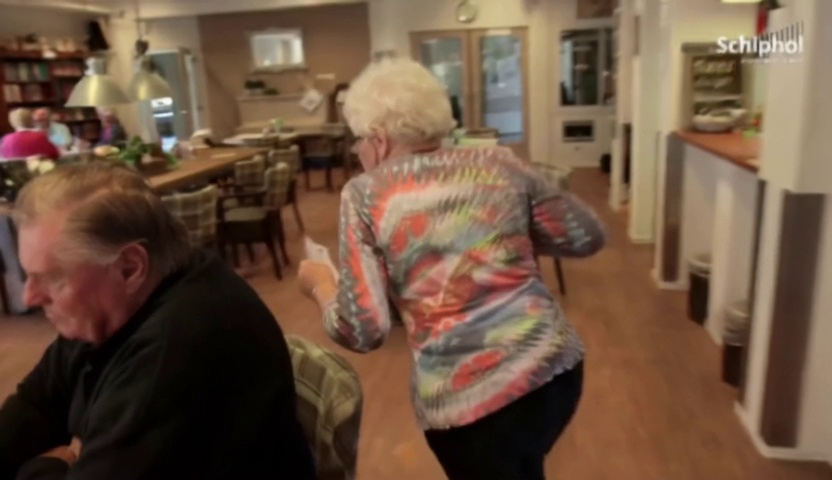}\hspace{0.5pt}  
        \includegraphics[width=0.17\textwidth]{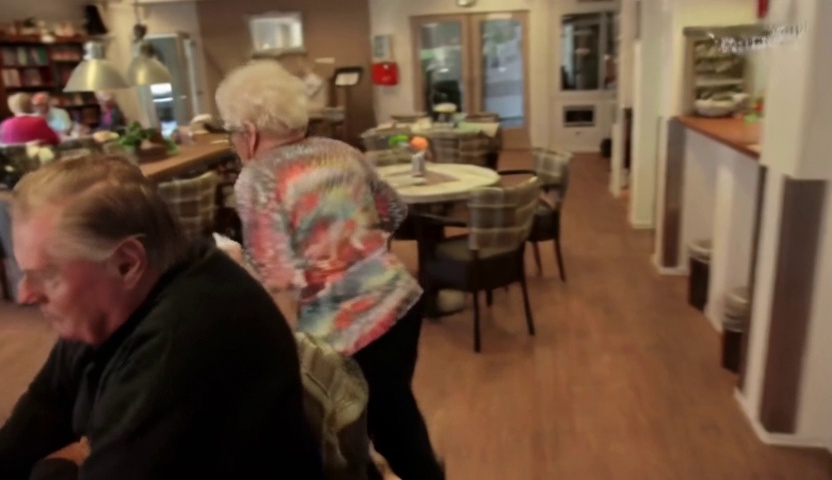}\hspace{0.5pt}  
        \includegraphics[width=0.17\textwidth]{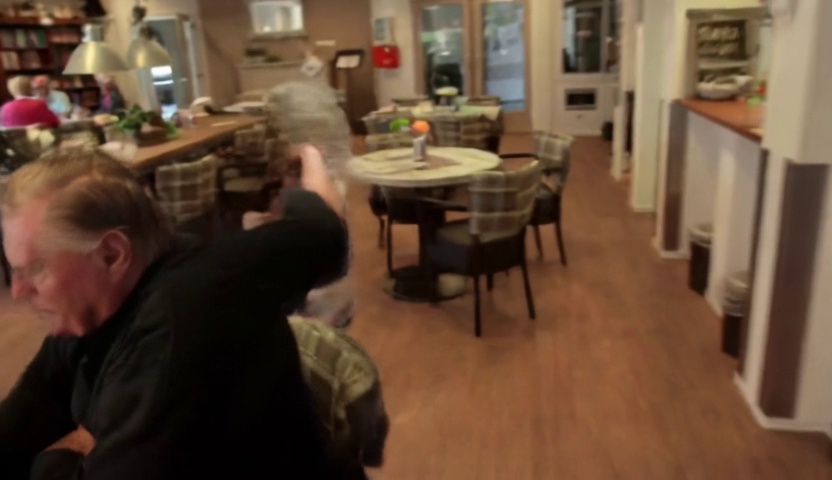}\hspace{0.5pt}  
        \includegraphics[width=0.17\textwidth]{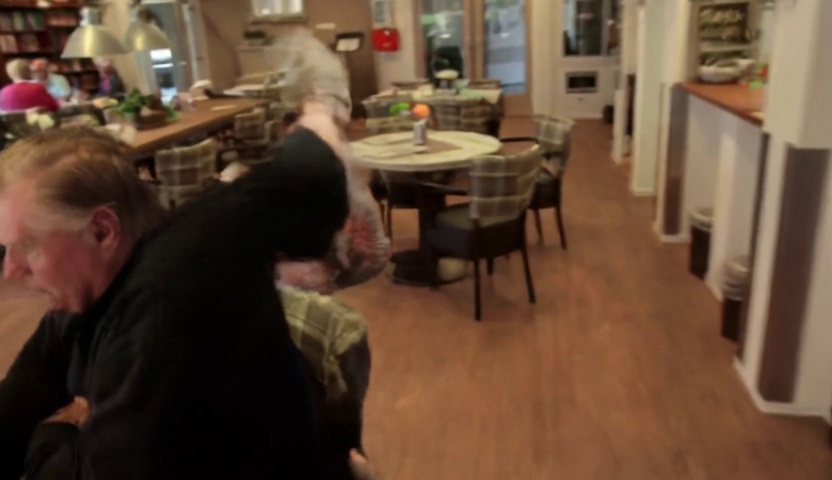}\hspace{0.5pt}  
        \includegraphics[width=0.17\textwidth]{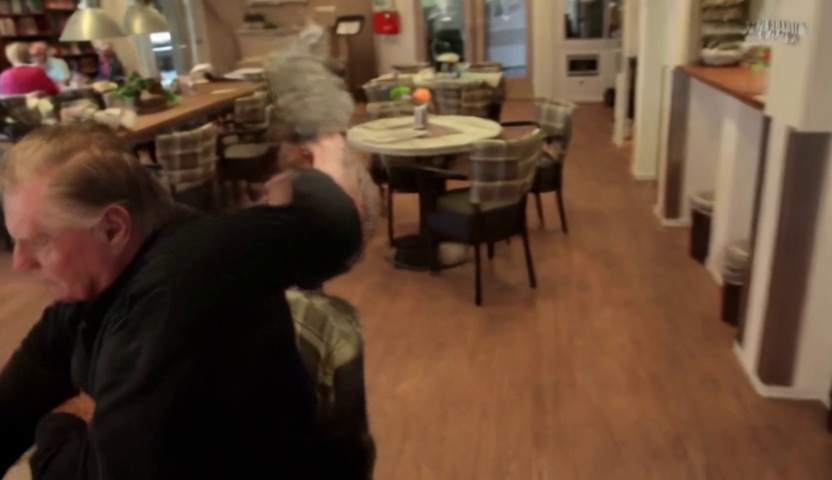}\hspace{0.5pt}  
        \\
        \raisebox{0.cm}{\rotatebox{90}{%
            \sffamily\tiny\parbox{1.1cm}{\centering TrajectorC}%
        }}   &
        \includegraphics[width=0.17\textwidth]{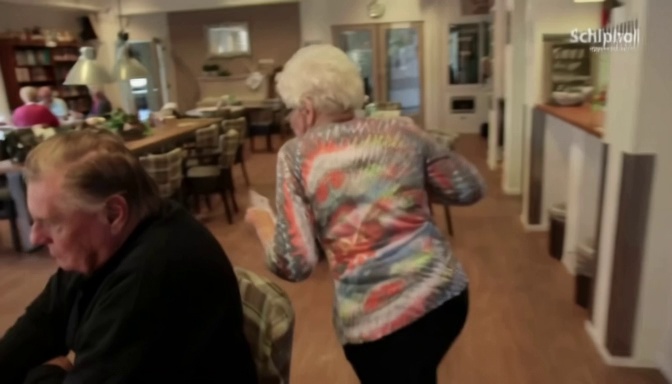}\hspace{0.5pt}\hspace{0.5pt}   
        \includegraphics[width=0.17\textwidth]{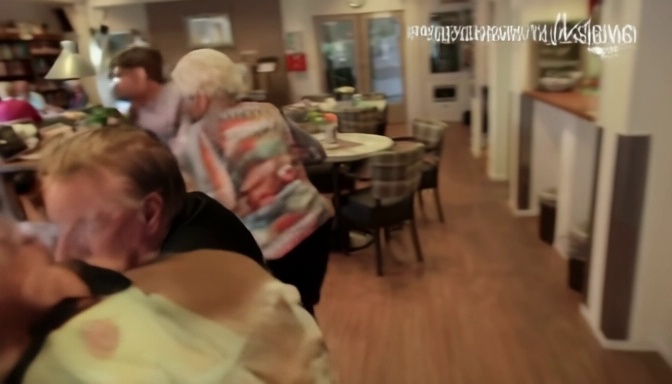}\hspace{0.5pt}  
        \includegraphics[width=0.17\textwidth]{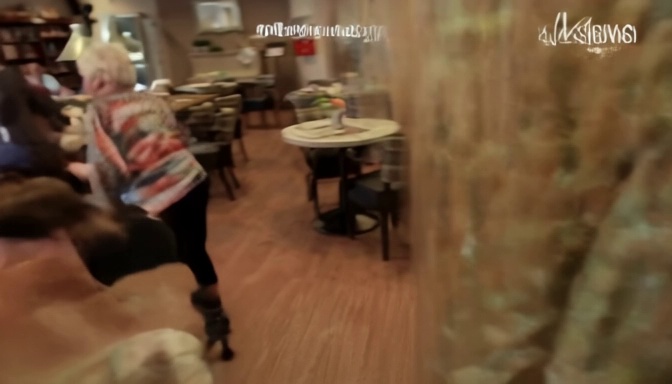}\hspace{0.5pt}  
        \includegraphics[width=0.17\textwidth]{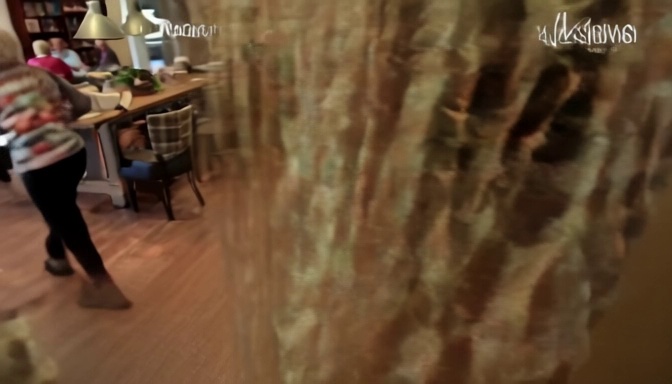}\hspace{0.5pt}  
        \includegraphics[width=0.17\textwidth]{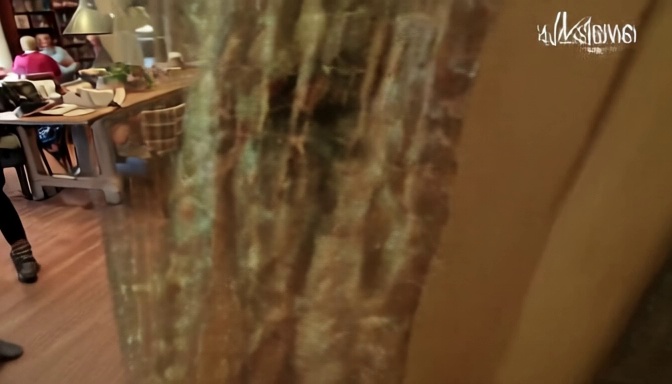}\hspace{0.5pt}     
        \\
        \raisebox{0.cm}{\rotatebox{90}{%
            \sffamily\tiny\parbox{1.1cm}{\centering Ours}%
        }} &
        \includegraphics[width=0.17\textwidth]{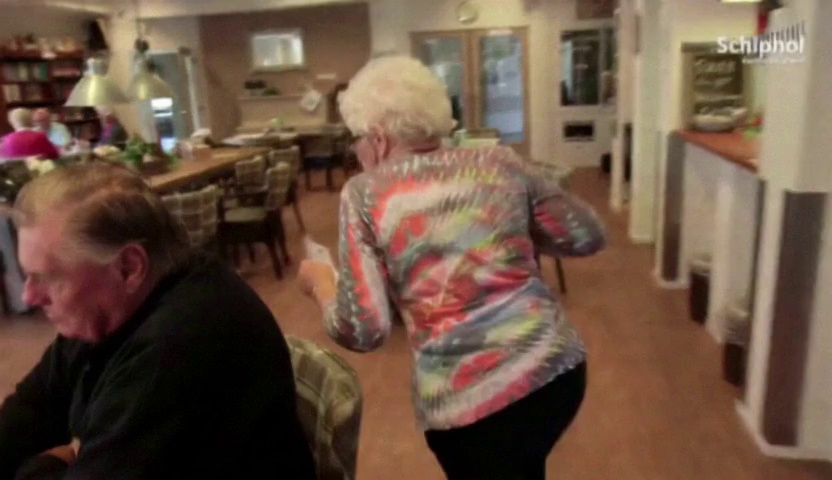}\hspace{0.5pt}  
        \includegraphics[width=0.17\textwidth]{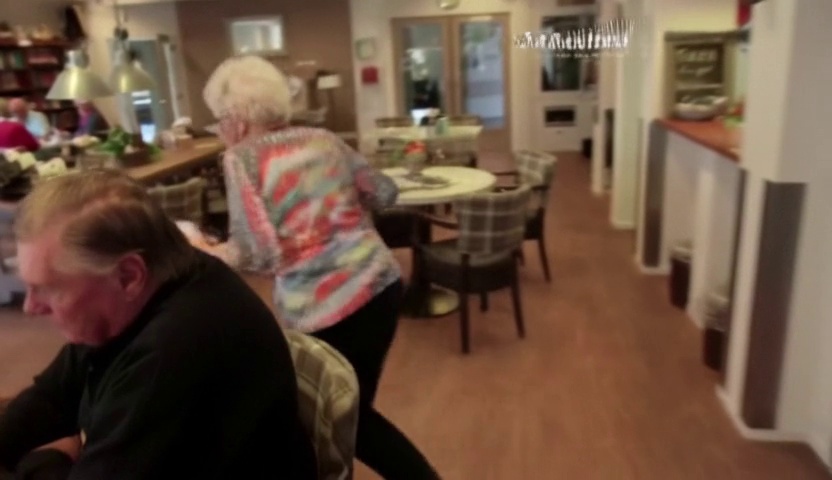}\hspace{0.5pt}  
        \includegraphics[width=0.17\textwidth]{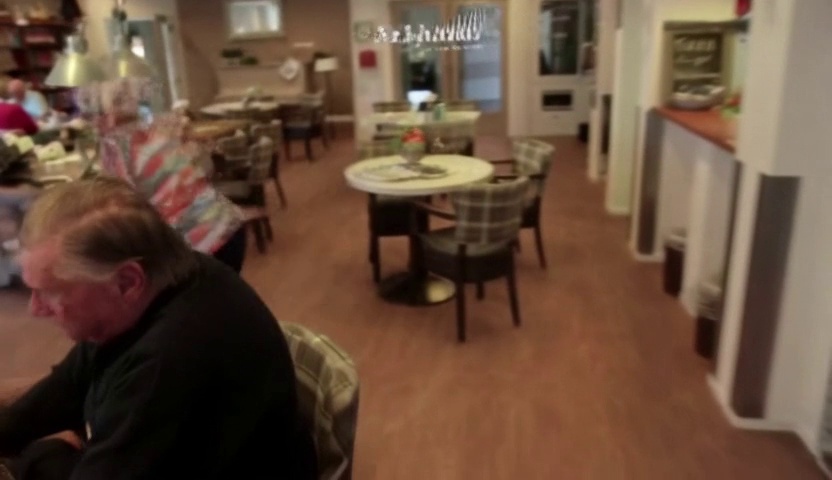}\hspace{0.5pt}  
        \includegraphics[width=0.17\textwidth]{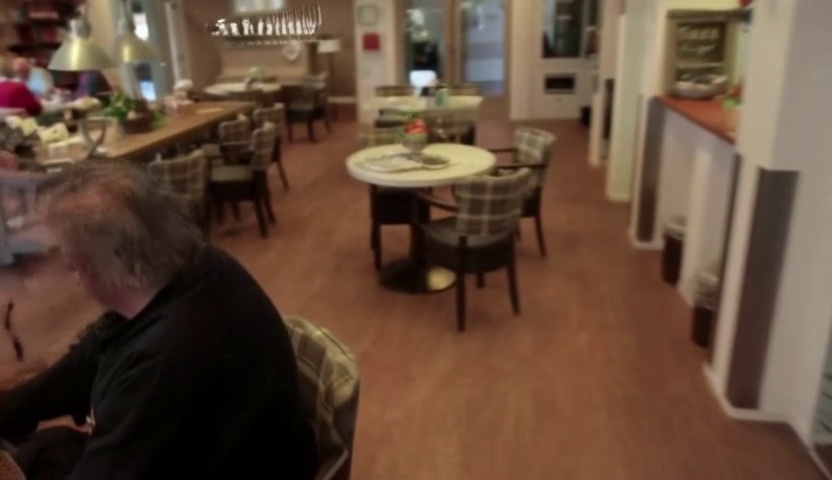}\hspace{0.5pt}  
        \includegraphics[width=0.17\textwidth]{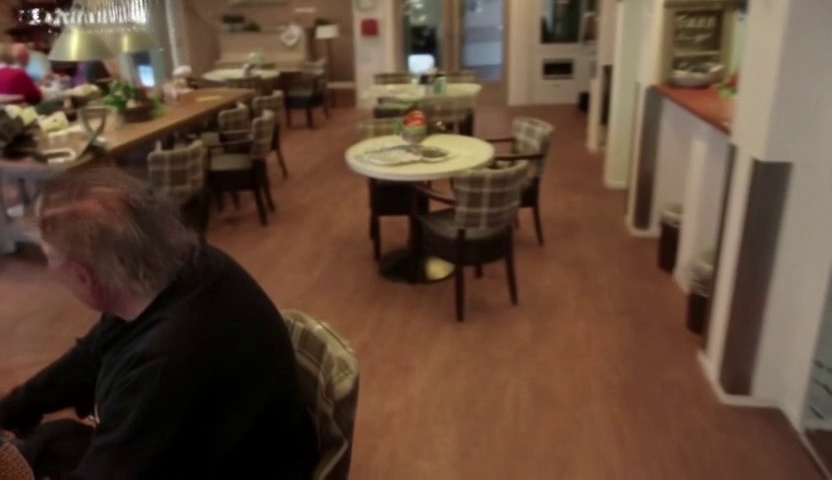}\hspace{0.5pt}   
   \end{tabular}
    \caption{\textbf{Additional qualitative comparison on the DAVIS dataset}.}
    \vspace{-0.1cm}
    \label{fig:app_add_qual_1}
\end{figure}

\begin{figure*}[htbp]
    \centering
    \begin{tabular}{@{}c@{\hspace{2pt}}c@{\hspace{2pt}}c@{\hspace{2pt}}}
        \raisebox{0.cm}{\rotatebox{90}{%
            \sffamily\tiny\parbox{1.1cm}{\centering ReCamMaster}%
        }}   &

        \includegraphics[width=0.17\textwidth]{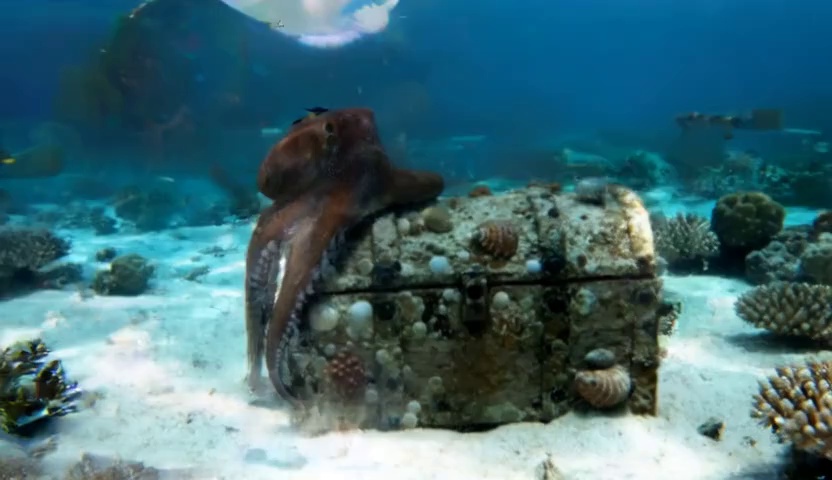}\hspace{0.5pt}  
        \includegraphics[width=0.17\textwidth]{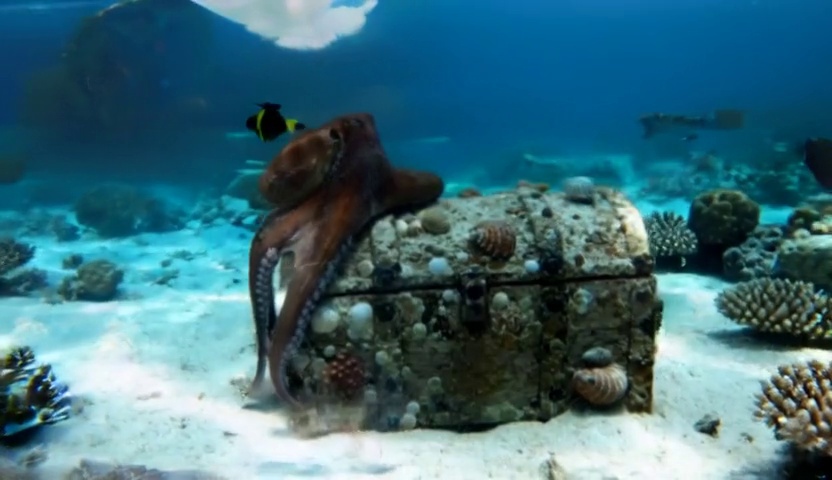}\hspace{0.5pt}  
        \includegraphics[width=0.17\textwidth]{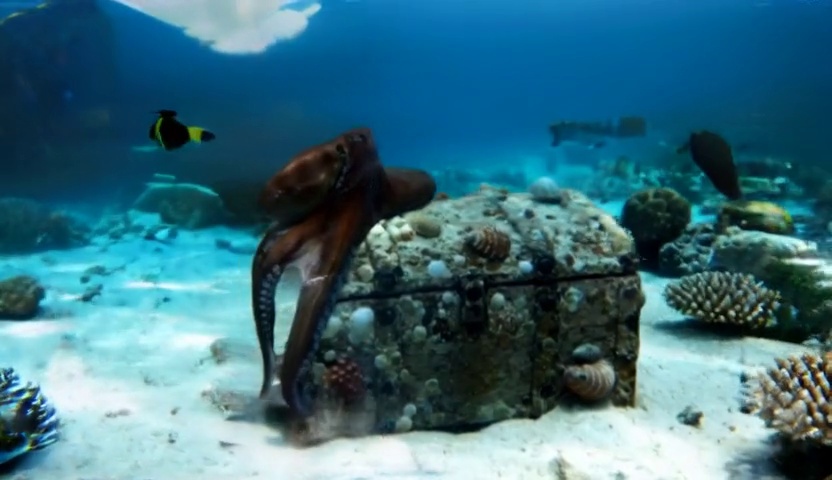}\hspace{0.5pt}  
        \includegraphics[width=0.17\textwidth]{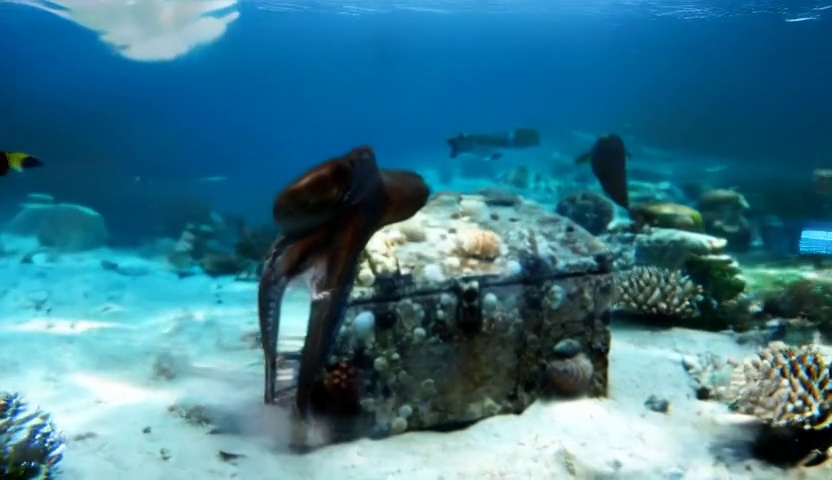}\hspace{0.5pt}  
        \includegraphics[width=0.17\textwidth]{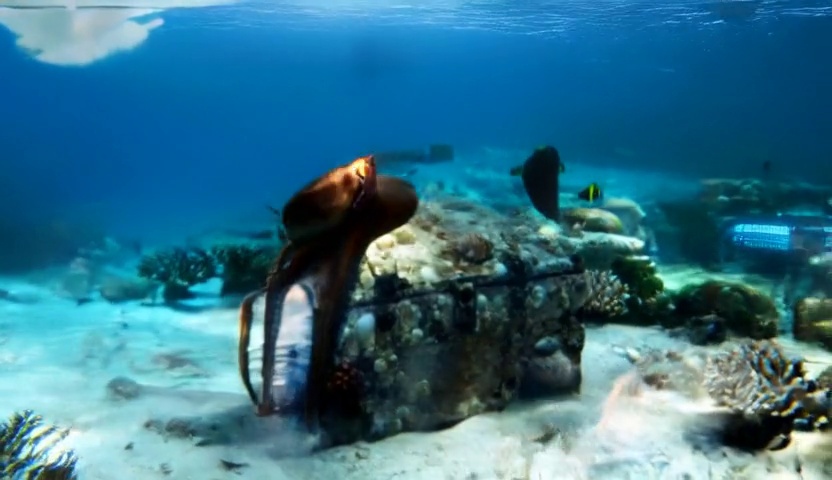}\hspace{0.5pt}  
        \\
        \raisebox{0.cm}{\rotatebox{90}{ 
            \sffamily\tiny\parbox{1.1cm}{\centering ReDirector} }}  &
        \includegraphics[width=0.17\textwidth]{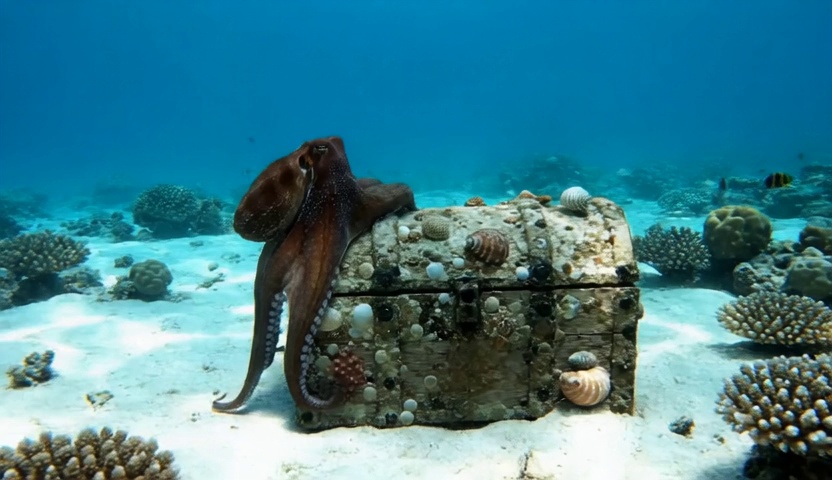}\hspace{0.5pt}  
        \includegraphics[width=0.17\textwidth]{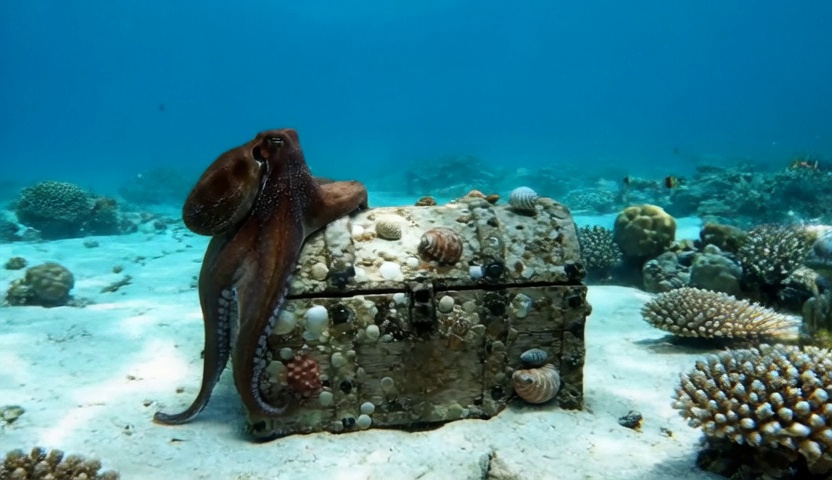}\hspace{0.5pt}  
        \includegraphics[width=0.17\textwidth]{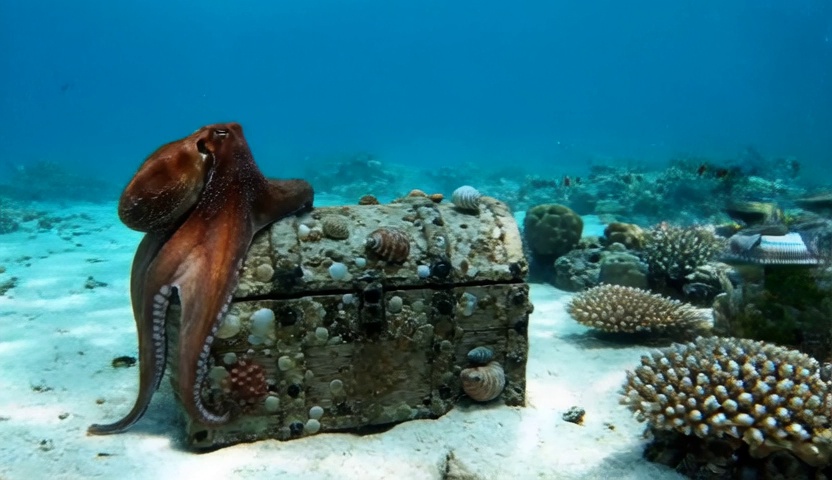}\hspace{0.5pt}  
        \includegraphics[width=0.17\textwidth]{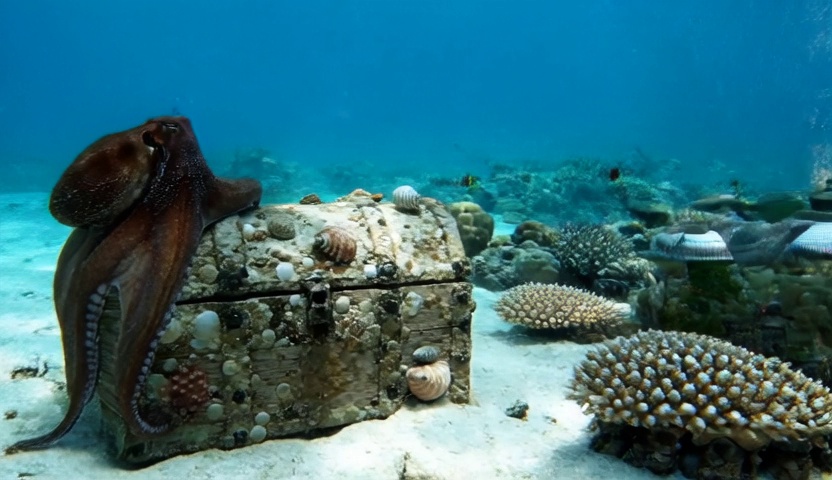}\hspace{0.5pt}  
        \includegraphics[width=0.17\textwidth]{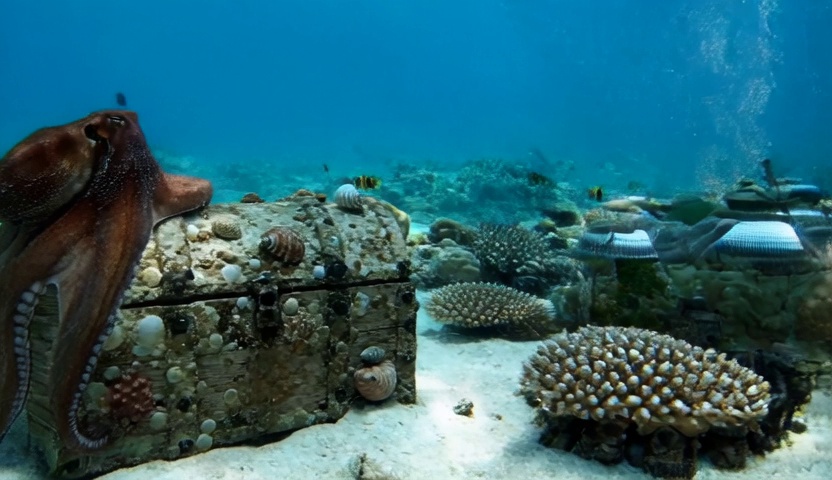}\hspace{0.5pt}  
        \\
        \raisebox{0.cm}{\rotatebox{90}{ 
            \sffamily\tiny\parbox{1.1cm}{\centering TrajectorC} }}  &
        \includegraphics[width=0.17\textwidth]{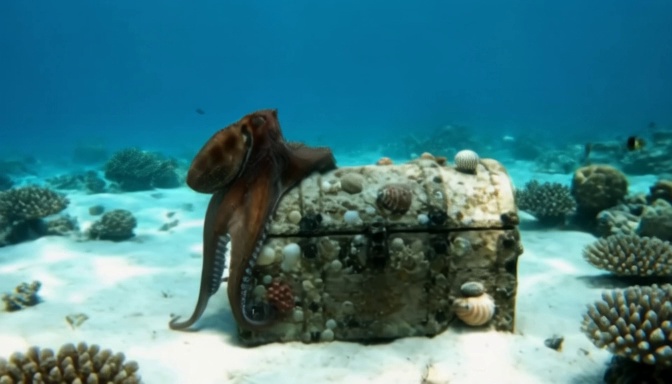}\hspace{0.5pt}  
        \includegraphics[width=0.17\textwidth]{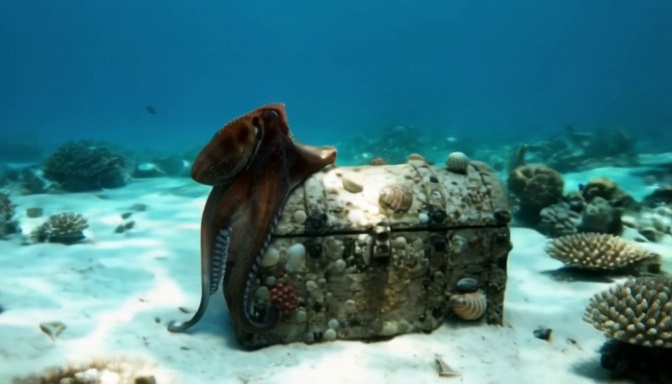}\hspace{0.5pt}  
        \includegraphics[width=0.17\textwidth]{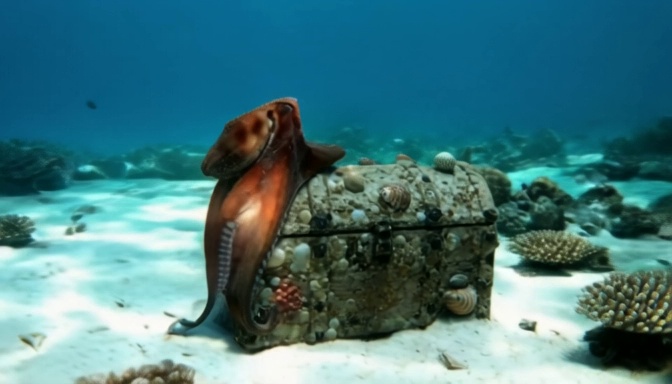}\hspace{0.5pt}  
        \includegraphics[width=0.17\textwidth]{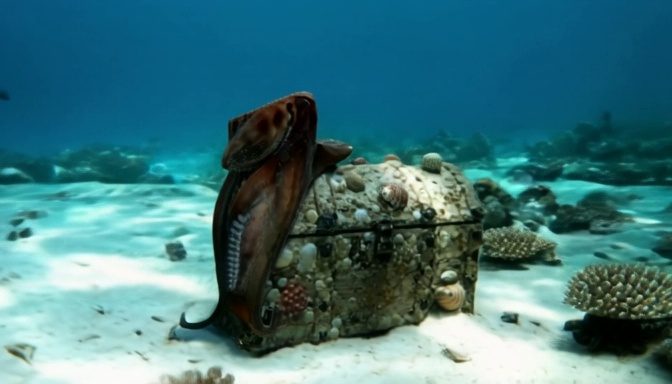}\hspace{0.5pt}    
        \includegraphics[width=0.17\textwidth]{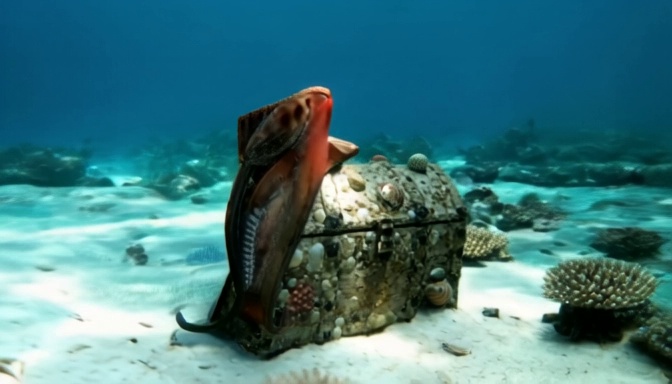}\hspace{0.5pt}   
        \\
        \raisebox{0.cm}{\rotatebox{90}{
            \sffamily\tiny\parbox{1.1cm}{\centering Ours} }}  &
        \includegraphics[width=0.17\textwidth]{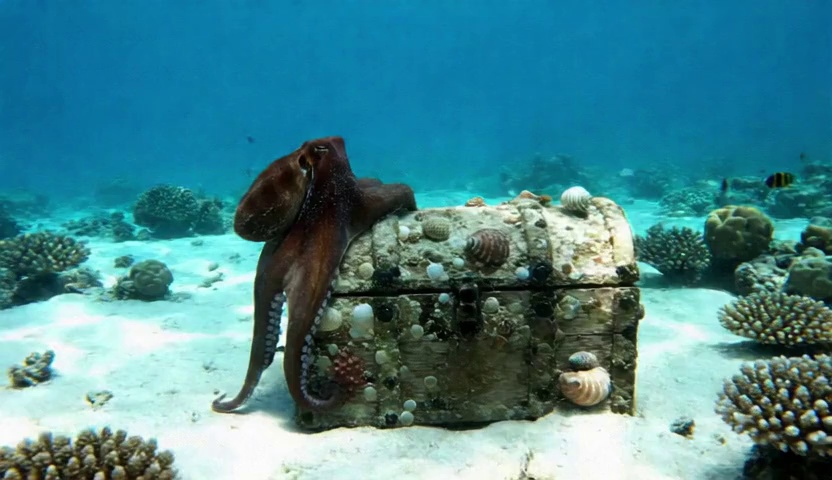}\hspace{0.5pt}  
        \includegraphics[width=0.17\textwidth]{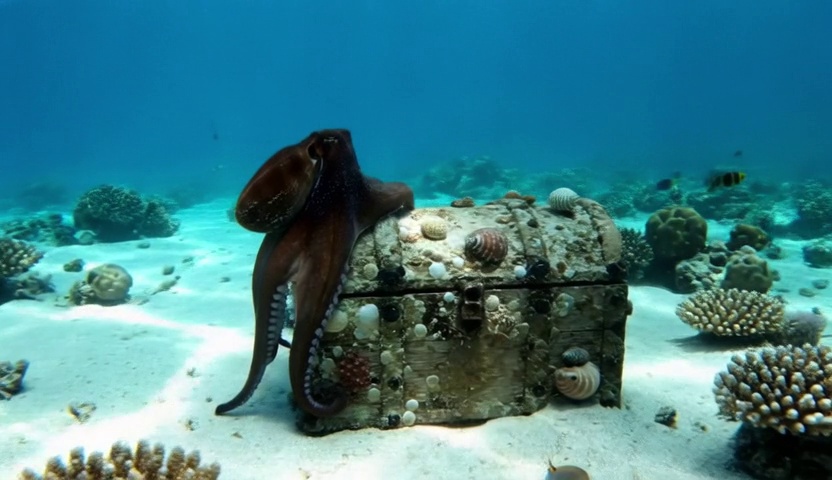}\hspace{0.5pt}  
        \includegraphics[width=0.17\textwidth]{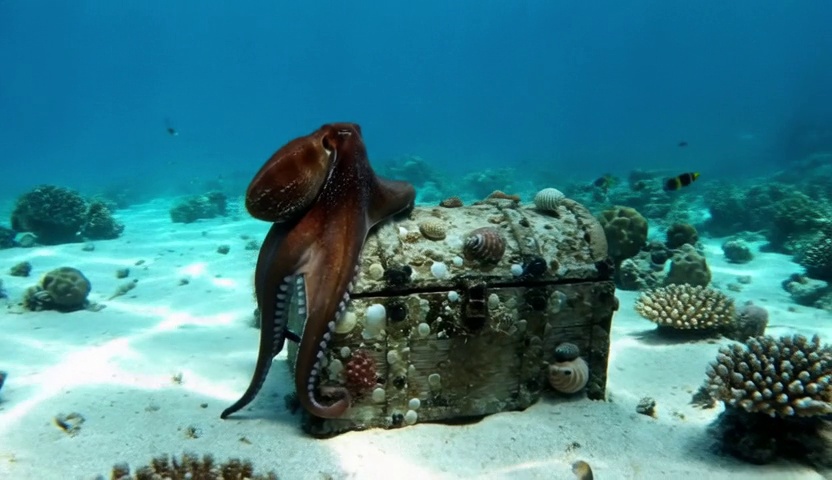}\hspace{0.5pt}  
        \includegraphics[width=0.17\textwidth]{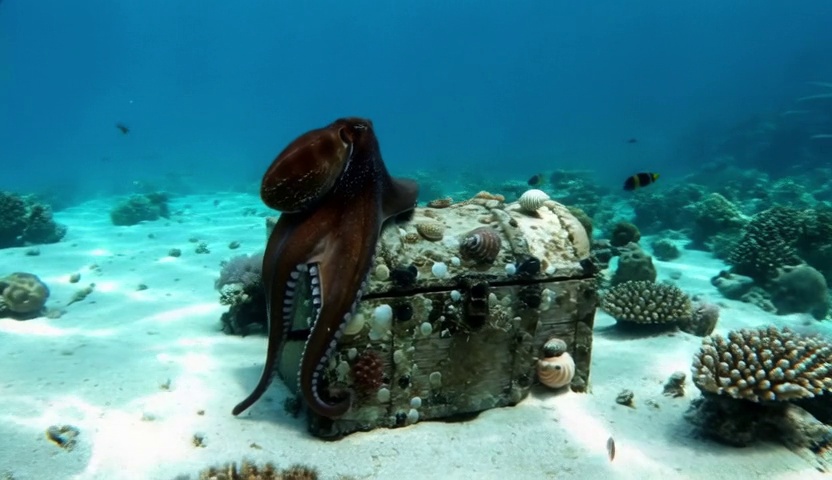}\hspace{0.5pt}  
        \includegraphics[width=0.17\textwidth]{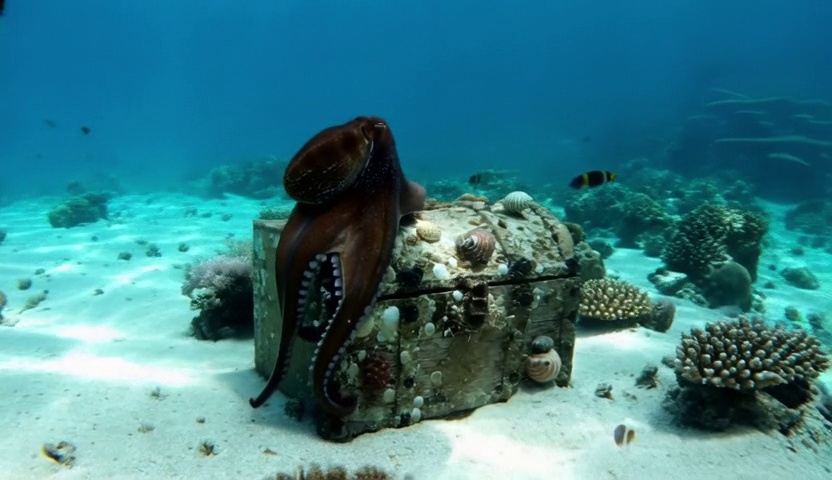}\hspace{0.5pt}  
        \\
        \\
        \raisebox{0.cm}{\rotatebox{90}{ 
            \sffamily\tiny\parbox{1.1cm}{\centering ReCamMaster} }}  &
        \includegraphics[width=0.17\textwidth]{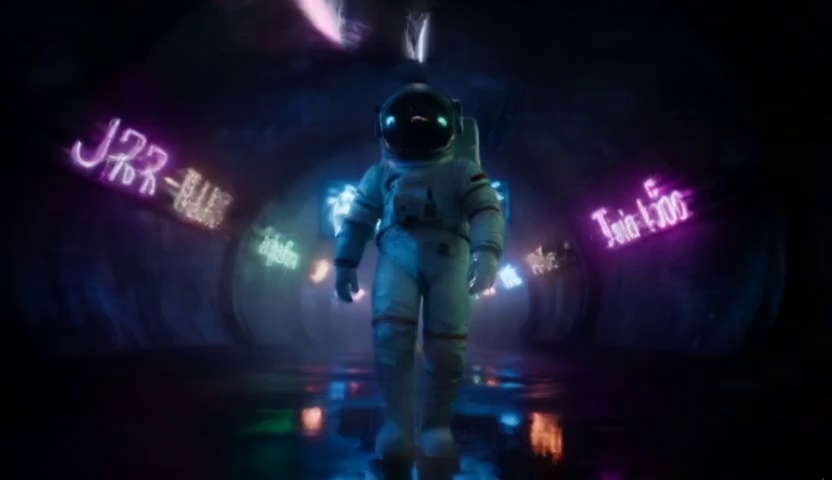}\hspace{0.5pt}  
        \includegraphics[width=0.17\textwidth]{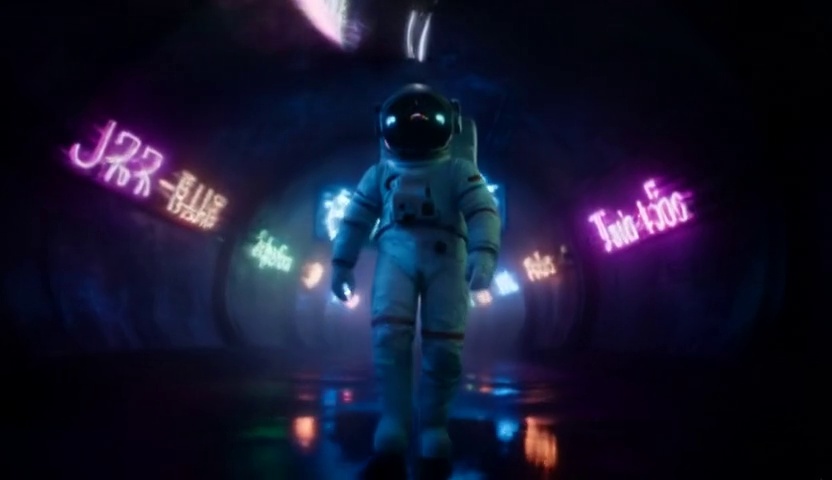}\hspace{0.5pt}  
        \includegraphics[width=0.17\textwidth]{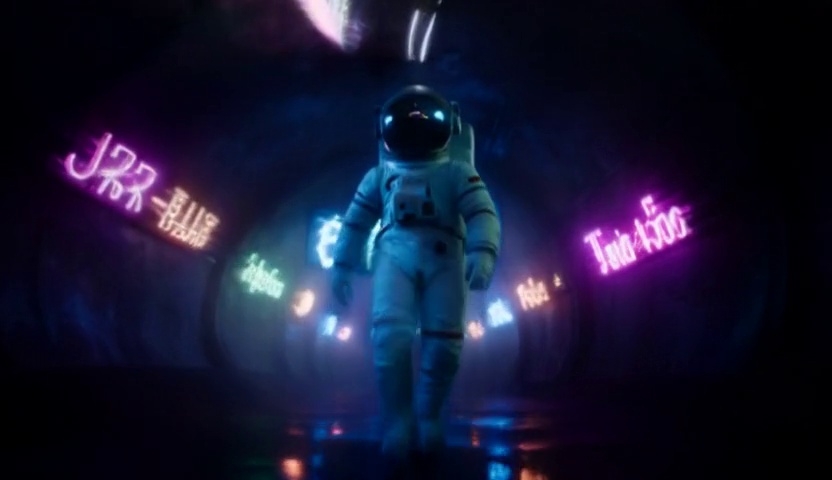}\hspace{0.5pt}  
        \includegraphics[width=0.17\textwidth]{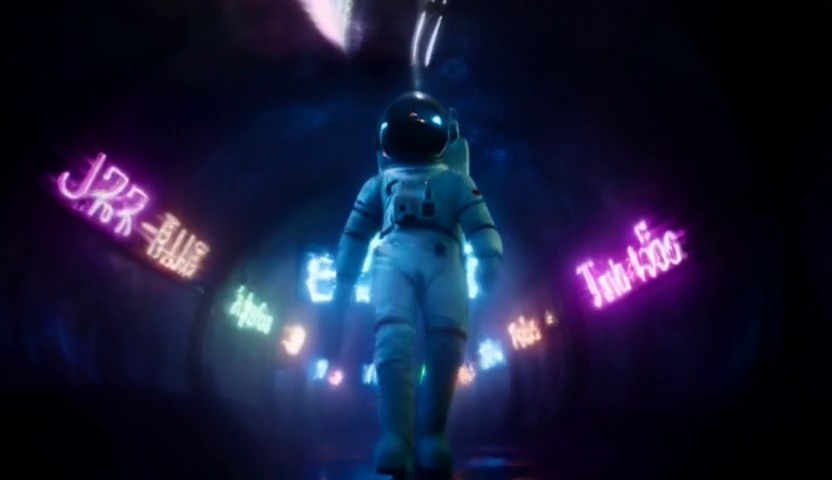}\hspace{0.5pt}   
        \includegraphics[width=0.17\textwidth]{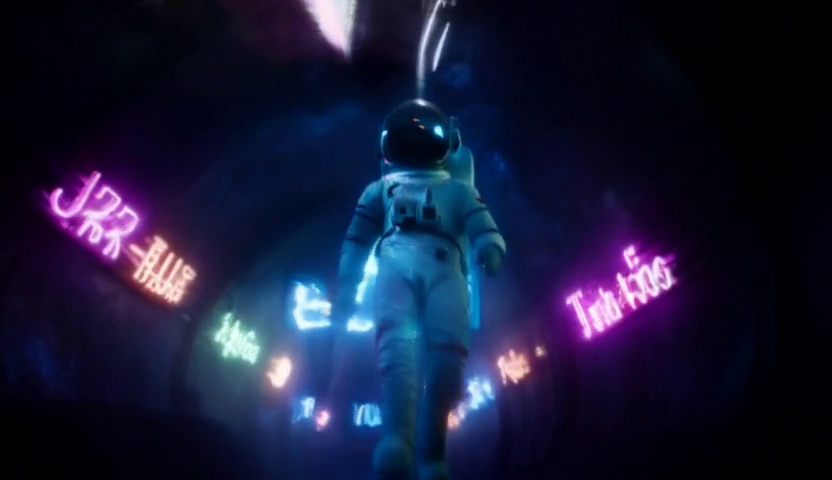}\hspace{0.5pt}   
        \\
        \raisebox{0.cm}{\rotatebox{90}{ 
            \sffamily\tiny\parbox{1.1cm}{\centering ReDirector} }}  &
        \includegraphics[width=0.17\textwidth]{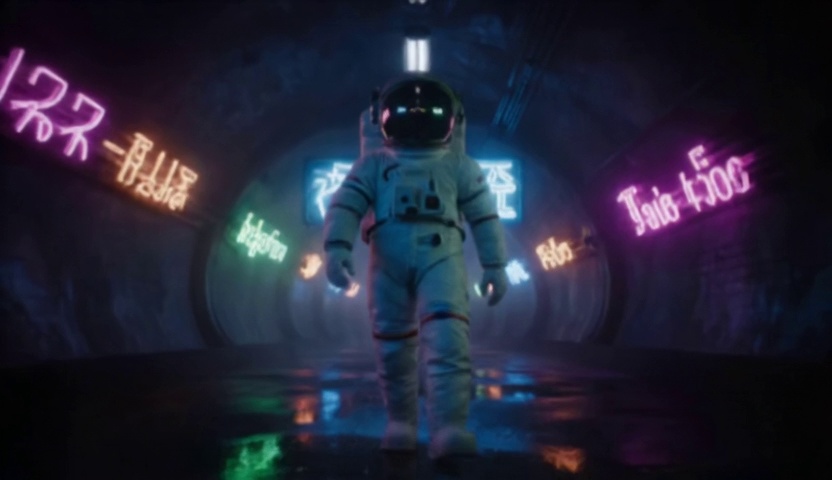}\hspace{0.5pt}  
        \includegraphics[width=0.17\textwidth]{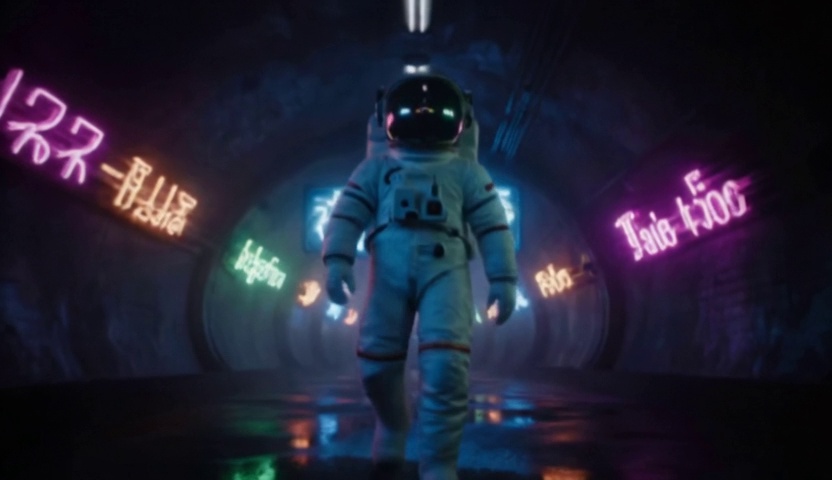}\hspace{0.5pt}  
        \includegraphics[width=0.17\textwidth]{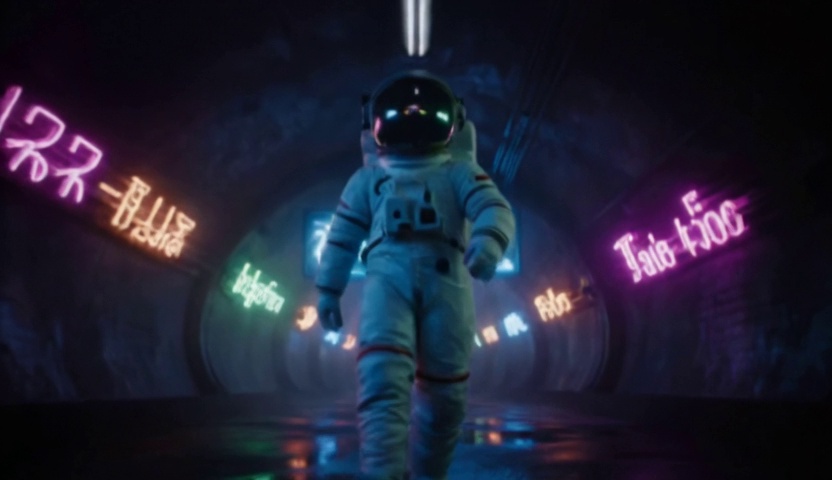}\hspace{0.5pt}  
        \includegraphics[width=0.17\textwidth]{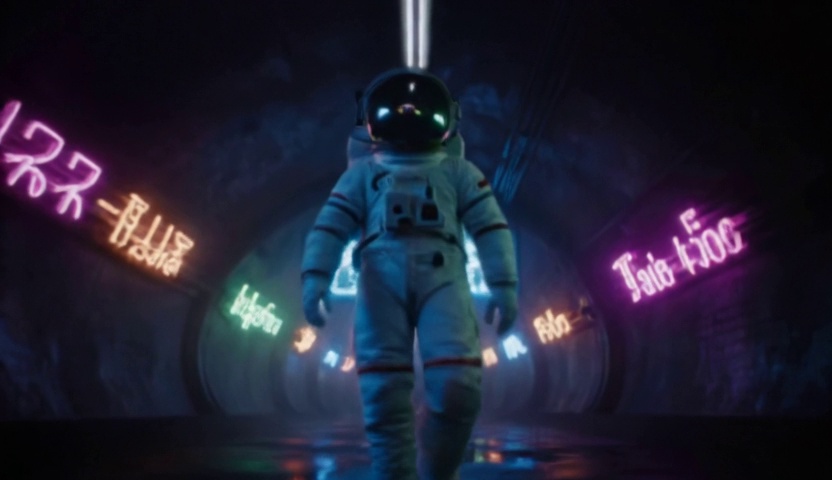}\hspace{0.5pt}  
        \includegraphics[width=0.17\textwidth]{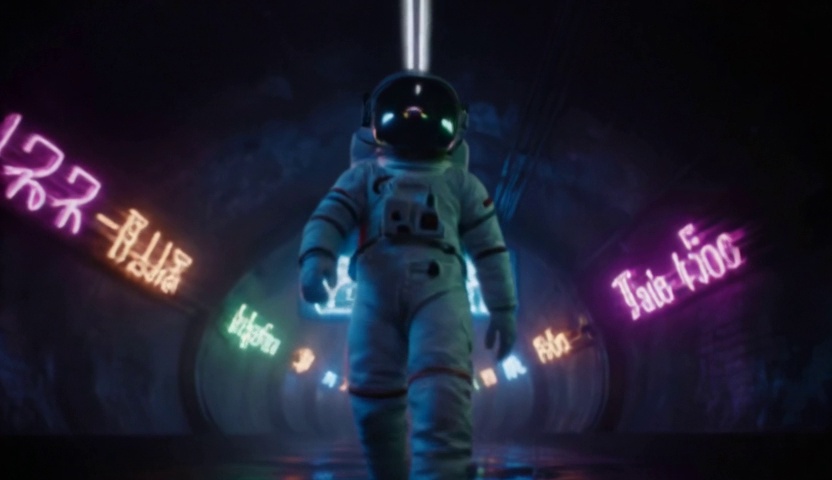}\hspace{0.5pt}  
        \\
        \raisebox{0.cm}{\rotatebox{90}{ 
            \sffamily\tiny\parbox{1.1cm}{\centering TrajectorC } }}  &
        \includegraphics[width=0.17\textwidth]{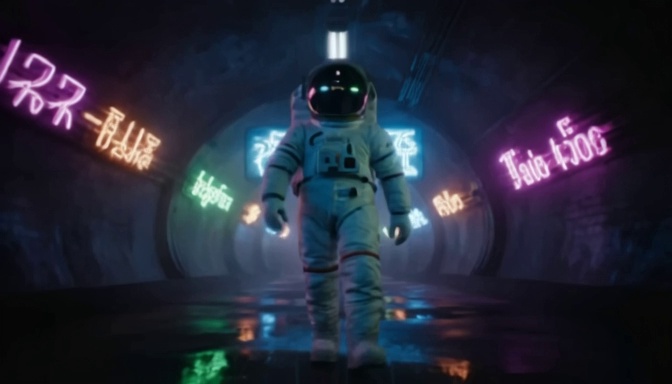}\hspace{0.5pt}  
        \includegraphics[width=0.17\textwidth]{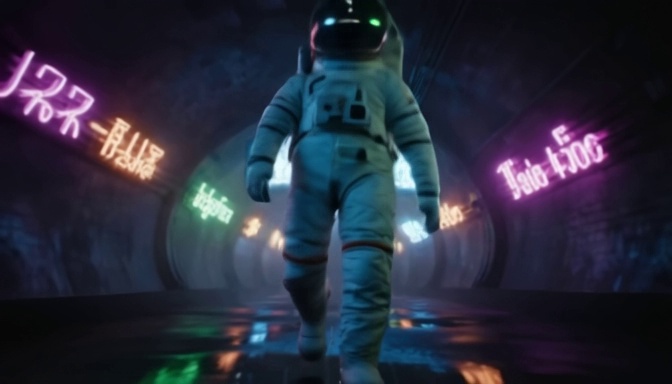}\hspace{0.5pt}  
        \includegraphics[width=0.17\textwidth]{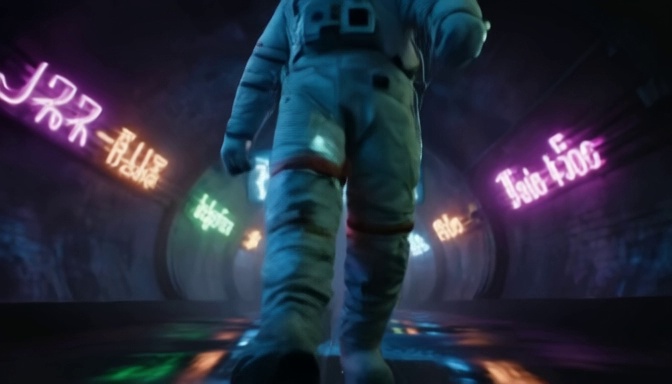}\hspace{0.5pt}  
        \includegraphics[width=0.17\textwidth]{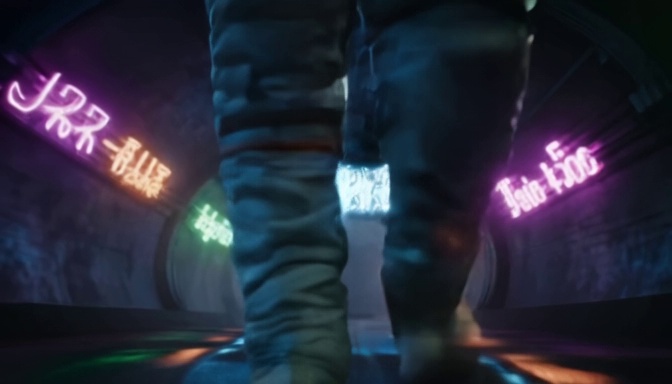}\hspace{0.5pt}     
        \includegraphics[width=0.17\textwidth]{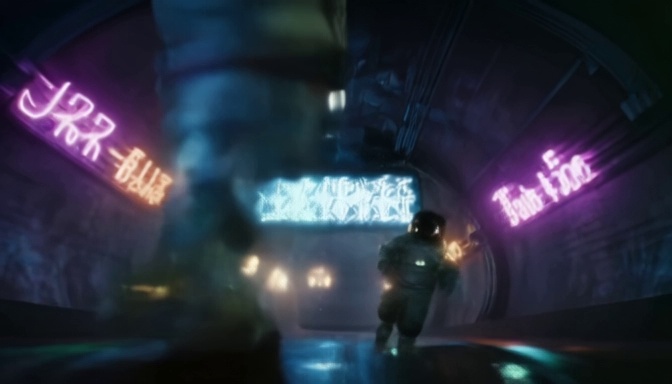}\hspace{0.5pt}   
        \\
        \raisebox{0.cm}{\rotatebox{90}{
            \sffamily\tiny\parbox{1.1cm}{\centering Ours} }}  &
        \includegraphics[width=0.17\textwidth]{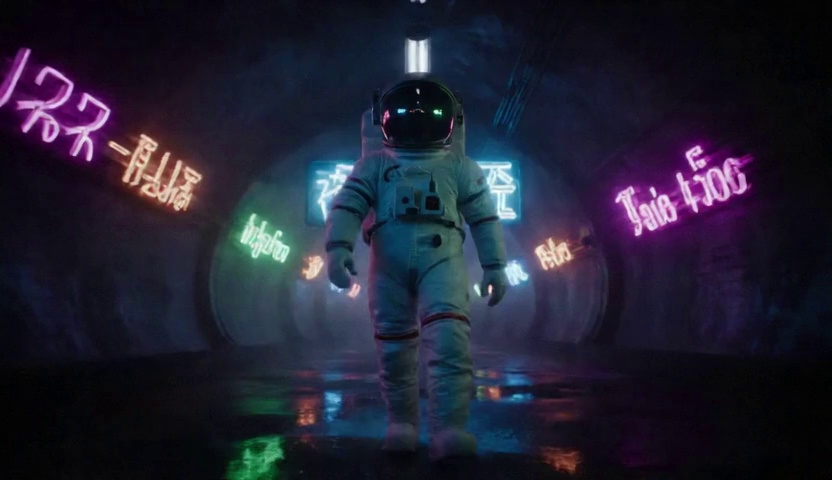}\hspace{0.5pt}  
        \includegraphics[width=0.17\textwidth]{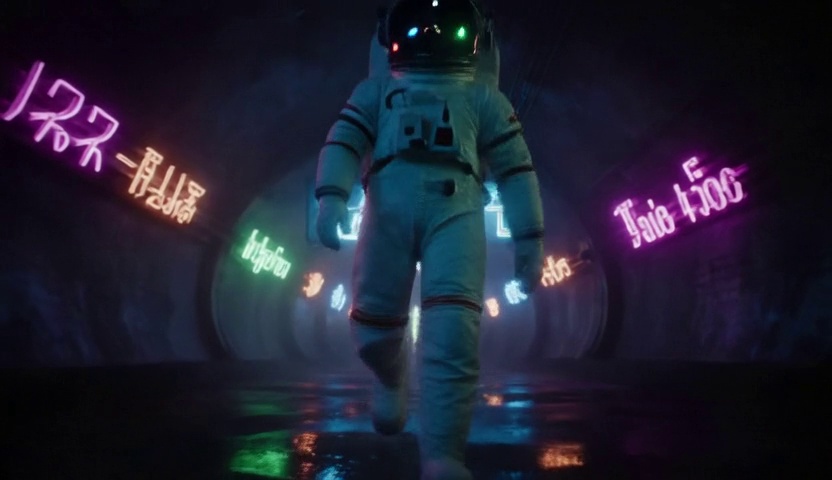}\hspace{0.5pt}  
        \includegraphics[width=0.17\textwidth]{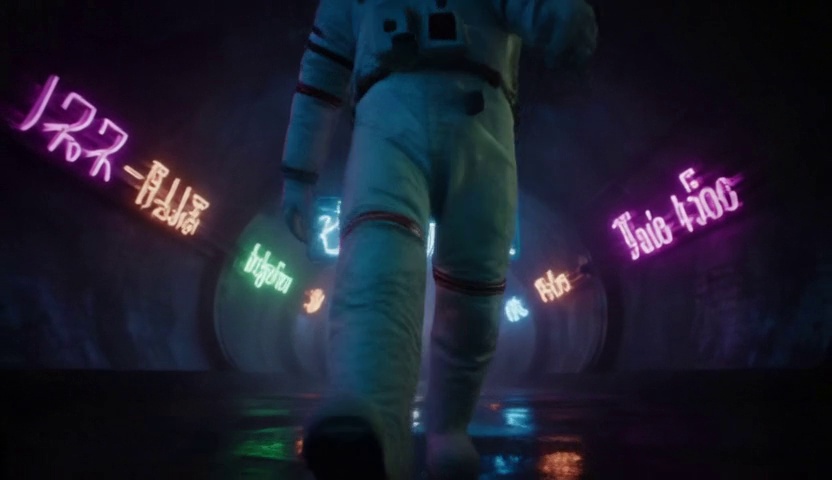}\hspace{0.5pt}  
        \includegraphics[width=0.17\textwidth]{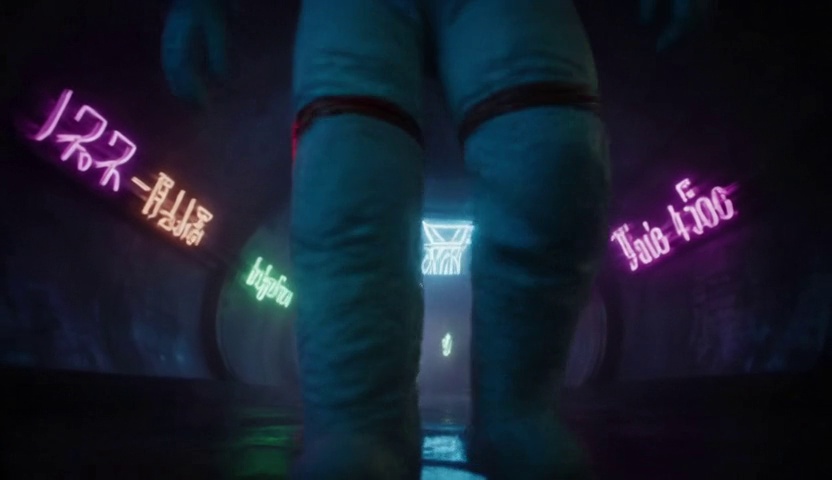}\hspace{0.5pt}   
        \includegraphics[width=0.17\textwidth]{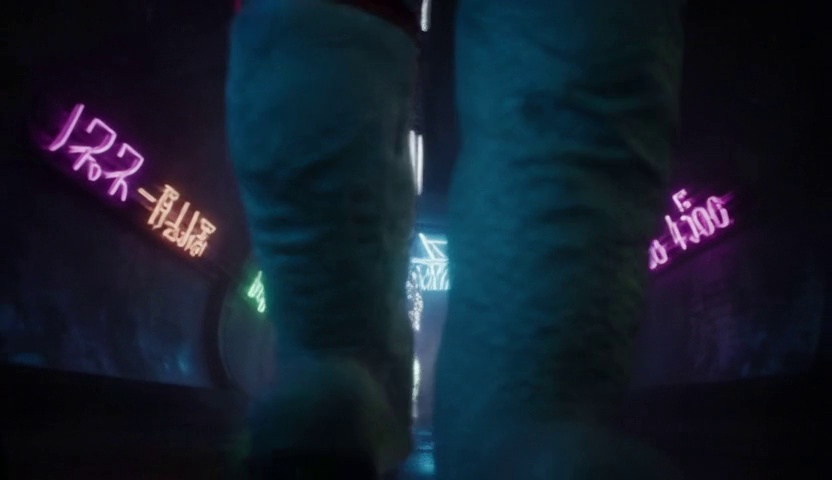}\hspace{0.5pt}  
        \\
        \\
        \raisebox{0.cm}{\rotatebox{90}{ 
            \sffamily\tiny\parbox{1.1cm}{\centering ReCamMaster} }}  &
        \includegraphics[width=0.17\textwidth]{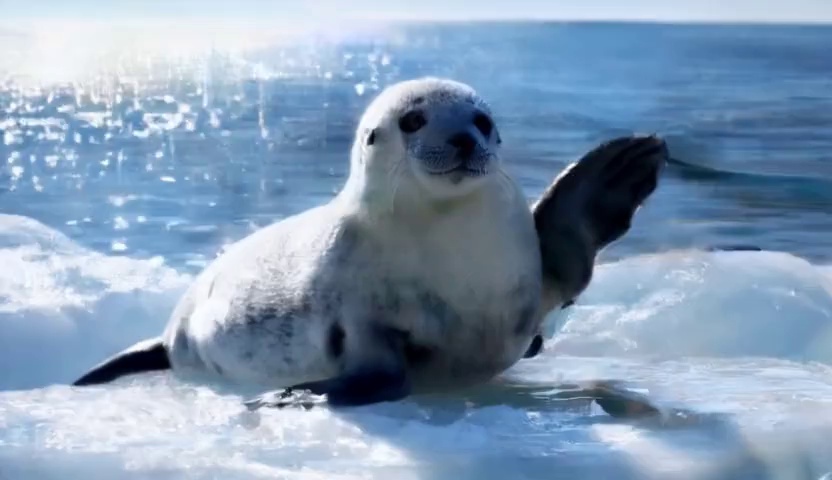}\hspace{0.5pt}  
        \includegraphics[width=0.17\textwidth]{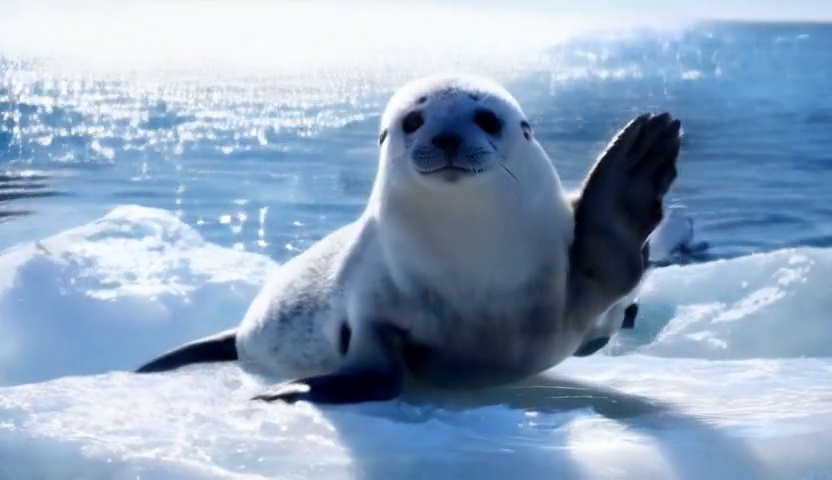}\hspace{0.5pt} 
        \includegraphics[width=0.17\textwidth]{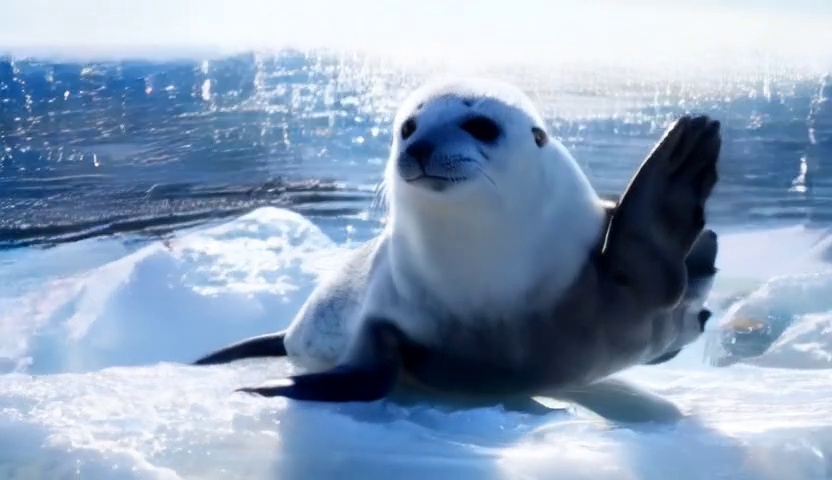}\hspace{0.5pt}   
        \includegraphics[width=0.17\textwidth]{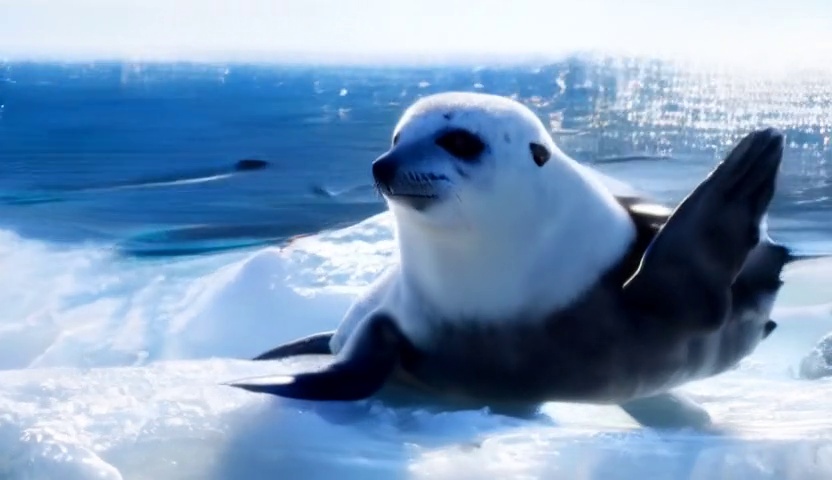}\hspace{0.5pt}  
        \includegraphics[width=0.17\textwidth]{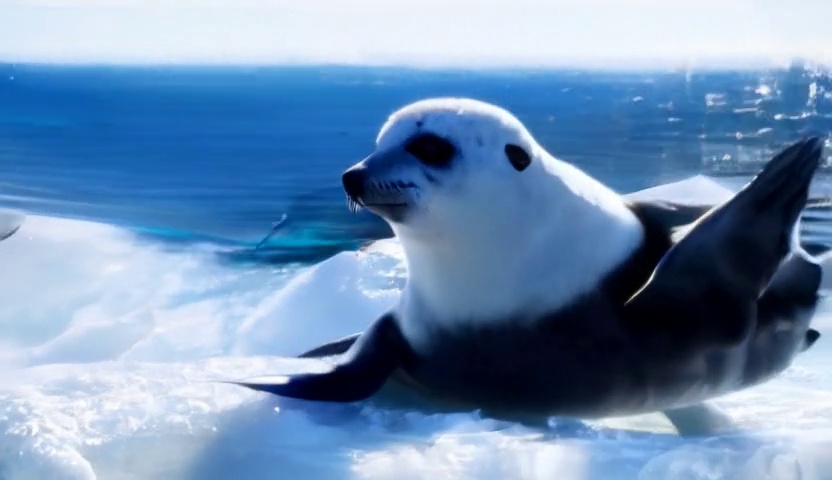}\hspace{0.5pt}  
        \\
        \raisebox{0.cm}{\rotatebox{90}{ 
            \sffamily\tiny\parbox{1.1cm}{\centering ReDirector} }}  &
        \includegraphics[width=0.17\textwidth]{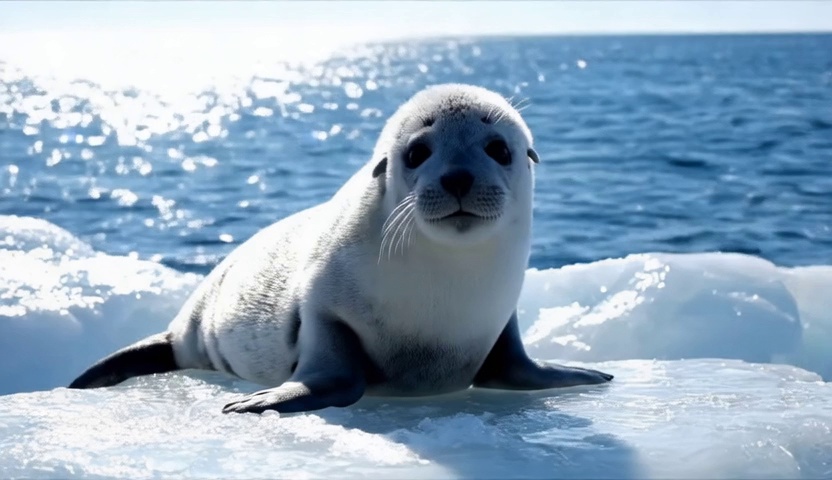}\hspace{0.5pt}  
        \includegraphics[width=0.17\textwidth]{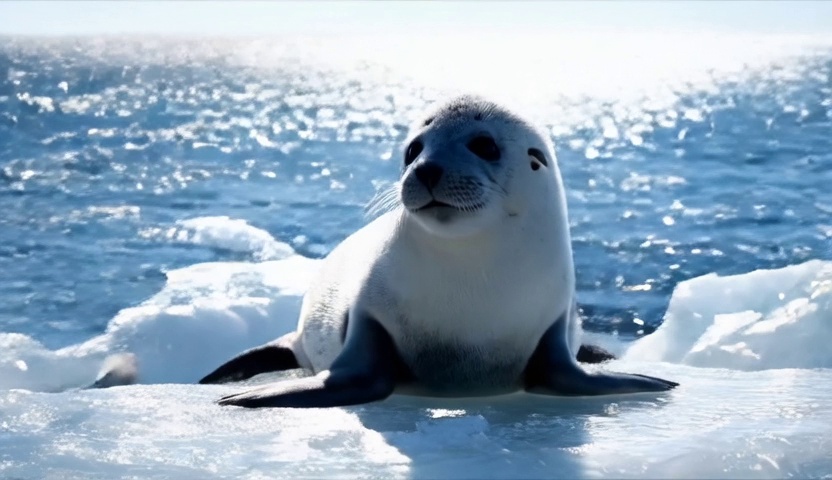}\hspace{0.5pt}  
        \includegraphics[width=0.17\textwidth]{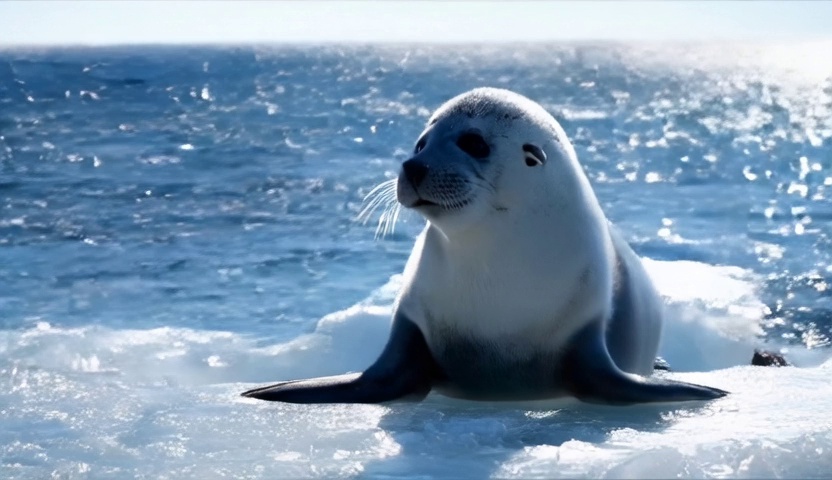}\hspace{0.5pt}  
        \includegraphics[width=0.17\textwidth]{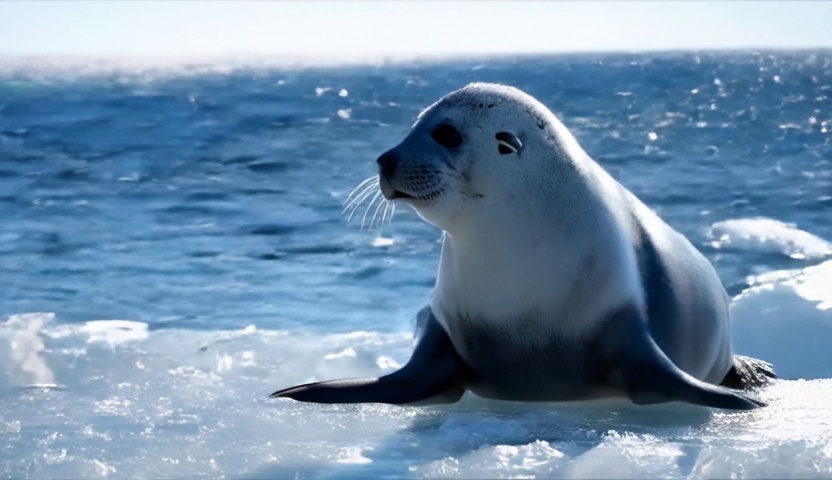}\hspace{0.5pt}  
        \includegraphics[width=0.17\textwidth]{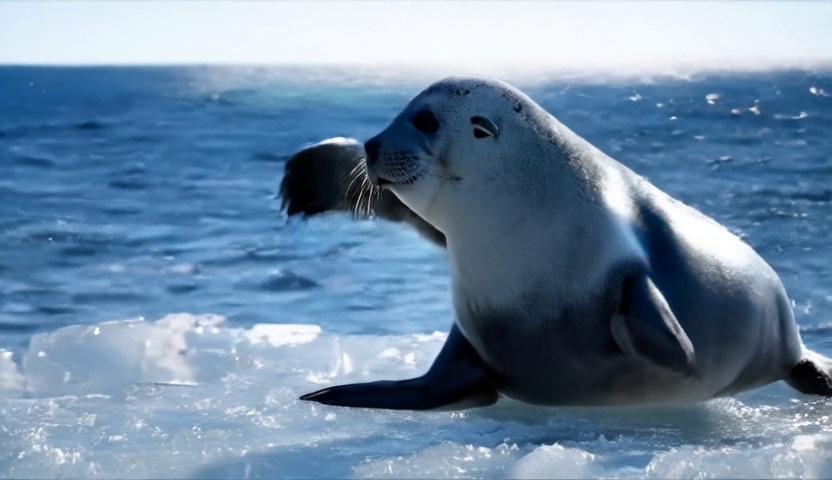}\hspace{0.5pt}  
        \\
        \raisebox{0.cm}{\rotatebox{90}{ 
            \sffamily\tiny\parbox{1.1cm}{\centering TrajectorC } }}  &
        \includegraphics[width=0.17\textwidth]{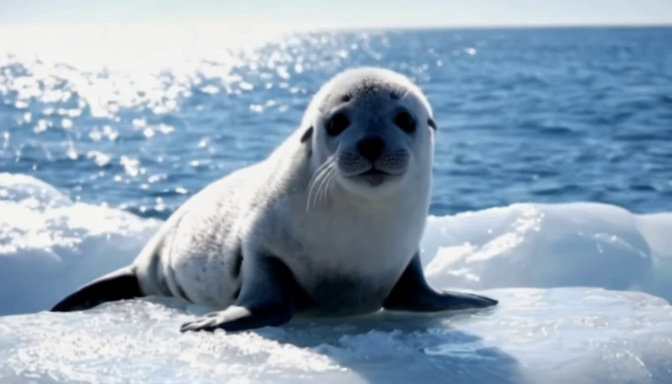}\hspace{0.5pt}  
        \includegraphics[width=0.17\textwidth]{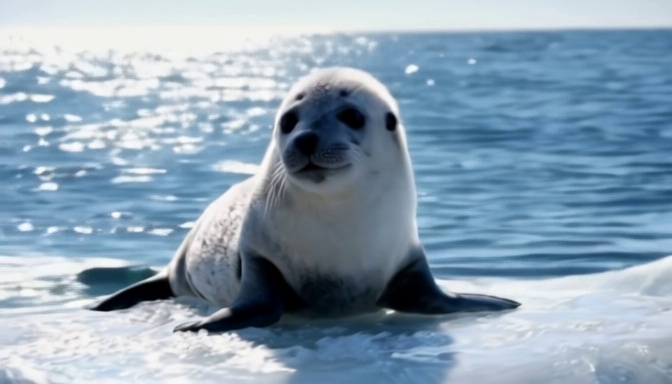}\hspace{0.5pt}  
        \includegraphics[width=0.17\textwidth]{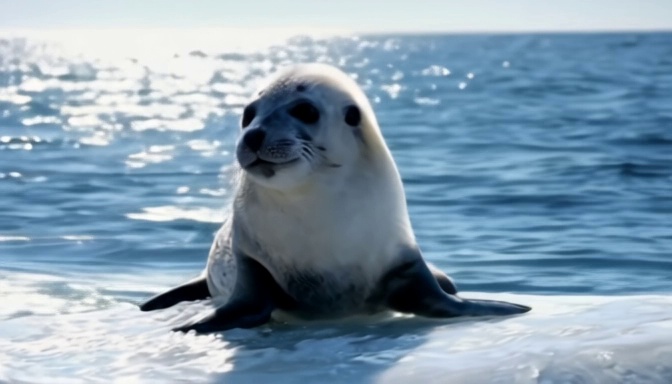}\hspace{0.5pt}  
        \includegraphics[width=0.17\textwidth]{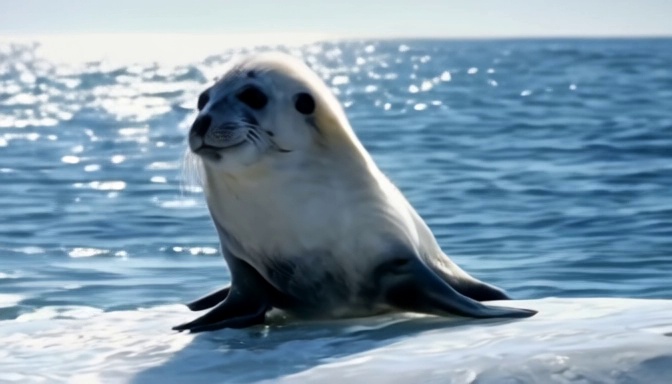}\hspace{0.5pt}  
        \includegraphics[width=0.17\textwidth]{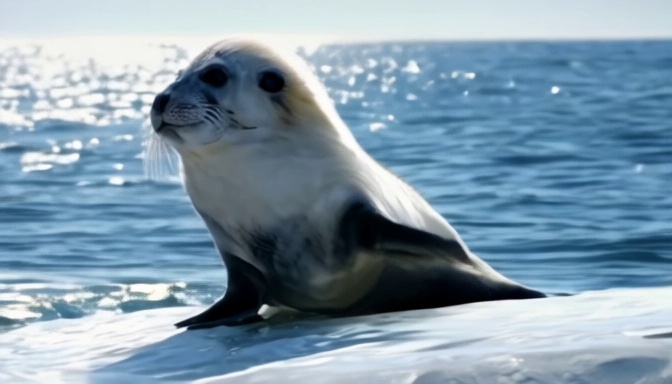}\hspace{0.5pt}     
        \\
        \raisebox{0.cm}{\rotatebox{90}{
            \sffamily\tiny\parbox{1.1cm}{\centering Ours} }}  &
        \includegraphics[width=0.17\textwidth]{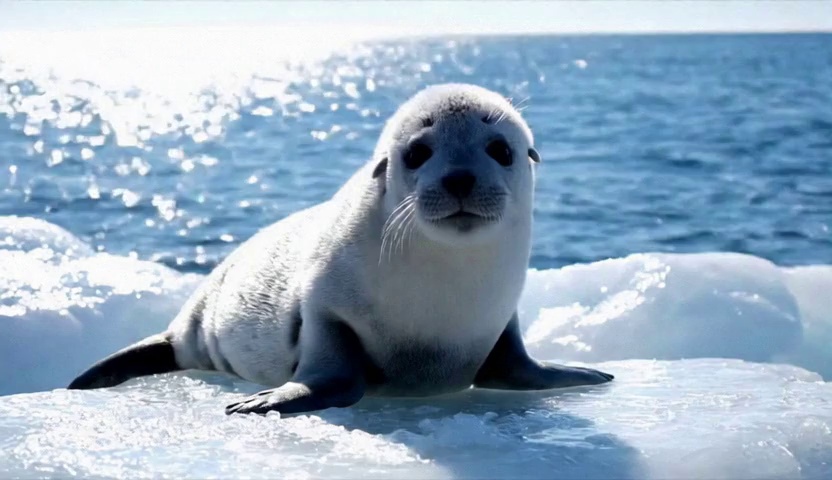}\hspace{0.5pt}  
        \includegraphics[width=0.17\textwidth]{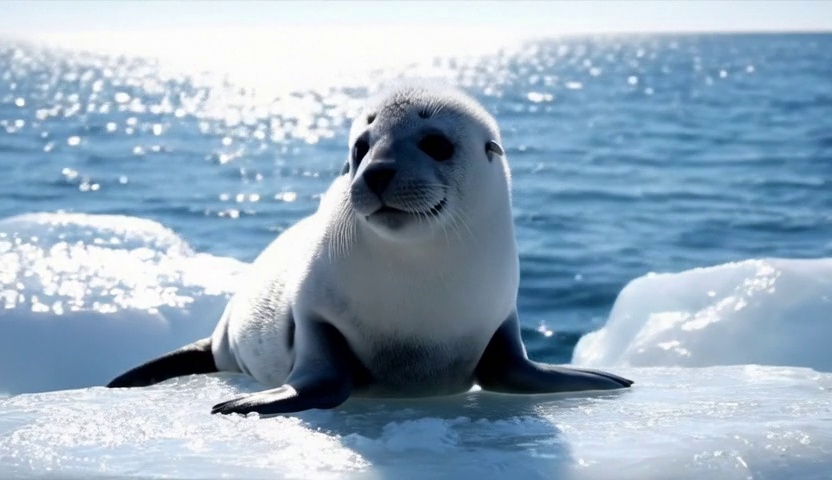}\hspace{0.5pt}  
        \includegraphics[width=0.17\textwidth]{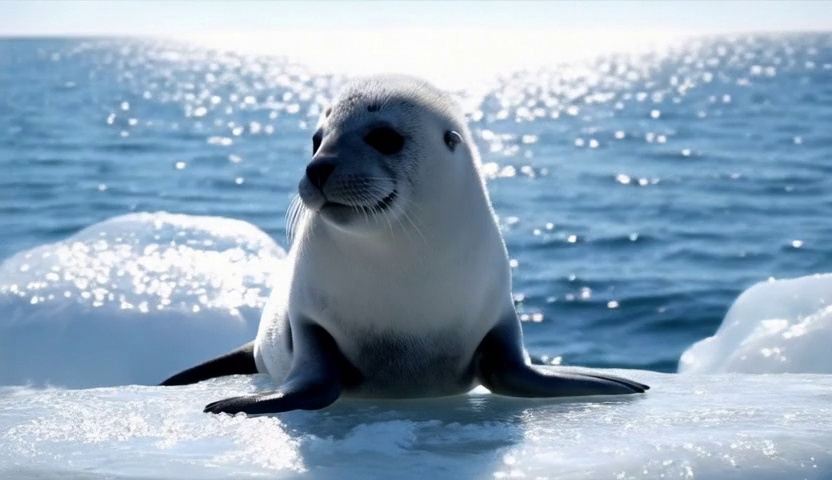}\hspace{0.5pt}  
        \includegraphics[width=0.17\textwidth]{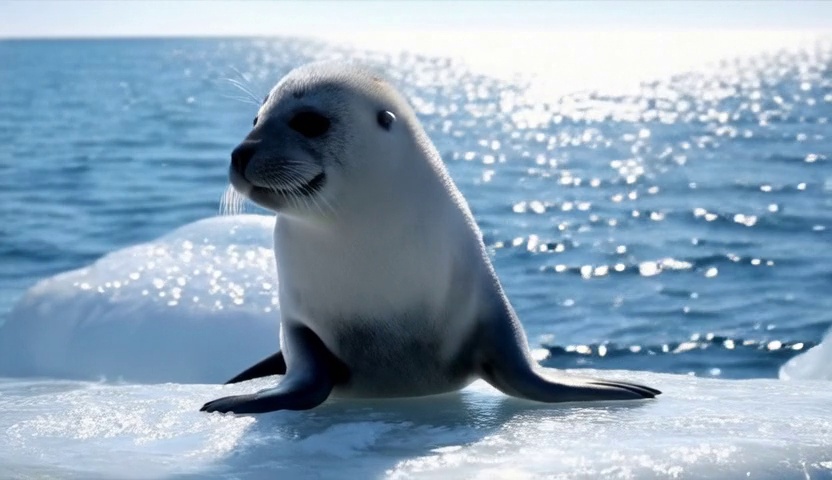}\hspace{0.5pt}  
        \includegraphics[width=0.17\textwidth]{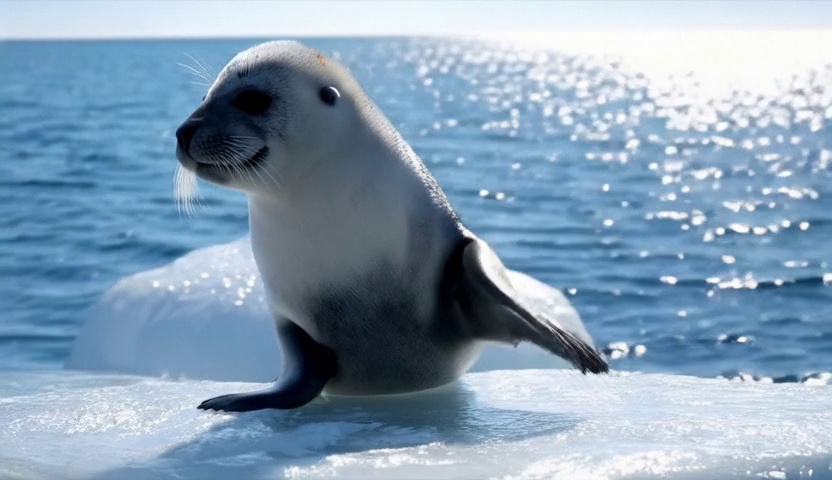}\hspace{0.5pt}   
   \end{tabular}
    \caption{\textbf{Additional qualitative comparison on generated video by Veo}.}
    \vspace{-0.1cm}
    \label{fig:con_compare_set1_p1}
\end{figure*}

\begin{figure*}[htbp]
    \centering
    \begin{tabular}{@{}c@{\hspace{2pt}}c@{\hspace{2pt}}c@{\hspace{2pt}}}
        \raisebox{0.cm}{\rotatebox{90}{%
            \sffamily\tiny\parbox{1.1cm}{\centering ReCamMaster}%
        }}   &

        \includegraphics[width=0.17\textwidth]{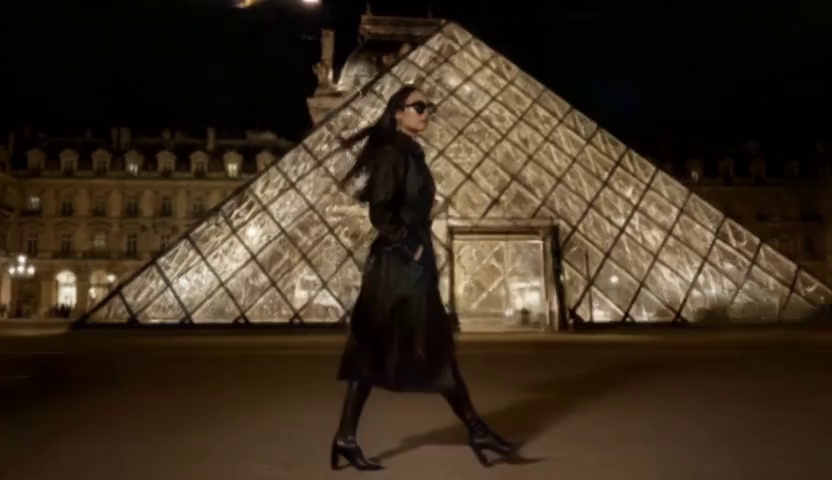}\hspace{0.5pt}  
        \includegraphics[width=0.17\textwidth]{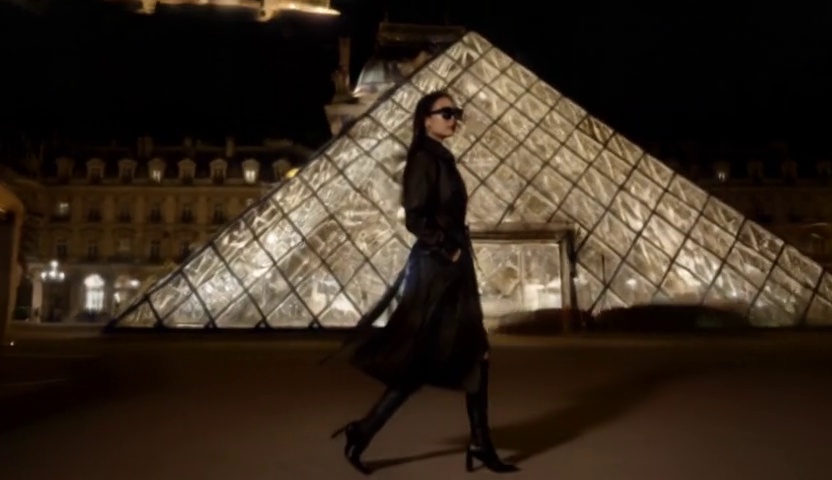}\hspace{0.5pt}  
        \includegraphics[width=0.17\textwidth]{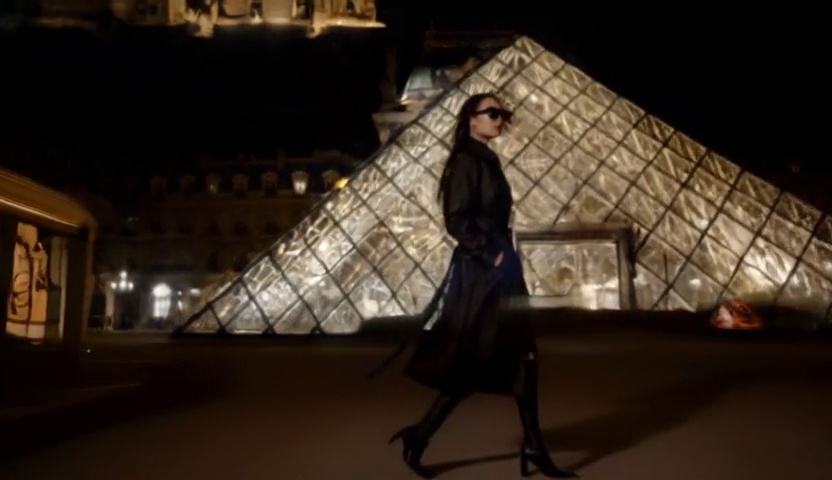}\hspace{0.5pt}  
        \includegraphics[width=0.17\textwidth]{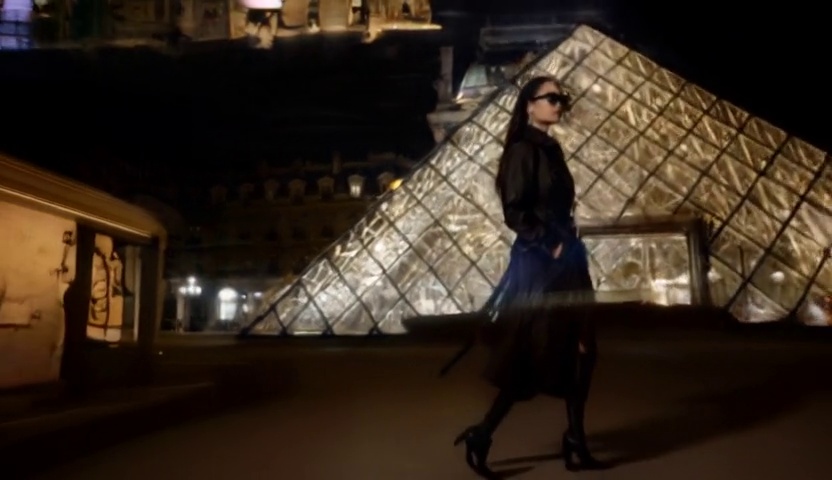}\hspace{0.5pt}  
        \includegraphics[width=0.17\textwidth]{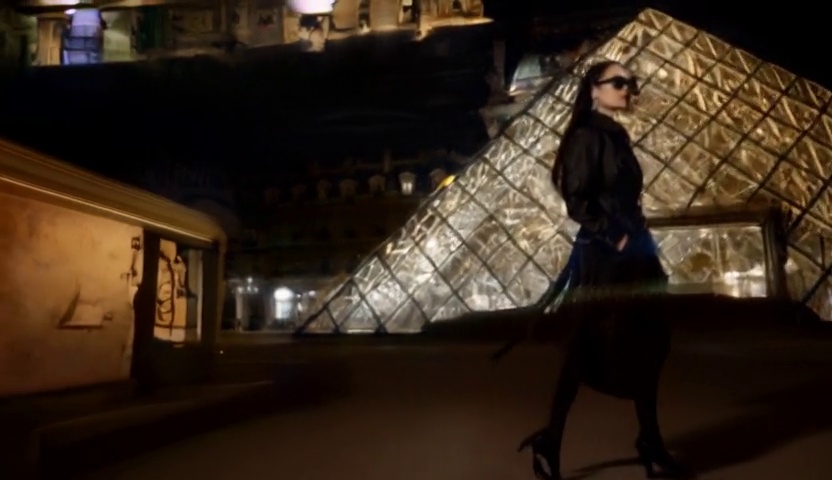}\hspace{0.5pt}  
        \\
        \raisebox{0.cm}{\rotatebox{90}{ 
            \sffamily\tiny\parbox{1.1cm}{\centering ReDirector} }}  &
        \includegraphics[width=0.17\textwidth]{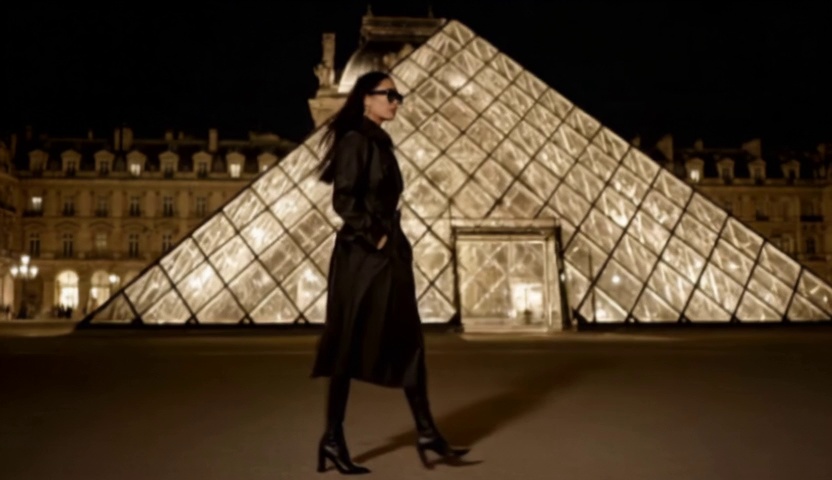}\hspace{0.5pt}  
        \includegraphics[width=0.17\textwidth]{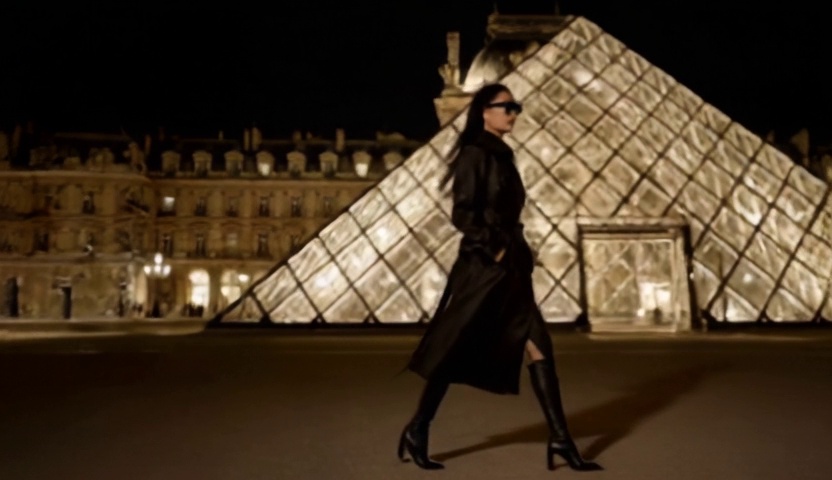}\hspace{0.5pt}  
        \includegraphics[width=0.17\textwidth]{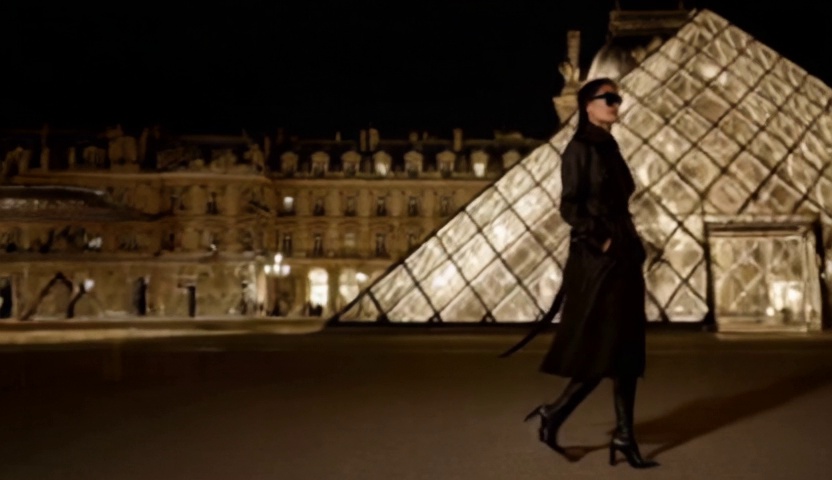}\hspace{0.5pt}  
        \includegraphics[width=0.17\textwidth]{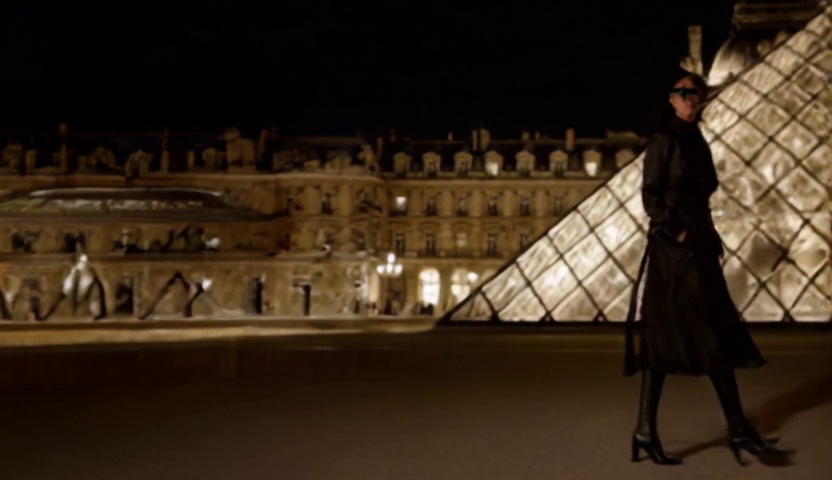}\hspace{0.5pt}  
        \includegraphics[width=0.17\textwidth]{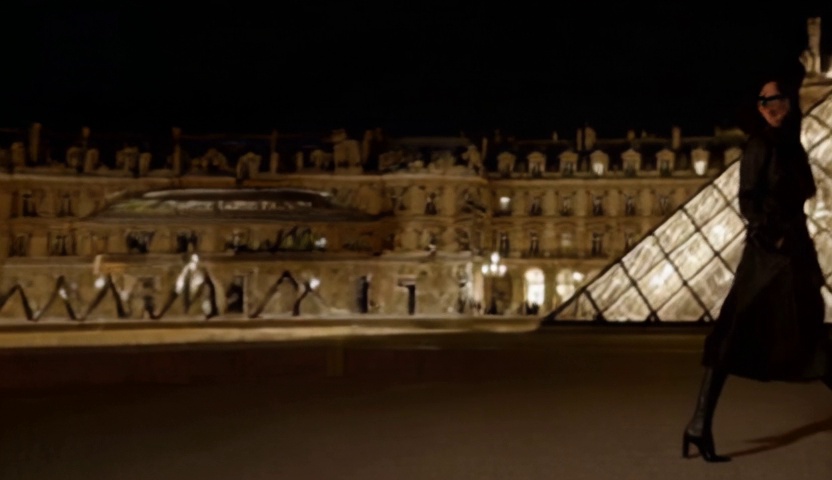}\hspace{0.5pt}  
        \\
        \raisebox{0.cm}{\rotatebox{90}{ 
            \sffamily\tiny\parbox{1.1cm}{\centering TrajectorC} }}  &
        \includegraphics[width=0.17\textwidth]{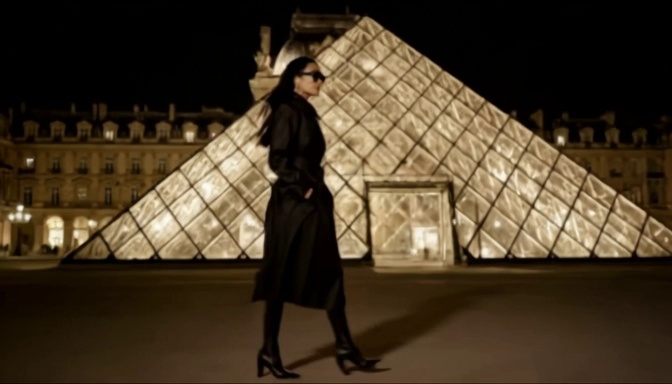}\hspace{0.5pt}  
        \includegraphics[width=0.17\textwidth]{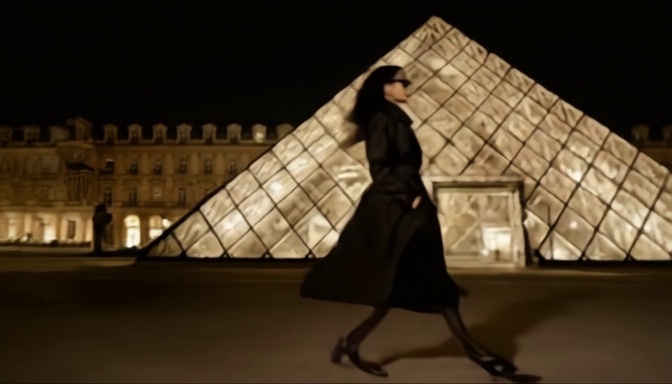}\hspace{0.5pt}  
        \includegraphics[width=0.17\textwidth]{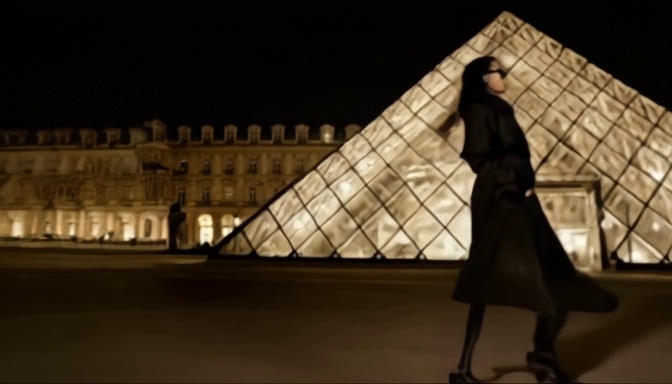}\hspace{0.5pt}  
        \includegraphics[width=0.17\textwidth]{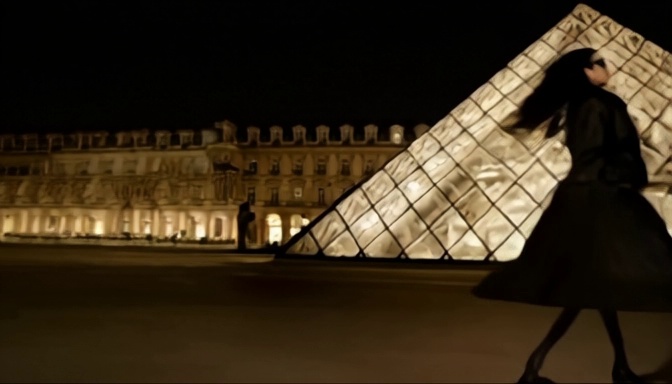}\hspace{0.5pt}    
        \includegraphics[width=0.17\textwidth]{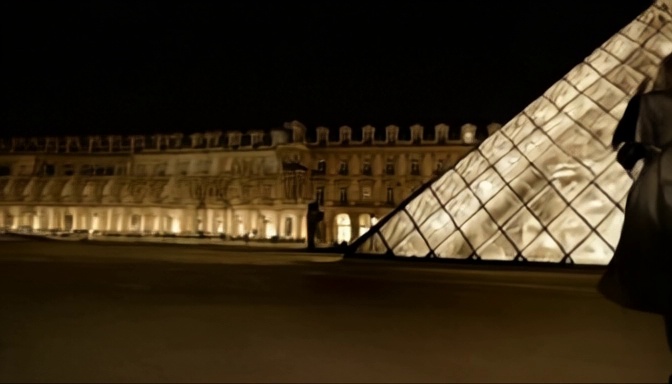}\hspace{0.5pt}   
        \\
        \raisebox{0.cm}{\rotatebox{90}{
            \sffamily\tiny\parbox{1.1cm}{\centering Ours} }}  &
        \includegraphics[width=0.17\textwidth]{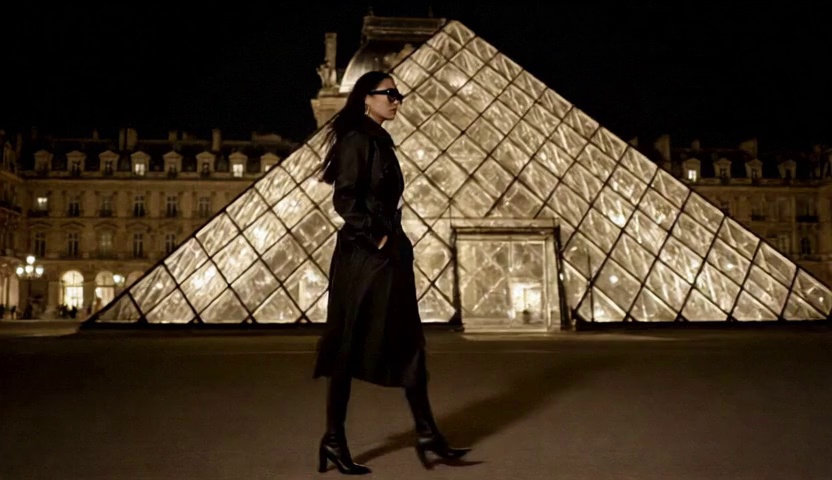}\hspace{0.5pt}  
        \includegraphics[width=0.17\textwidth]{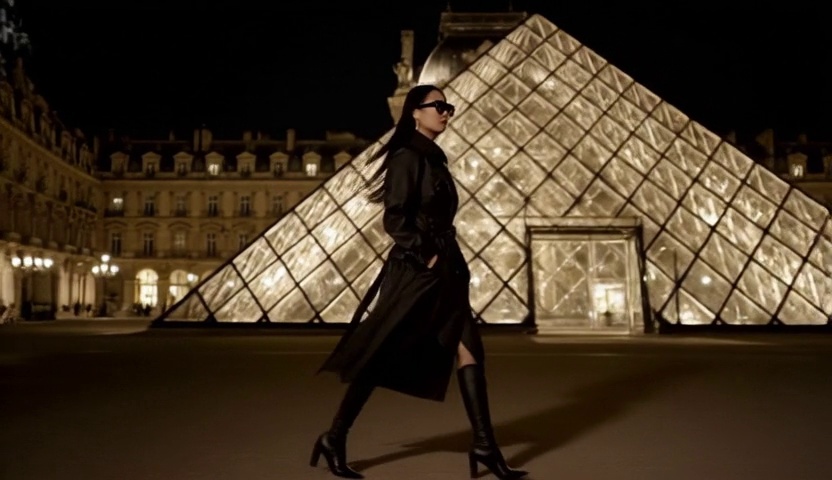}\hspace{0.5pt}  
        \includegraphics[width=0.17\textwidth]{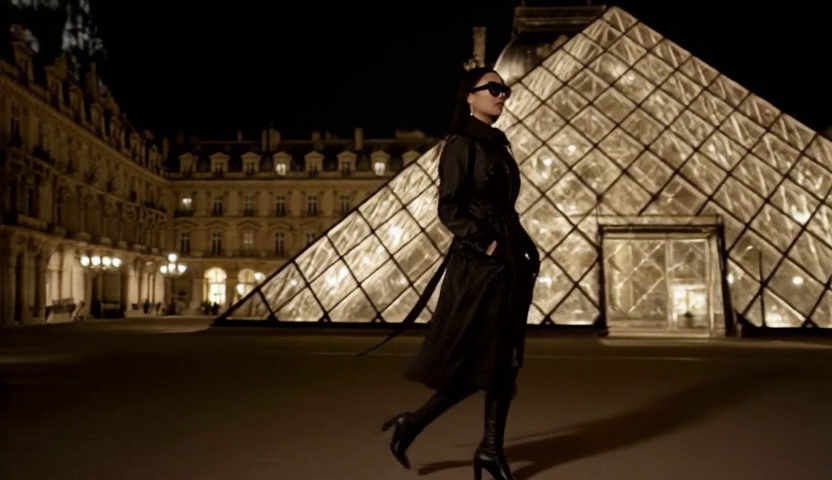}\hspace{0.5pt}  
        \includegraphics[width=0.17\textwidth]{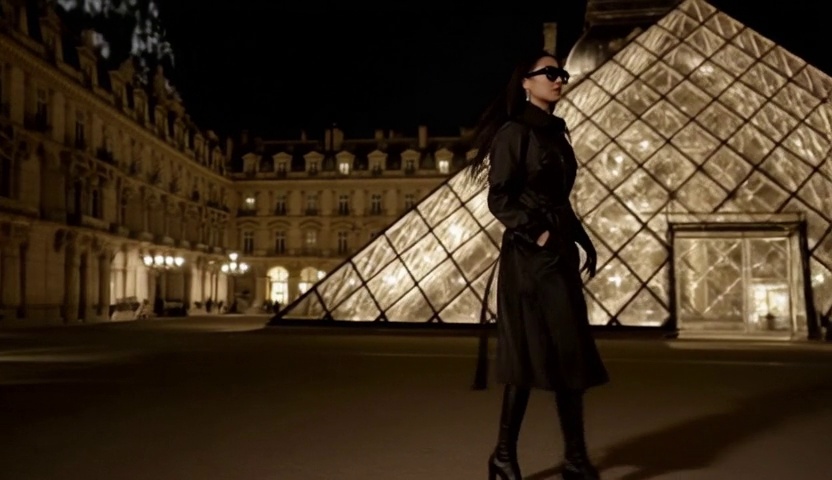}\hspace{0.5pt}  
        \includegraphics[width=0.17\textwidth]{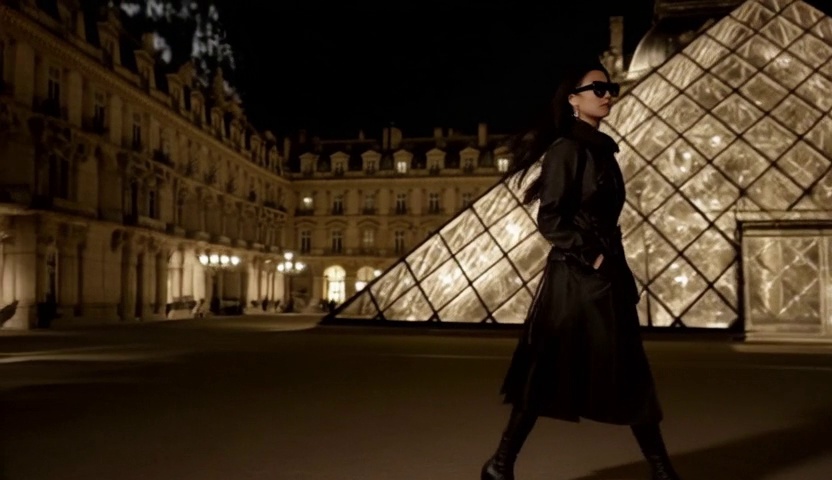}\hspace{0.5pt}  
        \\
        \\
        \raisebox{0.cm}{\rotatebox{90}{ 
            \sffamily\tiny\parbox{1.1cm}{\centering ReCamMaster} }}  &
        \includegraphics[width=0.17\textwidth]{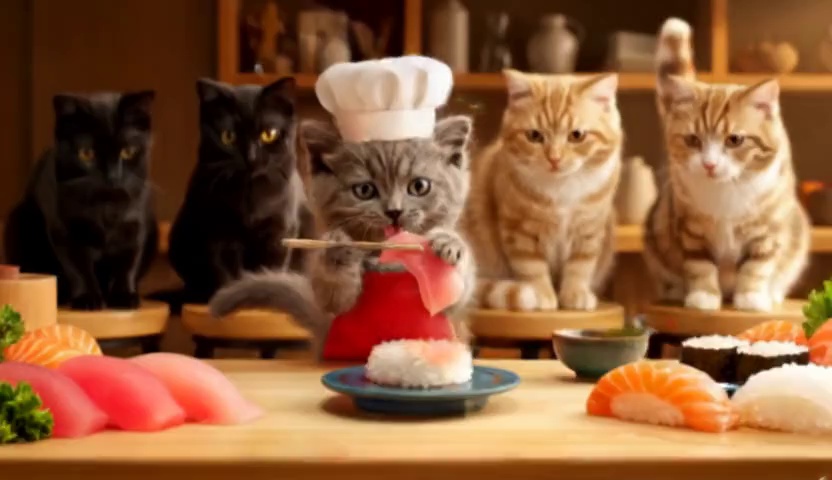}\hspace{0.5pt}  
        \includegraphics[width=0.17\textwidth]{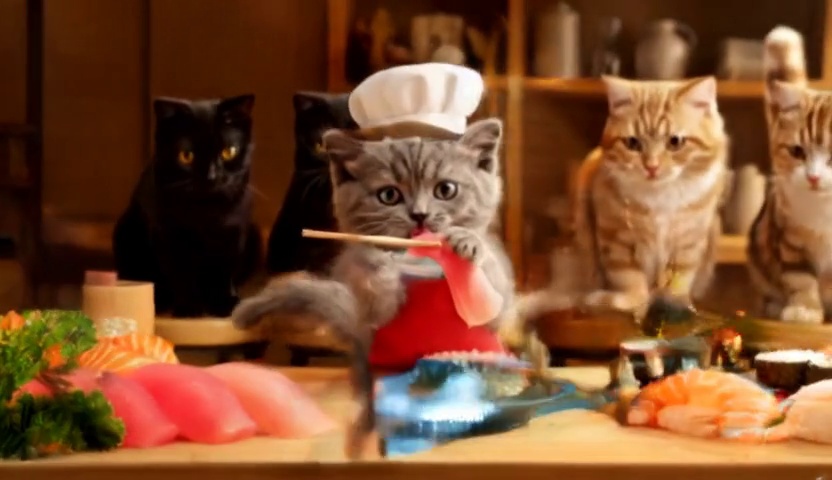}\hspace{0.5pt}  
        \includegraphics[width=0.17\textwidth]{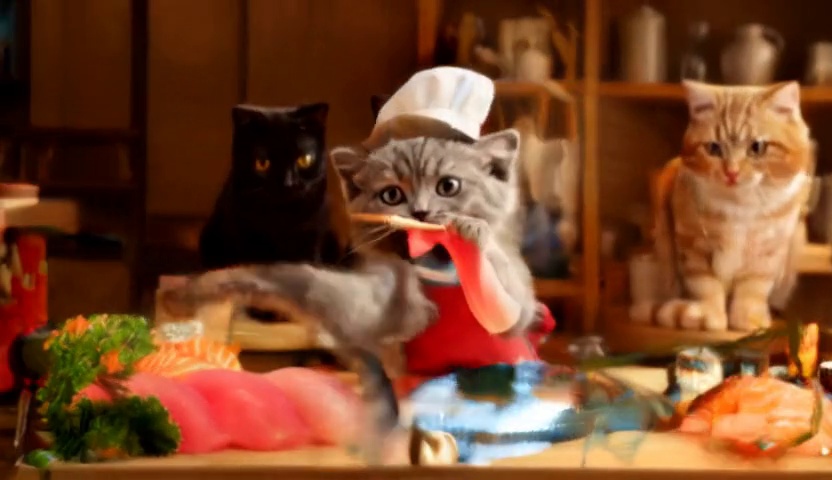}\hspace{0.5pt}  
        \includegraphics[width=0.17\textwidth]{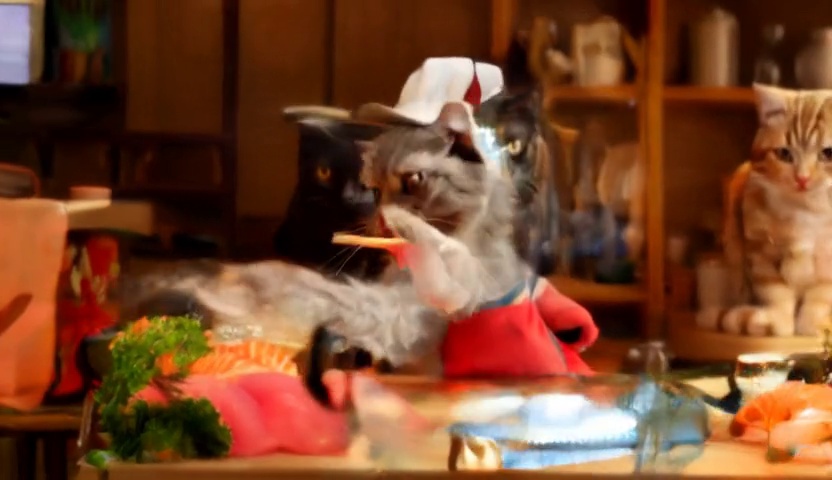}\hspace{0.5pt}   
        \includegraphics[width=0.17\textwidth]{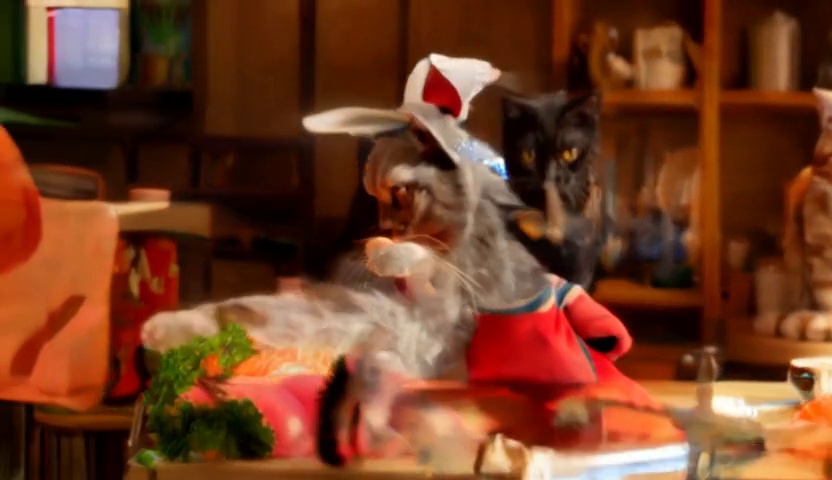}\hspace{0.5pt}   
        \\
        \raisebox{0.cm}{\rotatebox{90}{ 
            \sffamily\tiny\parbox{1.1cm}{\centering ReDirector} }}  &
        \includegraphics[width=0.17\textwidth]{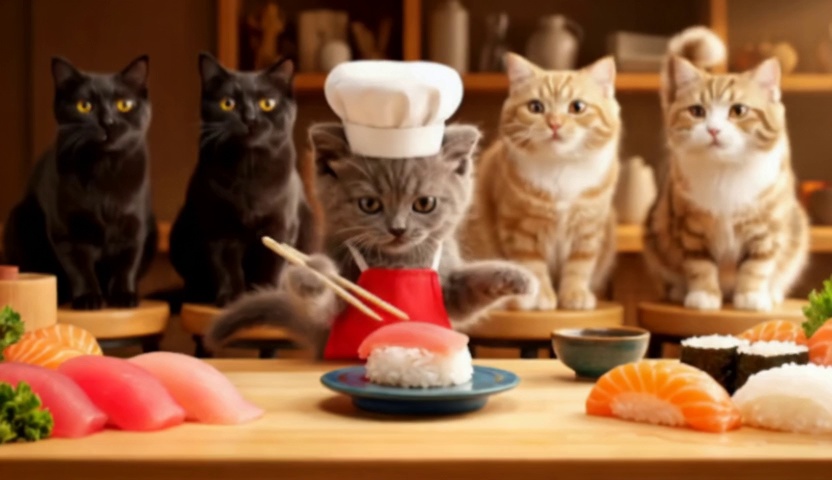}\hspace{0.5pt}  
        \includegraphics[width=0.17\textwidth]{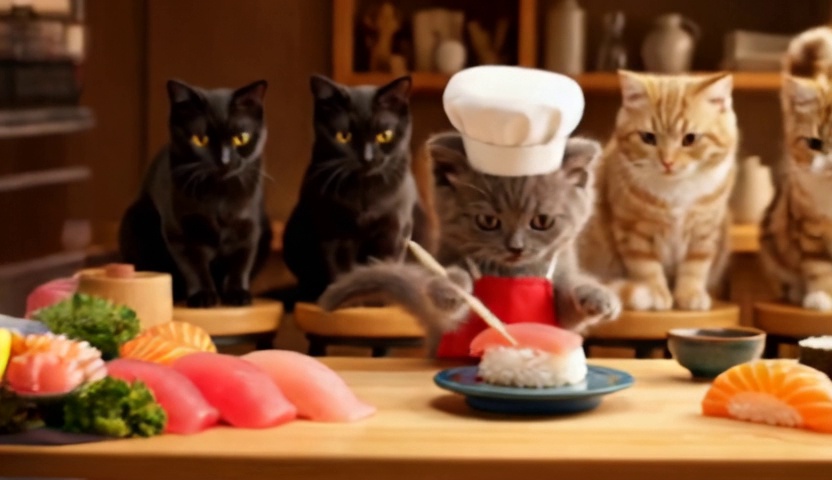}\hspace{0.5pt}  
        \includegraphics[width=0.17\textwidth]{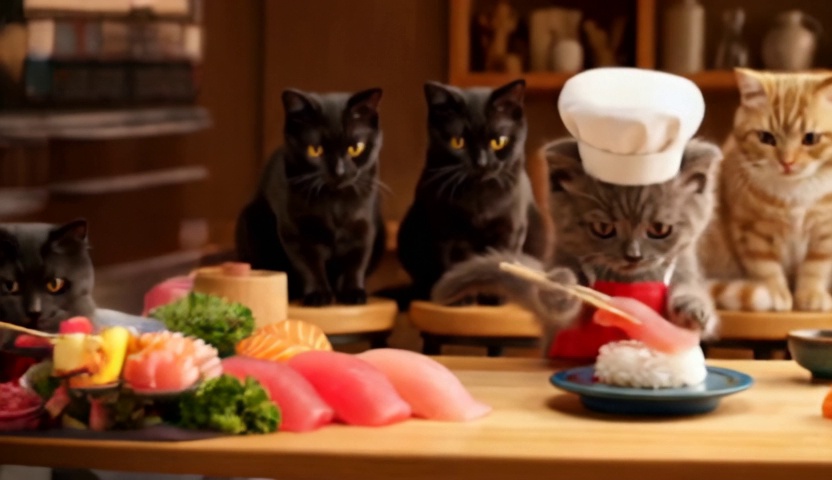}\hspace{0.5pt}  
        \includegraphics[width=0.17\textwidth]{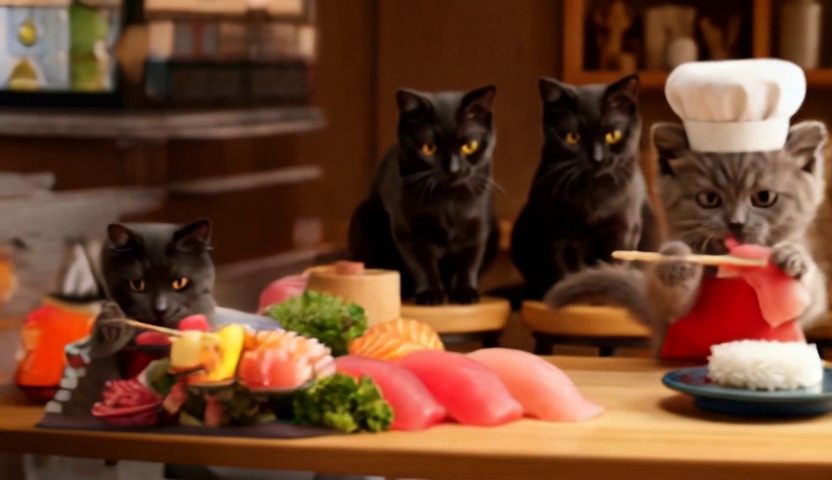}\hspace{0.5pt}  
        \includegraphics[width=0.17\textwidth]{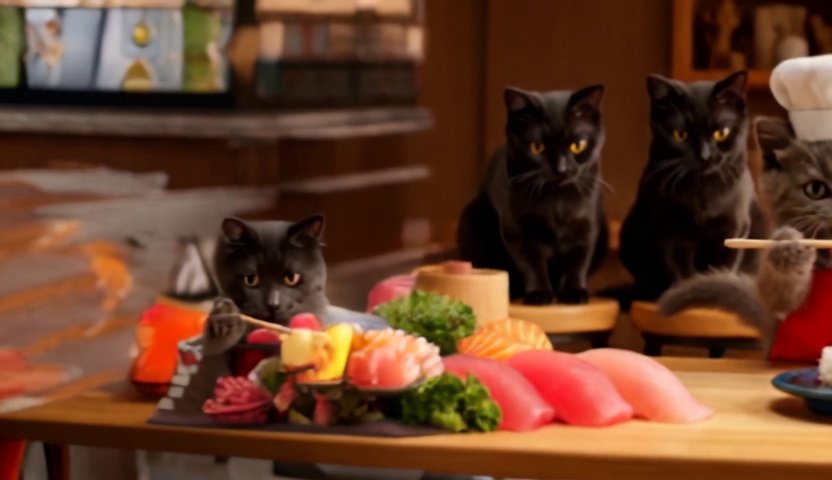}\hspace{0.5pt}  
        \\
        \raisebox{0.cm}{\rotatebox{90}{ 
            \sffamily\tiny\parbox{1.1cm}{\centering TrajectorC } }}  &
        \includegraphics[width=0.17\textwidth]{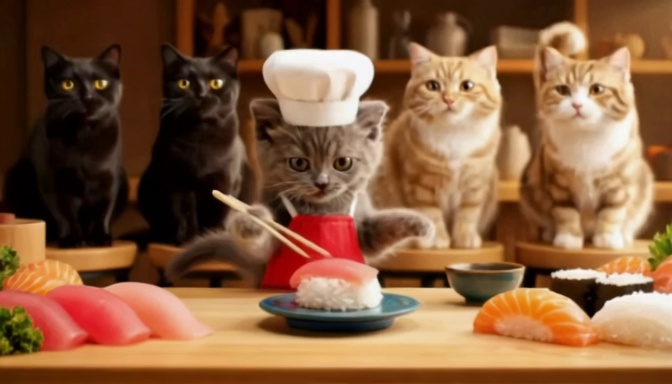}\hspace{0.5pt}  
        \includegraphics[width=0.17\textwidth]{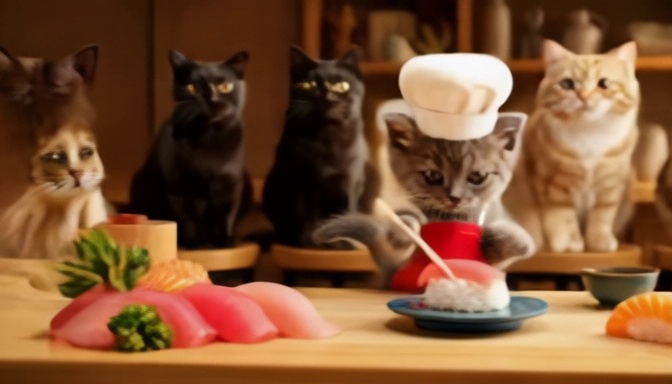}\hspace{0.5pt}  
        \includegraphics[width=0.17\textwidth]{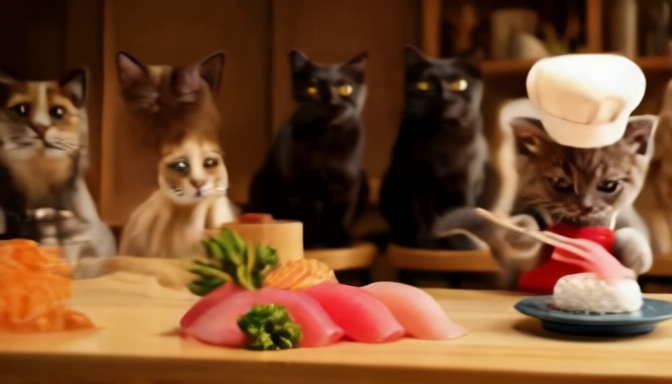}\hspace{0.5pt}  
        \includegraphics[width=0.17\textwidth]{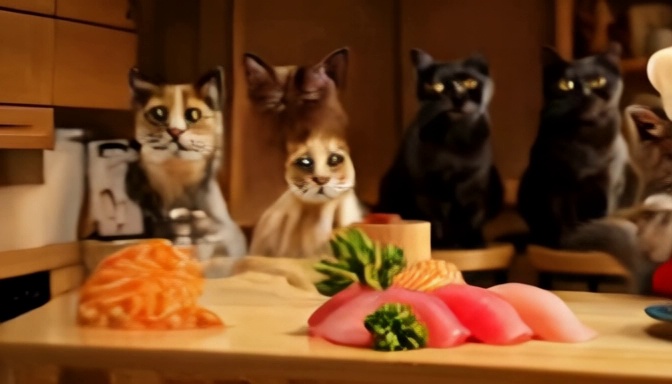}\hspace{0.5pt}     
        \includegraphics[width=0.17\textwidth]{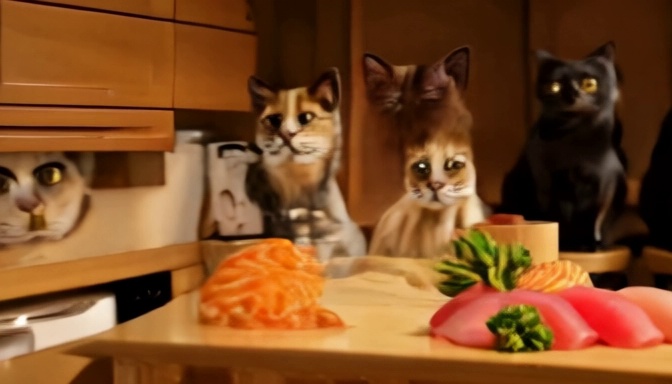}\hspace{0.5pt}   
        \\
        \raisebox{0.cm}{\rotatebox{90}{
            \sffamily\tiny\parbox{1.1cm}{\centering Ours} }}  &
        \includegraphics[width=0.17\textwidth]{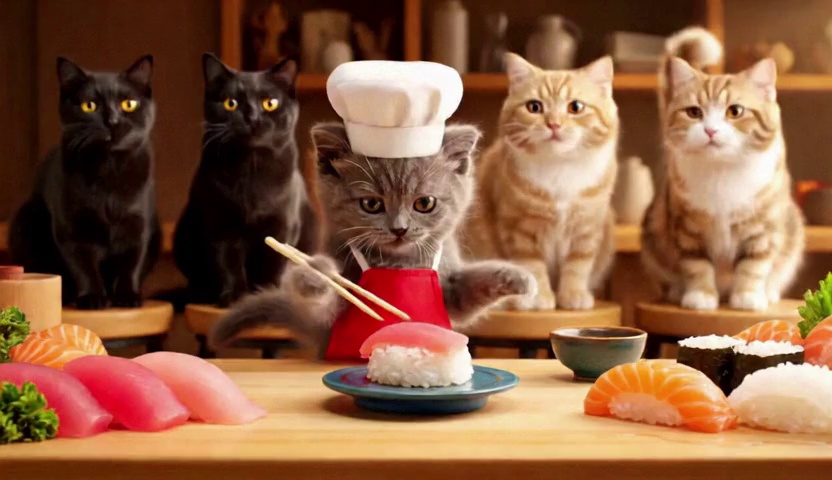}\hspace{0.5pt}  
        \includegraphics[width=0.17\textwidth]{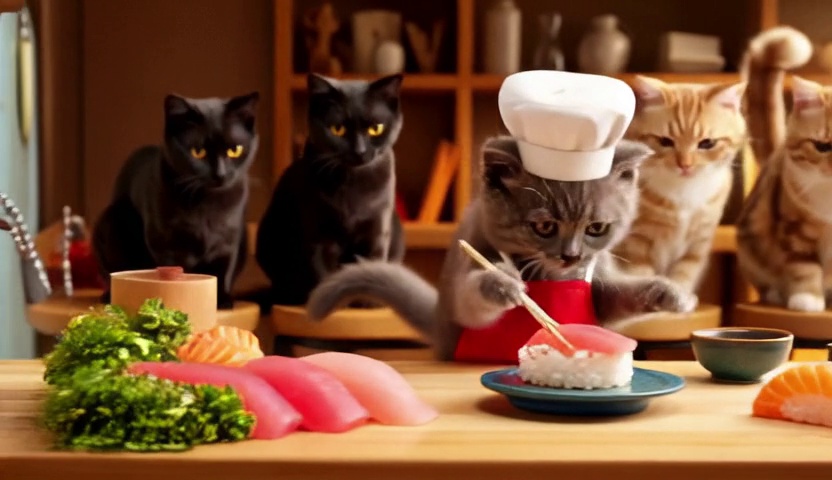}\hspace{0.5pt}  
        \includegraphics[width=0.17\textwidth]{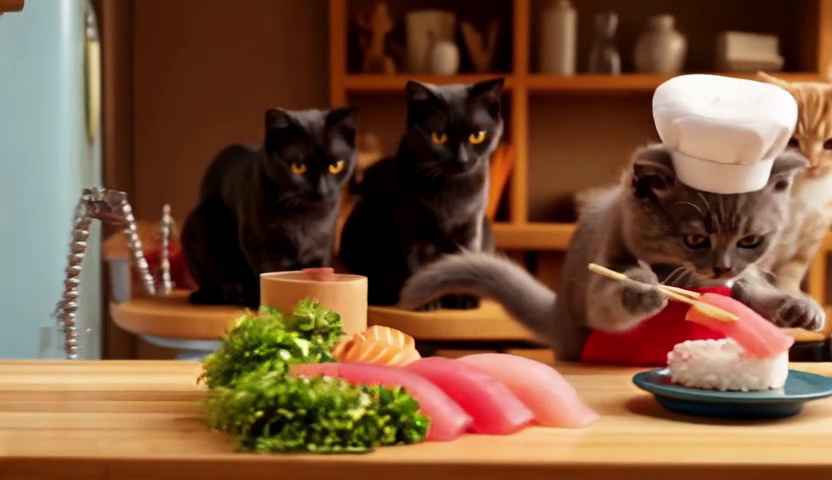}\hspace{0.5pt}  
        \includegraphics[width=0.17\textwidth]{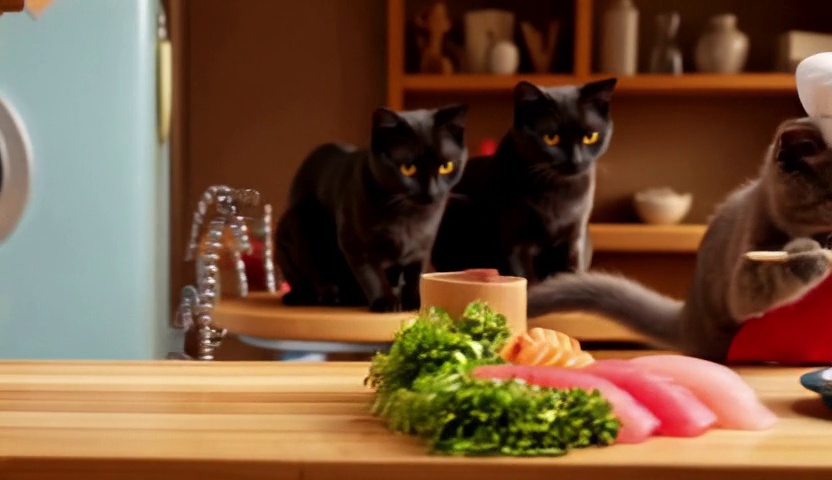}\hspace{0.5pt}   
        \includegraphics[width=0.17\textwidth]{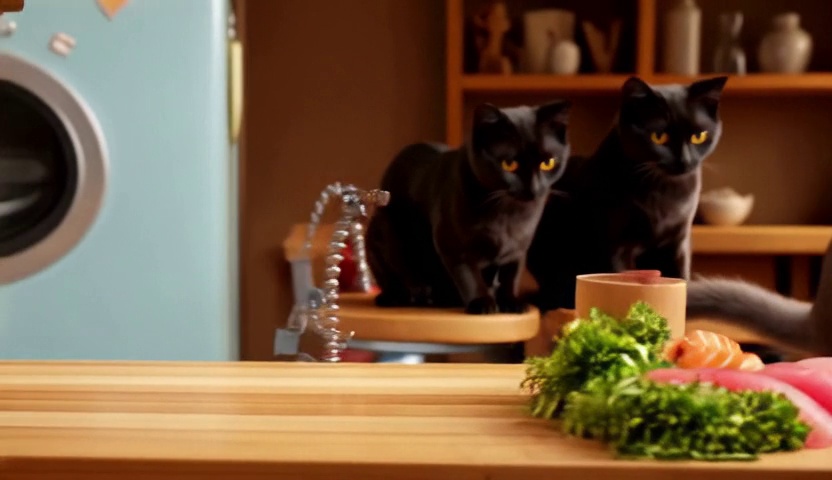}\hspace{0.5pt}  
        \\
        \\
        \raisebox{0.cm}{\rotatebox{90}{ 
            \sffamily\tiny\parbox{1.1cm}{\centering ReCamMaster} }}  &
        \includegraphics[width=0.17\textwidth]{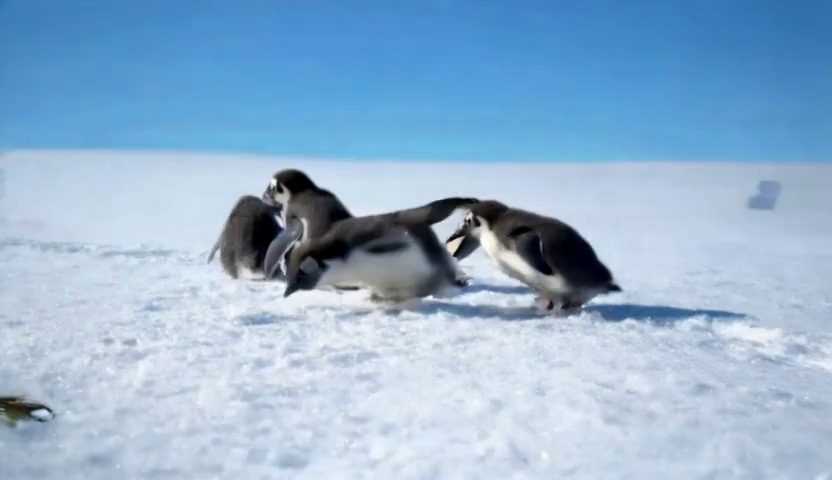}\hspace{0.5pt}  
        \includegraphics[width=0.17\textwidth]{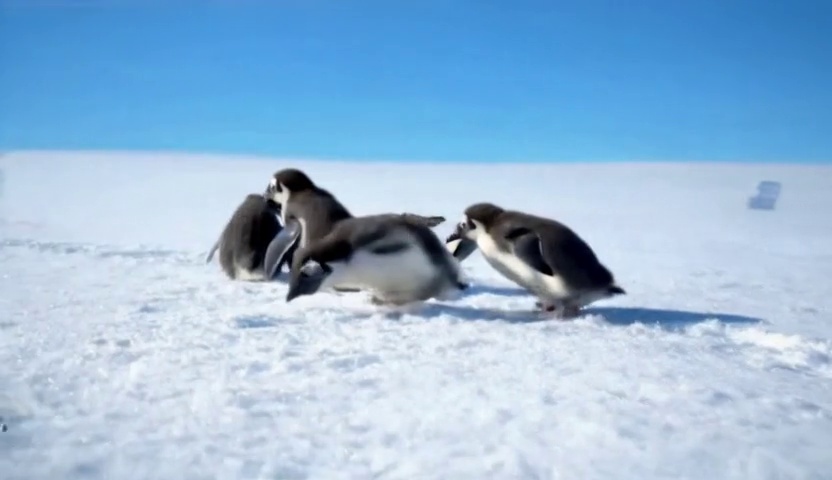}\hspace{0.5pt} 
        \includegraphics[width=0.17\textwidth]{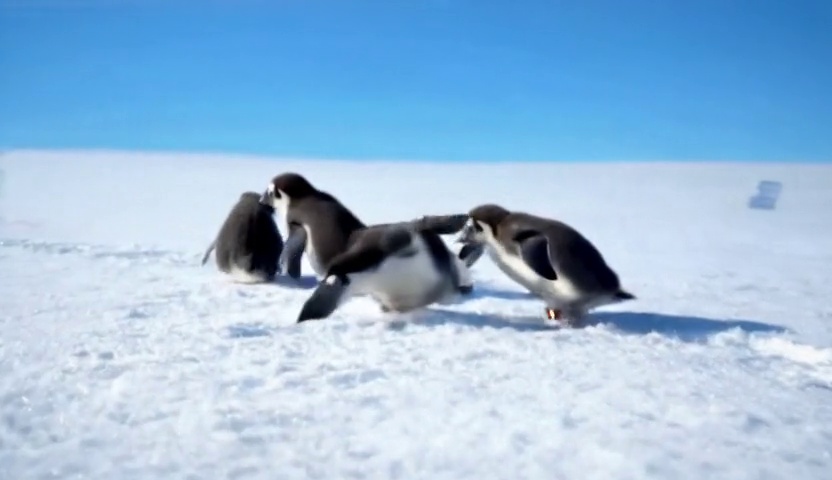}\hspace{0.5pt}   
        \includegraphics[width=0.17\textwidth]{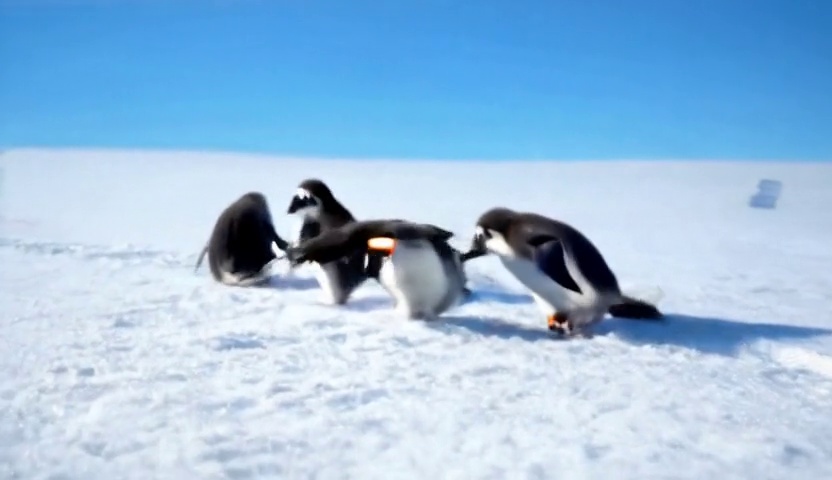}\hspace{0.5pt}  
        \includegraphics[width=0.17\textwidth]{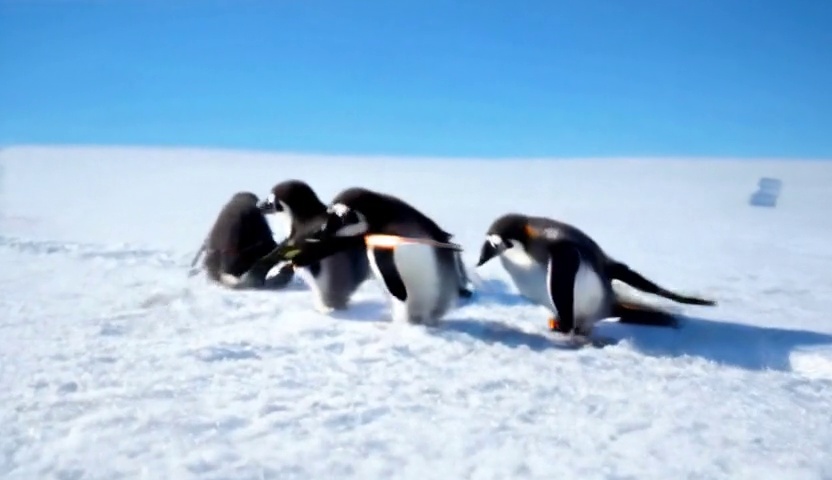}\hspace{0.5pt}  
        \\
        \raisebox{0.cm}{\rotatebox{90}{ 
            \sffamily\tiny\parbox{1.1cm}{\centering ReDirector} }}  &
        \includegraphics[width=0.17\textwidth]{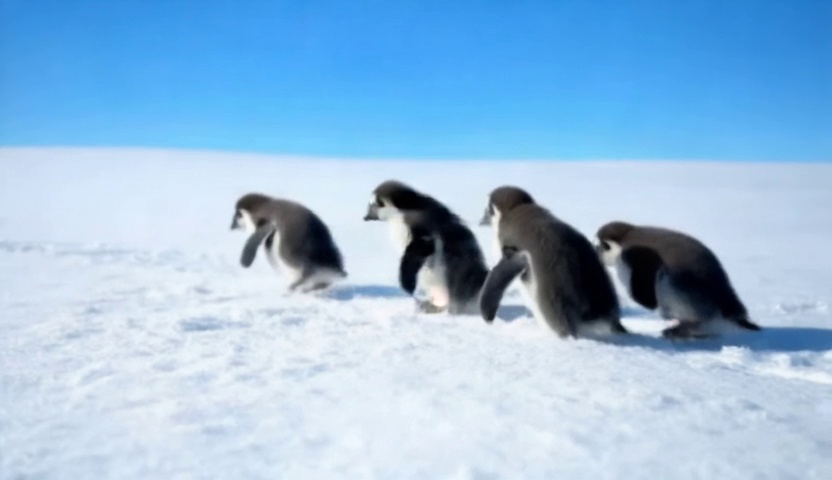}\hspace{0.5pt}  
        \includegraphics[width=0.17\textwidth]{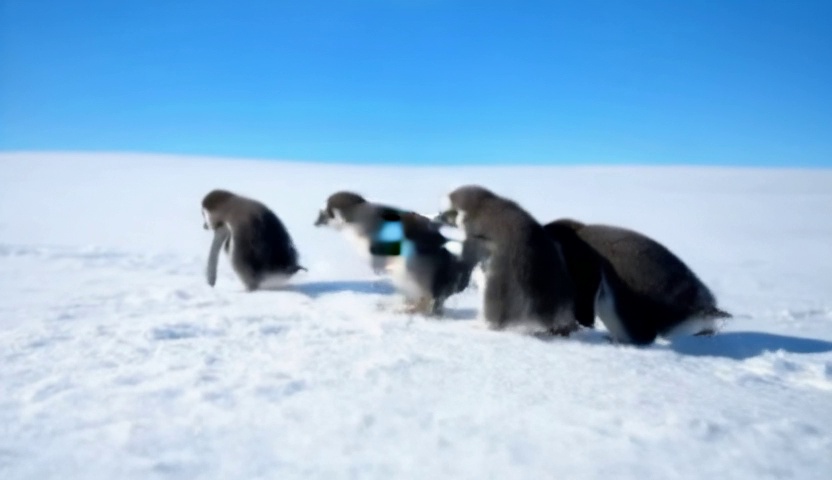}\hspace{0.5pt}  
        \includegraphics[width=0.17\textwidth]{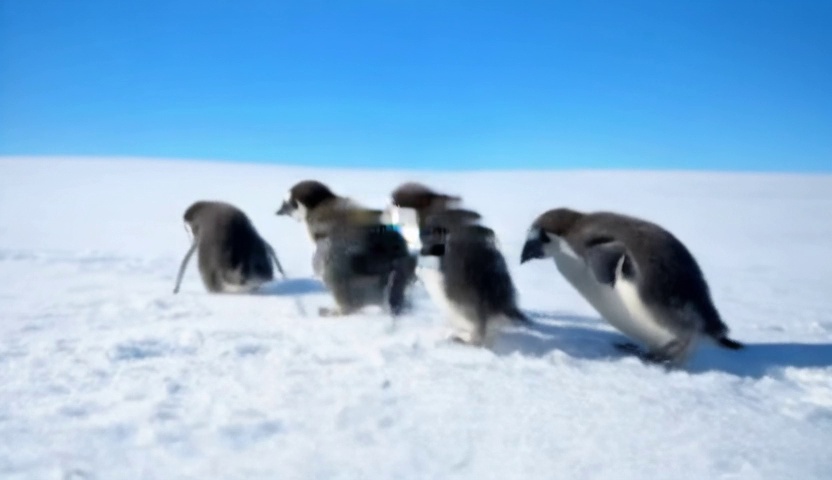}\hspace{0.5pt}  
        \includegraphics[width=0.17\textwidth]{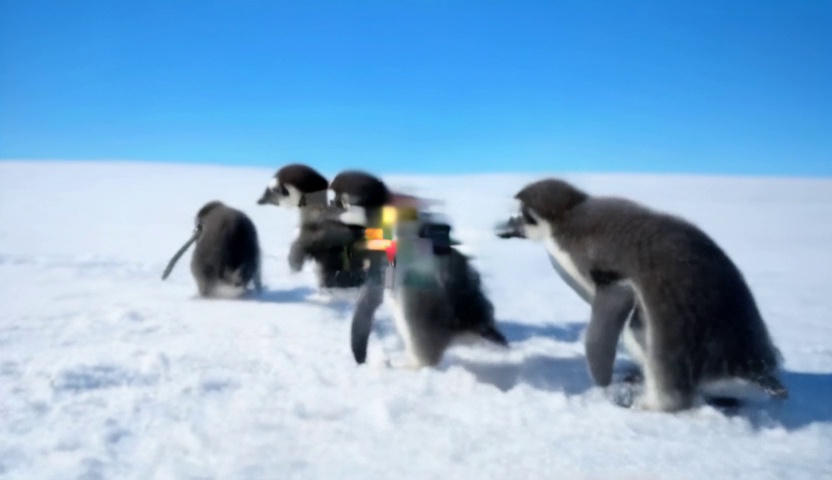}\hspace{0.5pt}  
        \includegraphics[width=0.17\textwidth]{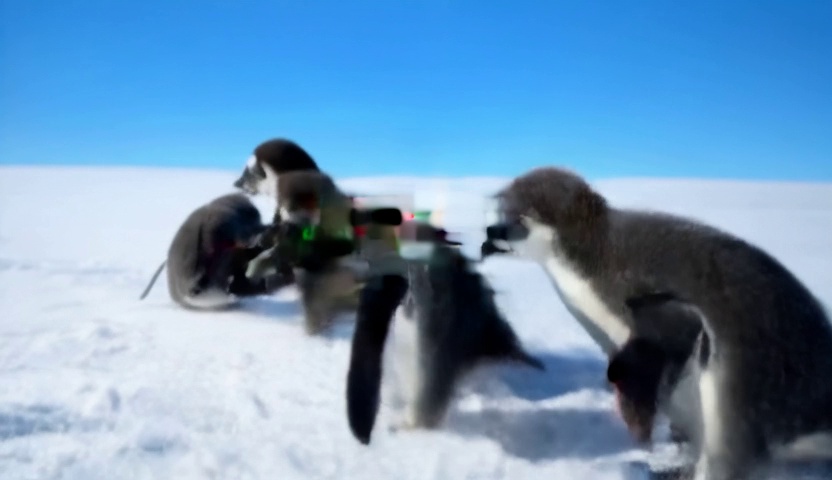}\hspace{0.5pt}  
        \\
        \raisebox{0.cm}{\rotatebox{90}{ 
            \sffamily\tiny\parbox{1.1cm}{\centering TrajectorC } }}  &
        \includegraphics[width=0.17\textwidth]{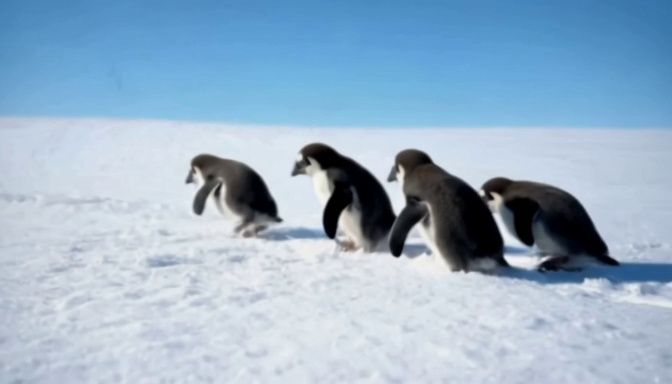}\hspace{0.5pt}  
        \includegraphics[width=0.17\textwidth]{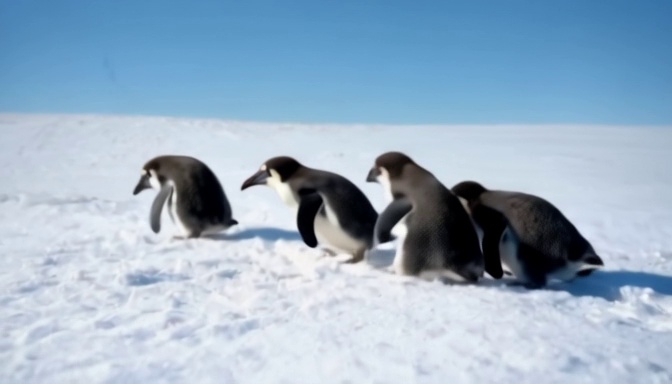}\hspace{0.5pt}  
        \includegraphics[width=0.17\textwidth]{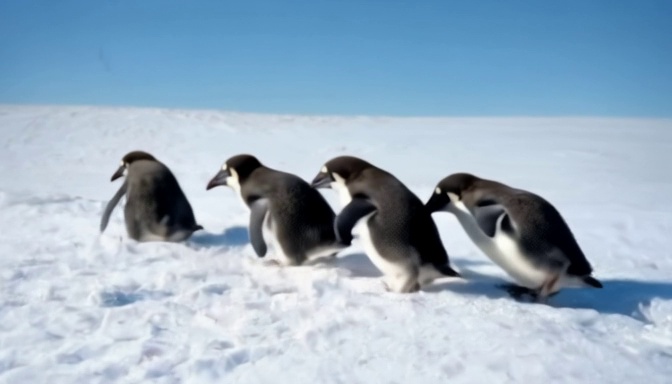}\hspace{0.5pt}  
        \includegraphics[width=0.17\textwidth]{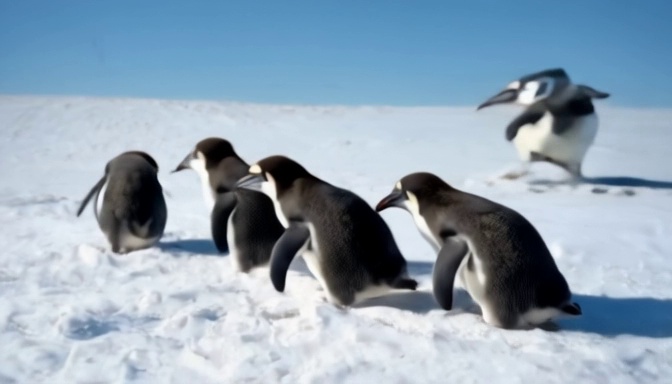}\hspace{0.5pt}  
        \includegraphics[width=0.17\textwidth]{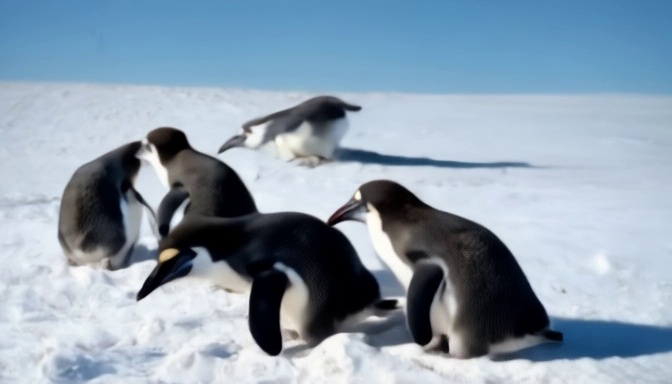}\hspace{0.5pt}     
        \\
        \raisebox{0.cm}{\rotatebox{90}{
            \sffamily\tiny\parbox{1.1cm}{\centering Ours} }}  &
        \includegraphics[width=0.17\textwidth]{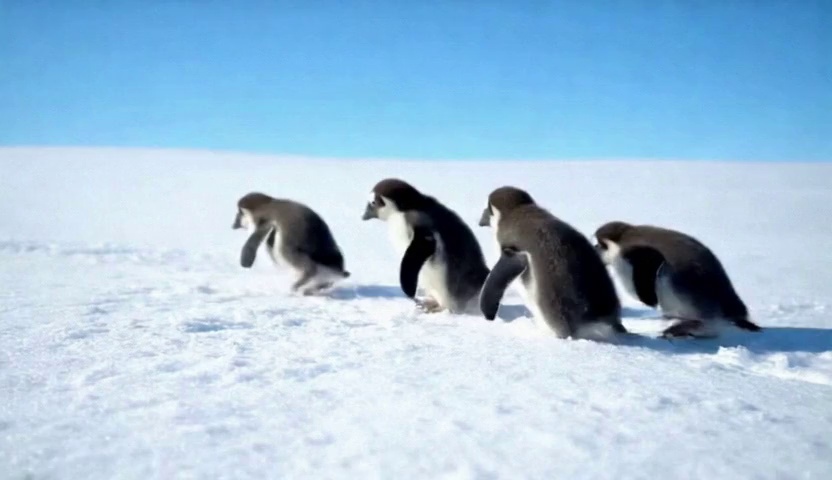}\hspace{0.5pt}  
        \includegraphics[width=0.17\textwidth]{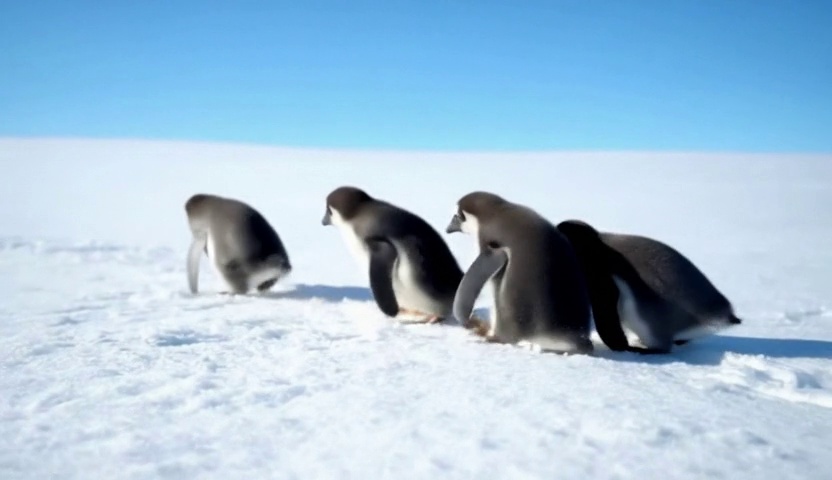}\hspace{0.5pt}  
        \includegraphics[width=0.17\textwidth]{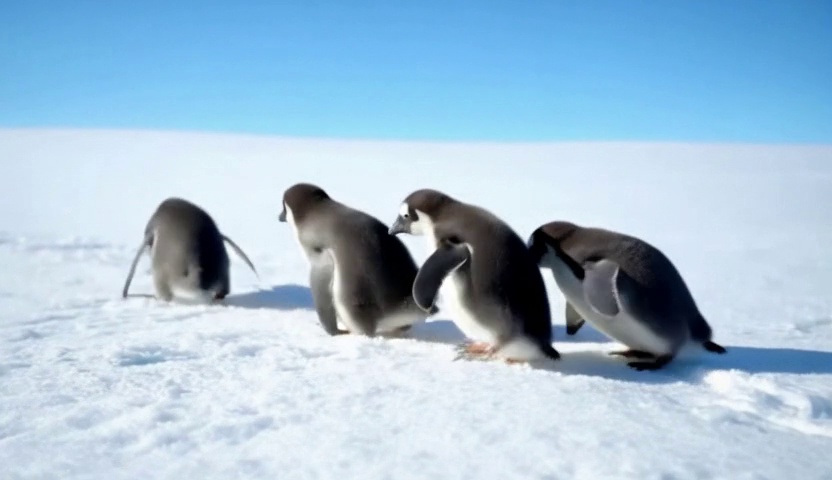}\hspace{0.5pt}  
        \includegraphics[width=0.17\textwidth]{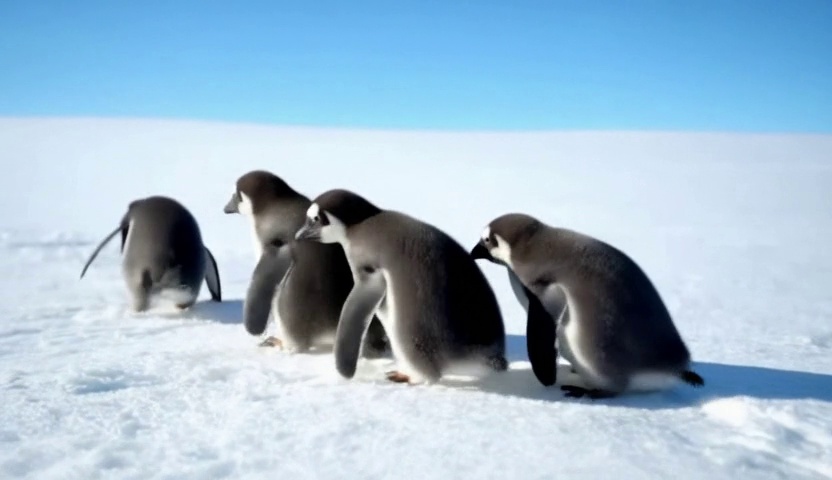}\hspace{0.5pt}  
        \includegraphics[width=0.17\textwidth]{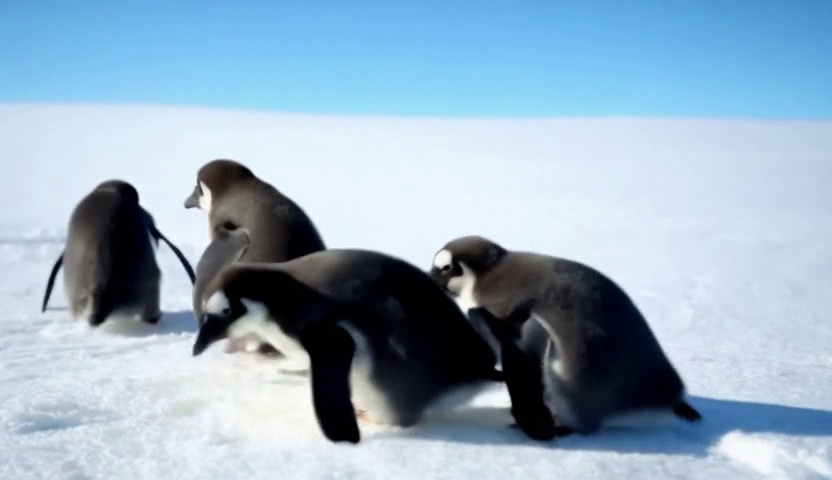}\hspace{0.5pt}   
   \end{tabular}
    \caption{\textbf{Additional qualitative comparison on generated video by Veo}.}
    \vspace{-0.1cm}
    \label{fig:con_compare_set1_p1}
\end{figure*}

\end{document}